\documentclass[11pt,letterpaper]{article}
\usepackage[T1]{fontenc}
\usepackage[textwidth=6.4in,top=0.88in,bottom=0.95in]{geometry}
\usepackage{newtxtext}
\usepackage{amsmath,amsthm}
\usepackage{newtxmath}
\usepackage{microtype,graphicx,booktabs}
\usepackage[round,authoryear]{natbib}
\usepackage[font=small,labelfont=bf,labelsep=period]{caption}
\usepackage{xcolor}
\definecolor{linkblue}{HTML}{234D70}
\usepackage[colorlinks=true,allcolors=linkblue,bookmarksnumbered=true]{hyperref}
\usepackage{bookmark}
\providecommand*{\doi}[1]{\href{https://doi.org/#1}{\nolinkurl{doi:#1}}}
\hypersetup{
  pdftitle={Gradient Orthogonalization: Optimality Thresholds and One-Step Gain Gaps},
  pdfauthor={Sixian Zeng},
  pdfsubject={Exact optimality and optimized one-step gain for polar updates},
  pdfkeywords={gradient orthogonalization, Muon, polar decomposition, high-dimensional probability}
}
\graphicspath{{figures/}}
\newtheorem{theorem}{Theorem}[section]
\newtheorem{lemma}[theorem]{Lemma}
\newtheorem{corollary}[theorem]{Corollary}
\newtheorem{proposition}[theorem]{Proposition}
\allowdisplaybreaks
\title{\bfseries Gradient Orthogonalization:\\
  Optimality Thresholds and One-Step Gain Gaps}
\author{Sixian Zeng\\[2pt]
  {\small\href{mailto:zengsx26@mails.tsinghua.edu.cn}{\nolinkurl{zengsx26@mails.tsinghua.edu.cn}}}}
\date{}
\begin{document}
\maketitle
\begin{abstract}
Giving every singular direction of a matrix gradient the same step length raises two questions. How steep must a displacement penalty be for this update to be exactly optimal, and how much one-step gain does it forgo at a specified penalty? We study these questions in the isotropic curvature model of Su and its random-radius extension by Xu. For uniform unit-sphere inputs and zero left slope at a kink, the least right slope, divided by the gradient nuclear norm, is within universal factors of \((1+\|T-I\|_F)/\sqrt{\det T}\), where the eigenvalues of \(T\) are the gradient singular values divided by their mean. One exponentially weak direction can force an exponential threshold even though the gain lost at a slope equal to the input width tends to zero. For a fixed number of nonuniform macroscopic spectral blocks, we construct thin-shell and Gaussian inputs with matching radial means and variances. The shell gap vanishes at a finite critical exponential slope rate. The Gaussian gap stays positive at every finite rate. Finally, in frozen checkpoints of three language-model training runs, rotating an update changed its measured loss cost while preserving every input displacement length. The model thus quantifies the effect of imposing a common step length under a radial cost, while the measurements identify a contribution from output orientation in the tested layers.
\end{abstract}

\section{Exact Optimality and the Gain Gap for Polar Updates}\label{sec:questions}
The singular values of a matrix gradient quantify the first-order loss decrease per unit step along its singular directions. A polar update keeps the singular directions and gives each the same length. This choice is central to Muon, which approximately orthogonalizes a momentum matrix \citep{muon}. We study one update with a fixed gradient. Two different comparisons arise even in this setting. One asks what would make a prescribed polar update the best among \emph{all} matrices. The other fixes the loss model and measures how much a polar update gives up after its step length has been tuned.

Let \(W\in\mathbb R^{m\times n}\) be a linear layer, \(x\in\mathbb R^n\) its input, and \(G=\nabla_W L(W)\) the gradient. Subtracting \(Q\) from \(W\) changes the output by \(-Qx\). The first-order loss change is \(-\langle G,Q\rangle\), with \(\langle G,Q\rangle=\operatorname{tr}(G^\top Q)\). The isotropic curvature model of \citet[Sec.~2.1, Eq.~(2)]{su} charges a finite, convex, nondecreasing cost \(H\) for the length of this displacement. With input law \(\nu\), the objective is
\begin{equation}\label{eq:su}
 J_\nu(Q)=-\langle G,Q\rangle+\mathbb E_{x\sim\nu}H(\|Qx\|_2).
\end{equation}
Here \(H\) models loss beyond the linear approximation. In Su's spherical model \(x=\zeta\) is uniform on the unit sphere. In the random-radius model of \citet[Sec.~2.2]{xu}, \(x=R\zeta\) with \(R\ge0\) independent of \(\zeta\). Thus \(R=\|x\|_2\) is an input length and \(r=\|Qx\|_2\) is the output displacement length to which \(H\) applies.

Throughout the main theory \(G\) has full column rank, with thin singular value decomposition \(G=U\Sigma V^\top\). Write \(P=UV^\top\), \(S=(G^\top G)^{1/2}=V\Sigma V^\top\), and \(\tau=\operatorname{tr}S=\sum_i\sigma_i(G)\). Then \(G=PS\) and \(P^\top P=I_n\). For a right singular vector \(v_i\), the gradient step \(\eta G\) gives a displacement of length \(\eta\sigma_i(G)\), whereas \(tP\) gives length \(t\) on every unit input. On an input of radius \(R\), the latter displacement has length \(tR\).

\paragraph{What is already known about orthogonalization?}
\citet[Thm.~1]{su} gives a growth condition under which an optimal update can share the singular directions of the fixed gradient while having smaller or equal ratios between corresponding larger and smaller singular values. Complete equality of the step lengths requires a stronger condition. On the sphere, Su's Theorem~2 shows that sufficiently small left and sufficiently large right slopes at a kink \(c>0\) can make \(cP\) optimal. Remark~4.3 asks how large the slopes must be for a given gradient. Su's empirical power-law observation and the kink assumption have different roles: Figure~1 of \citet{su} measures an average cost over a range of displacements, whereas the slopes at \(c\) determine whether a prescribed step is exactly optimal.
\citet[Sec.~3.1, Thm.~1 and Cor.~3.2]{xu} gives a geometric criterion for this optimality and shows that, for atomless input radii under its integrability assumptions, a positive polar step can be exactly optimal only for equal gradient singular values. Xu's separate minimax model optimizes over input-direction laws. In \eqref{eq:su} we average over a specified law.

On matrix quadratics, \citet[Sec.~3]{shulgin_quadratic} compute the one-step decrease along the polar direction after optimizing its step length. In a positive-definite quadratic model over momentum singular-value bands, \citet[Thm.~3]{sengupta_equal} quantify the gap between the best common step and the best choice of a step for each band. For the radial objective \eqref{eq:su}, we estimate the slope threshold for exact optimality on the sphere and determine how the input radius distribution affects the gain forgone at a prescribed slope.

\paragraph{What slope makes a prescribed polar step optimal?}
First take \(R=1\). On the ray \(Q=tP\) the loss is \(\ell(t)=H(t)-t\tau\). Write \(A=H'_-(c)\), \(B=H'_+(c)\). Moving towards \(c\) from the left must have nonpositive slope and moving away to the right must have nonnegative slope. For this convex function the two conditions are also sufficient:
\[
 c\in\arg\min_{t\ge0}\ell(t)
 \quad\Longleftrightarrow\quad A-\tau\le0\le B-\tau .
\]
Thus \(c\) is an optimal choice of \(t\) on the ray whenever \(A\le\tau\le B\). Changing the relative lengths of different singular directions could still improve the objective. Hold \(G,c,A\) and the input law fixed. We denote by \(B_{\min}\) the least right slope for which \(cP\) minimizes \eqref{eq:su} over all matrices, with \(+\infty\) if no finite slope suffices. For example, \(H_B(r)=Ar+(B-A)(r-c)_+\), where \(y_+=\max(y,0)\), varies that slope and holds the kink and left slope fixed. For spherical inputs, every direction under \(cP\) lands at the kink, so the same threshold applies to curved profiles with the same one-sided slopes there.

Our first result, Theorem~\ref{conf:threshold}, estimates \(B_{\min}\) for each full-rank gradient when \(A=0\) and \(n\ge2\). Put \(T=nS/\tau\), whose eigenvalues are the singular values divided by their mean. Up to constants independent of width and spectrum, \(B_{\min}/\tau\) equals \((1+\|T-I_n\|_F)/\sqrt{\det T}\). Sections~\ref{sec:criterion} and~\ref{conf:geometry} show how the known optimality criterion leads to this estimate.

\paragraph{How much gain does the best polar step forgo?}
For this question we \emph{specify} the cost and optimize the step. Divide the objective by \(\tau>0\), scaling \(G\) and \(H\) together so that \(\tau=1\). Choose the reference displacement \(c=1\) and charge \(H(r)=B_n(r-1)_+\). The input width \(n\) is the number of singular directions. The subscript allows the prescribed marginal cost per unit of excess displacement to vary with width. We compare two gains on the very same gradient, input law and penalty:
\[
 V_n=-\inf_Q J_\nu(Q),\qquad
 V_{\mathrm{pol},n}=-\inf_{t\ge0}J_\nu(tP).
\]
Both are measured relative to the zero step, which has loss zero. When both are finite their difference is
\begin{equation}\label{conf:gap}
 \Delta_n=V_n-V_{\mathrm{pol},n}\ge0 .
\end{equation}
On the sphere the ray gain is \(t-B_n(t-1)_+\). For \(B_n>1\) its unique maximum is one at \(t=1\). For \(B_n=1\) every \(t\ge1\) attains one. For \(B_n<1\) even the ray gain is unbounded, and \eqref{conf:gap} is undefined.

For an atomless radius of finite mean, the hinge gives a short illustration of Xu's result. At any fixed \(t>0\), \(\Pr(tR=1)=0\). Only samples with \(tR>1\) contribute a penalty derivative. Independence and \(\mathbb E[\zeta\zeta^\top]=I_n/n\) give the full loss gradient
\[
 \nabla_Q J_\nu(tP)=\lambda(t)P-G,\qquad
 \lambda(t)=\frac{B_n}{n}\mathbb E[R\mathbf1_{\{tR>1\}}].
\]
At an interior minimizing length \(t_*\) on the ray, the component along \(P\) vanishes, so \(n\lambda(t_*)=\tau\). Full optimality also requires all the other components to vanish. For unequal singular values, the perturbation \(D=G-(\tau/n)P\) then gives a strict descent derivative \(-\|D\|_F^2\). The same obstruction at a differentiable sphere profile appears in the proof of \citet[Prop.~3.4]{su}. Following \citet[Sec.~2.2]{xu}, the effective sphere profile after averaging over \(R\) is \(\bar H(t)=\mathbb E_R[H(tR)]\). For our hinge and atomless radii of finite mean, \(\bar H'(t)=n\lambda(t)\) at every positive \(t\), so the kink in the sample penalty disappears after averaging.

Xu's criterion detects this failure of exact optimality. Our second result, Theorem~\ref{conf:matched}, quantifies the loss in gain: for the same nonuniform block spectrum it can vanish for a thin shell yet stay positive for a Gaussian input with the same first two radial moments.

\paragraph{One weak direction can require an exponential slope even as the gain gap vanishes.}
We can see both questions on one spherical example. Compare two gradients of nuclear norm one at width \(n\). The flat reference has every singular value equal to \(1/n\). Write \(a_i=\sigma_i/(1/n)=n\sigma_i\), the strength of direction \(i\) relative to the mean. The actual singular values still sum to one; the \(a_i\) have mean one. Fix \(\gamma>0\) and choose
\[
 a_1=e^{-\gamma n},\qquad a_2=\cdots=a_n=\frac{n-e^{-\gamma n}}{n-1}.
\]
On this spectrum we first ask for the least slope making \(P\) optimal, then compute an upper bound on the gain gap at the specified slope \(B_n=n\):
\begin{equation}\label{conf:outlier}
 \frac{\log B_{\min}}{n}\longrightarrow\frac{\gamma}{2},
 \qquad \Delta_n=O(n^{-1/2}).
\end{equation}
The weak direction matters differently in a product and in an average. The geometric mean of the actual singular values divided by that of the flat reference is
\[
 \rho_n=\left(\prod_i a_i\right)^{1/n}
       =e^{-\gamma}a_2^{(n-1)/n}\longrightarrow e^{-\gamma}.
\]
The determinant factor in the threshold is \(\rho_n^{-n/2}\), giving the exponent \(\gamma/2\). In the squared difference from the flat spectrum, the single weak direction instead receives weight \(1/n\). Section~\ref{conf:outlier-proof} computes that difference and shows why it controls the gain lost.

Sections~\ref{sec:criterion} and~\ref{conf:geometry} develop the sphere threshold, and Section~\ref{conf:gains} compares the two radial laws. Section~\ref{sec:checkpoint-tests} asks a complementary empirical question to Su's radial measurements: if every displacement length stays fixed, how much can the loss cost change when the output directions rotate? Section~\ref{sec:discussion} discusses the reach of both comparisons. A \hyperref[guide:appendices]{guide to the appendices} locates the proofs, the finite-width computations, and extensions to rank-deficient gradients and finite input samples.

\section{When Is a Prescribed Polar Step Globally Optimal?}\label{sec:criterion}
Return to the unit-sphere law \(x=\zeta\), which we denote by \({\rm sph}\). Fix a kink at output length \(c>0\) with slopes \(A,B\). Under \(Q=tP\) every input moves by length \(t\); choosing \(t=c\) therefore puts every input exactly at the kink. Section~\ref{sec:questions} gave the ray criterion \(A\le\tau\le B\). To test optimality over all matrices we now allow each input's displacement to change at a different rate. We derive the sphere criterion used in \citet[proof of Thm.~2]{su} and stated explicitly by \citet[Sec.~3.1]{xu}, then use it to pose the least-slope problem.

\paragraph{Which perturbations can improve the prescribed step?}
Change \(Q\) from \(cP\) to \(cP+\varepsilon D\), where \(D\in\mathbb R^{m\times n}\) is fixed and \(\varepsilon\downarrow0\). For input \(\zeta\), the output length changes from \(c\) to \(r_\varepsilon=\|(cP+\varepsilon D)\zeta\|_2\). Expanding the squared norm gives
\[
 \frac{r_\varepsilon-c}{\varepsilon}
 =\frac{2c\langle P\zeta,D\zeta\rangle+\varepsilon\|D\zeta\|_2^2}{r_\varepsilon+c}
 \longrightarrow d_\zeta=\langle P\zeta,D\zeta\rangle .
\]
If \(d_\zeta>0\) the input crosses to the right of the kink and the penalty derivative is \(B d_\zeta\). If \(d_\zeta<0\) it moves to the left and the derivative is \(A d_\zeta\). When \(d_\zeta=0\) the derivative is zero by local Lipschitz continuity. These are precisely the values of \(\max_{A\le a\le B}a d_\zeta\). Appendix~\ref{app:selector} justifies passing this pointwise derivative through the expectation.

For any measurable choice \(h(\zeta)\in[A,B]\), collect the contributions of the unit inputs into a matrix
\[
 M(h)=\mathbb E[h(\zeta)P\zeta\zeta^\top],\qquad
 \mathcal M_{A,B}=\{M(h): A\le h\le B\}.
\]
Its inner product with \(D\) is \(\mathbb E[h(\zeta)d_\zeta]\), because \(\langle P\zeta\zeta^\top,D\rangle=\langle P\zeta,D\zeta\rangle\). Thus \(M(h)\) records how one choice of marginal costs responds to matrix perturbations. For this particular \(D\) the right derivative has already selected a slope: take \(h_D=B\) where \(d_\zeta>0\) and \(A\) elsewhere. Every other \(h\) gives \(h(\zeta)d_\zeta\le h_D(\zeta)d_\zeta\) pointwise. Consequently
\[
 J'_{\rm sph}(cP;D)
 =\langle M(h_D)-G,D\rangle
 =\max_{M\in\mathcal M_{A,B}}\langle M-G,D\rangle .
\]
If \(G\in\mathcal M_{A,B}\), the maximum includes the value zero for every \(D\). Convexity then makes \(cP\) a global minimizer. If \(G\) lies outside this compact convex set, take its nearest point \(M_0\) in Frobenius norm. The component of \(G\) left unmatched by the allowed slopes gives a concrete improving perturbation:
\[
 D=G-M_0,\qquad J'_{\rm sph}(cP;D)=-\|G-M_0\|_F^2<0 .
\]
Indeed the projection inequality gives \(\langle M-M_0,D\rangle\le0\) for every feasible \(M\), with equality at \(M_0\). A sufficiently small positive move along \(D\) therefore lowers the loss.

Membership \(G\in\mathcal M_{A,B}\) means one \emph{common} selector \(h_*\) balances the entire gradient. Substitute \(G=PS\) and multiply on the left by \(P^\top\). Since \(P^\top P=I_n\), the known criterion takes the form
\begin{equation}\label{eq:selector}
 cP\in\operatorname*{arg\,min}_Q J_{\rm sph}(Q)
 \quad\Longleftrightarrow\quad
 S=\mathbb E[h_*(\zeta)\zeta\zeta^\top]
 \quad\text{for some measurable }A\le h_*\le B .
\end{equation}

\paragraph{The least right slope for a given gradient.}
Equation~\eqref{eq:selector} tests a specified interval \([A,B]\). We now fix the gradient and ask how small \(B\) can be while the test still passes. Start with \(A=0\). Taking the trace of the matrix identity gives \(\mathbb E h_*=\operatorname{tr}S=\tau\), since \(\operatorname{tr}(\zeta\zeta^\top)=1\). Dividing the identity by \(\tau\) normalizes both sides. The weight \(f=h_*/\tau\) integrates to one under uniform sphere probability, the target is the trace-one matrix \(C=S/\tau\), and the permitted density cap is \(B/\tau\). The least cap is
\begin{equation}\label{eq:density}
 K(C)=\inf\big\{\|f\|_\infty:
 f\ge0,\ \mathbb E f=1,\ \mathbb E[f(\zeta)\zeta\zeta^\top]=C\big\}.
\end{equation}
Here \(\|f\|_\infty\) is the essential supremum relative to uniform sphere probability. Every feasible density of cap \(L\) supplies a selector \(h_*=\tau f\) of cap \(\tau L\), and the converse follows by normalization. A finite infimum is attained (Appendix~\ref{app:attainment}), so \(B_{\min}=\tau K(C)\).

This is also a way to measure the size of the admissibility set in \citet[Eq.~(4)]{xu}. Write \(\mathcal K_n=\{\mathbb E[a(\zeta)\zeta\zeta^\top]:0\le a\le1\}\). For \(k>0\), if \(C\in k\mathcal K_n\), then \(C=\mathbb E[k a(\zeta)\zeta\zeta^\top]\). Taking traces gives \(1=\operatorname{tr}C=\mathbb E[ka]\), so \(f=ka\) is a density with cap \(k\). Conversely \(a=f/k\) lies in \([0,1]\) for any feasible density capped by \(k\). Thus \(K(C)=\inf\{k>0:C\in k\mathcal K_n\}\). The same quantity has a classical interpretation: \(1/K(C)\) is the zonoid depth of \(C\) for the random projector \(\zeta\zeta^\top\) \citep{zonoid}. The quantitative task here is to estimate this cap for the prescribed spectrum.

For comparison, Xu's necessary spectral bound at \(A=0\) is \(S\preceq (B/n)I_n\). It yields \(B_{\min}/\tau\ge n\lambda_{\max}(C)\). This is exact on a flat spectrum: \(f=1\) realizes \(C=I_n/n\) with cap one. On the single-weak-direction spectrum of \eqref{conf:outlier}, the same lower bound tends to one. Theorem~\ref{conf:threshold} shows that the actual cap grows exponentially there.

A positive left slope already contributes \(AI_n/n\) to the required matrix moment. For \(0<A<\tau\), subtract that contribution and write \(h=A+(\tau-A)f\). A finite threshold exists exactly when \(A<n\sigma_{\min}(G)\), in which case
\[
 C_A=\frac{S-(A/n)I_n}{\tau-A},\qquad B_{\min}=A+(\tau-A)K(C_A).
\]
At \(A=\tau\) only a flat spectrum is feasible, with limiting threshold \(\tau\) (attained when equal one-sided slopes are allowed). For \(A>\tau\), \(\mathbb E h=\tau\) rules out every choice. Appendix~\ref{app:selector} gives the boundary argument and the modification for an orientation-dependent cost.

\section{The Critical Slope for a Given Gradient Spectrum}\label{conf:geometry}
On the unit sphere, take a kink at \(c>0\) with left slope \(A=0\).
The moment criterion used by \citet[proof of Thm.~2]{su} and stated by
\citet[Sec.~3.1]{xu} characterizes when the prescribed step \(cP\) is
globally optimal. We estimate the least right slope \(B\) that satisfies
this criterion for a given full-column-rank gradient, answering the
quantitative question posed in \citet[Rem.~4.3]{su} for \(A=0\).
By \eqref{eq:density}, this threshold is \(B_{\min}=\tau K(C)\), with
\(C=S/\tau\). Thus it suffices to estimate the smallest density cap
under which the target moment \(C\) can be realized.

Write \(v\) for a point on the unit sphere, denoted by \(\zeta\) in
\eqref{eq:density}. Here the input \(x=R\zeta\) has \(R=1\), so the
input equals its direction. Let \(\sigma\) be uniform probability on the
sphere and \(\mu_f\) the probability obtained by weighting directions by
a density \(f\). For any bounded observable \(u\),
\[
 \mathbb E_{v\sim\mu_f}[u(v)]
 =\int_{\mathbb S^{n-1}}u(v)f(v)\,d\sigma(v)
 =\mathbb E_{v\sim\sigma}[f(v)u(v)].
\]
We abbreviate these expectations by \(\mathbb E_f\) and
\(\mathbb E_\sigma\), respectively. The constraint in \eqref{eq:density}
is \(\mathbb E_f[vv^\top]=C\). For a unit eigenvector \(e\) of
\(C\), the uniform mean squared projection is
\(\mathbb E_\sigma[(e^\top v)^2]=1/n\). Its target under the weighted
law is \(\mathbb E_f[(e^\top v)^2]=e^\top C e=\lambda_e(C)\).
The matrix \(T=nC\) measures these target values relative to the uniform
baseline. One density must match the entire matrix, including its
off-diagonal entries.

\begin{theorem}[Critical slope]\label{conf:threshold}
Let \(n\ge2\), let \(C\) be positive definite with trace one, and put
\(T=nC\) and \(F=\|T-I_n\|_F\). There are constants \(c_0,C_0>0\),
independent of \(n\) and \(C\), such that
\begin{equation}\label{conf:sharp}
 c_0(1+F)\ \le\ K(C)\sqrt{\det T}\ \le\ C_0(1+F).
\end{equation}
Writing \(\|T\|_{\rm op}\) for the largest eigenvalue of \(T\), we also have
\begin{equation}\label{conf:uniform}
 1\ \le\ K(C)\sqrt{\det T}\
 \le\ \min\{500n,\,800\sqrt{n\|T\|_{\rm op}}\}.
\end{equation}
\end{theorem}

\paragraph{Reading the slope threshold from the singular values.}
Let \(\mathrm{AM}=\tau/n\) and \(\mathrm{GM}=(\prod_i\sigma_i(G))^{1/n}\)
be the arithmetic and geometric means of the singular values. The eigenvalues
of \(T\) are \(a_i=\sigma_i(G)/\mathrm{AM}\), so
\[
 \rho_n:=(\det T)^{1/n}
 =\left(\prod_{i=1}^n a_i\right)^{1/n}
 =\frac{\mathrm{GM}}{\mathrm{AM}},\qquad
 F^2=\sum_{i=1}^n(a_i-1)^2.
\]
The flat spectrum of the same width and nuclear norm has every singular
value equal to \(\mathrm{AM}\). Thus \(\rho_n\) is the geometric-mean
ratio in the example in \eqref{conf:outlier}. Substituting in
\eqref{conf:sharp} gives the spectrum-specific estimate
\[
 \frac{B_{\min}}{\tau}\asymp (1+F)\rho_n^{-n/2}.
\]
Equation~\eqref{conf:uniform} gives explicit constants. In particular,
the minimum on its right is always at most \(500n\). This bound is
already sufficient for the exponential rate, since with
\(d_n=-\tfrac12\log\rho_n\) it gives
\[
 0\le \frac1n\log(B_{\min}/\tau)-d_n
 \le \frac{\log(500n)}{n}\longrightarrow0.
\]
Hence \(d_n\to d>0\) gives \(B_{\min}/\tau=\exp(nd+o(n))\).
For the single weak direction in \eqref{conf:outlier}, \(\tau=1\),
\(a_1=e^{-\gamma n}\), and the other \(n-1\) eigenvalues are
\(a_2=(n-e^{-\gamma n})/(n-1)\to1\). Consequently
\(\rho_n=e^{-\gamma}a_2^{(n-1)/n}\to e^{-\gamma}\), giving the claimed
rate \(\gamma/2\). The maximum eigenvalue tends to one in this example,
so the second term in \eqref{conf:uniform} bounds the correction by a
multiple of \(\sqrt n\) for large \(n\).
Section~\ref{conf:outlier-proof} uses \eqref{conf:sharp} for the
tighter threshold estimate and treats the gain at \(B_n=n\).

Two spectra illustrate the size of the multiplicative correction at
finite width. For a flat spectrum, \(T=I_n\) and the uniform density
\(f=1\) attains \(K=1\), so \(B_{\min}=\tau\). Now fix \(0<s<1\)
and let two singular values each equal \(s\tau/2\), with the remaining
\(n-2\) sharing \((1-s)\tau\) equally. For \(n>2/s\) these are the
two largest singular values. The exact two-level calculation in
Appendix~\ref{sec:determinant} (proof of Theorem~\ref{thm:uniform-det-cap}) gives
\[
 K(C)\sqrt{\det T}=\frac{ns}{2}.
\]
For \(s=1/2\), using the determinant factor alone underestimates the
threshold by a factor \(n/4\). Appendix~\ref{sec:beta},
Proposition~\ref{prop:beta}, gives the finite-width formula for
arbitrary multiplicities of the two singular values. Figure~\ref{fig:threshold}
illustrates its dependence on multiplicity at a fixed condition number.
For a fixed bound on the condition number, Theorem~\ref{thm:specht-cap}
gives the sharp large-width asymptotic for the largest normalized threshold
$B_{\min}/\tau$ over all spectra satisfying that bound.

\subsection*{Proof outline for Theorem~\ref{conf:threshold}}

We first establish the determinant lower bound in \eqref{conf:uniform}
and reduce its upper bound to a spherical probability estimate.
Appendix~\ref{sec:determinant} supplies the explicit constants.
We then explain the matching spectrum-dependent bounds in
\eqref{conf:sharp}, proved in Appendix~\ref{sec:frobenius}.

\paragraph{The determinant lower bound for the density cap.}
For the lower bound in \eqref{conf:uniform}, seek a positive function
with a known mean under uniform directions and a lower bound on its
mean under every feasible weighting. Set \(k(v)=v^\top T^{-1}v\).
The moment constraint forces
\[
 \mathbb E_f k
 =\operatorname{tr}\bigl(T^{-1}\mathbb E_f[vv^\top]\bigr)
 =\operatorname{tr}(T^{-1}C)=1 .
\]
The useful power of \(k\) comes from a Gaussian integral. In
\(\int_{\mathbb R^n}\exp(-y^\top T^{-1}y/2)\,dy\), the substitution
\(y=T^{1/2}z\) multiplies the standard integral by \(\sqrt{\det T}\).
In spherical coordinates \(y=rv\), integration along a fixed direction
gives the same radial integral as for \(T=I_n\), times
\(k(v)^{-n/2}\). Indeed, \(s=r\sqrt{k(v)}\) gives
\(r^{n-1}dr=k(v)^{-n/2}s^{n-1}ds\). Comparing the two evaluations yields
\(\mathbb E_\sigma[k^{-n/2}]=\sqrt{\det T}\). Now Jensen's inequality
for \(u\mapsto u^{-n/2}\) gives
\[
 1=(\mathbb E_f k)^{-n/2}\le\mathbb E_f[k^{-n/2}]
 \le \|f\|_\infty\,\mathbb E_\sigma[k^{-n/2}]
 =\|f\|_\infty\sqrt{\det T}.
\]
Taking the infimum over feasible \(f\) proves the lower bound
(Theorem~\ref{thm:det-cap}).

\paragraph{A uniform halfspace probability gives a feasible density cap.}
For the upper bound we need a density that realizes every entry of \(C\).
First consider a single linear test. A symmetric \(D\) assigns to a
moment \(M\) the score \(\operatorname{tr}(DM)\), a weighted sum of
entries. By convention the test passes when this score is at least
\(\operatorname{tr}(DC)\). Reversing the inequality gives the test for
\(-D\), so either convention yields the same family of halfspaces.
The projector \(vv^\top\) passes on the spherical event
\[
 E_D=\{v:v^\top Dv\ge\operatorname{tr}(DC)\}
     =\{v:v^\top D_0v\ge0\},
 \qquad D_0=D-\operatorname{tr}(DC)I_n .
\]
For \(C=\operatorname{diag}(3/4,1/4)\) on the unit circle, taking
\(D=\operatorname{diag}(1,0)\) tests just \(v_1^2\ge3/4\). The two
passing arcs occupy one third of the circle. Conditioning the uniform
law on these arcs gives density three there and zero elsewhere.
The conditional mean of \(v_1^2\) is at least \(3/4\). For \(D=I_2\),
both scores equal one and every direction passes.

Suppose there is one \(p>0\) such that the actual passing probability
\(q_D=\sigma(E_D)\) is at least \(p\) for \emph{every} symmetric \(D\).
For each test, draw \(v\) uniformly on the sphere and retain it when
it passes. Dividing the integral over the retained directions by their
probability gives the conditional moment
\[
 M_D=\frac{\mathbb E_\sigma[\mathbf1_{E_D}(v)vv^\top]}{q_D}
     =\mathbb E_\sigma[f_D(v)vv^\top],
 \qquad f_D(v)=\frac{\mathbf1_{E_D}(v)}{q_D}.
\]
This density integrates to one and satisfies \(f_D\le1/p\). Averaging
the passing scores gives \(\operatorname{tr}(DM_D)\ge\operatorname{tr}(DC)\).

Different tests can produce different \(M_D\). To recover one density
matching \(C\), consider the compact convex set of \emph{all} moments
attainable under the common cap \(1/p\) (Lemma~\ref{conf:halfspace}).
If \(C\) were outside this set, let \(M_0\) be its nearest point and
set \(D_*=C-M_0\). The projection inequality gives, for every attainable \(M\),
\[
 \operatorname{tr}(D_*M)\le\operatorname{tr}(D_*M_0)
 =\operatorname{tr}(D_*C)-\|C-M_0\|_F^2<\operatorname{tr}(D_*C).
\]
The same probability bound holds for \(D_*\). Its conditional moment
\(M_{D_*}\) is attainable and passes that very test, a contradiction.
Thus \(C\) is attainable and \(K(C)\le1/p\).

\paragraph{Estimating the passing probability under a Gaussian law.}
Write \(v=g/\|g\|_2\) for \(g\sim N(0,I_n)\). The positive radius
preserves signs, so \(q_D=\mathbb P_0(X\ge0)\) for
\(X=g^\top D_0 g\), where \(\mathbb P_0\) denotes the standard Gaussian
law. If \(D_0=0\) the test always passes. For nonzero \(D_0\) with
\(\mathbb E_0X=\operatorname{tr}D_0\ge0\), the fourth-moment estimate
in Theorem~\ref{thm:det-cap} gives \(q_D\ge1/60\).

When \(\mathbb E_0 X<0\), give higher scores more weight by multiplying
the Gaussian density by \(e^{tX}/m_t\), with \(m_t=\mathbb E_0 e^{tX}\).
Combining the quadratic exponents yields a new Gaussian law
\(\mathbb P_t=N(0,\Sigma_t)\), where
\(\Sigma_t=(I_n-2tD_0)^{-1}\) and \(m_t=\sqrt{\det\Sigma_t}\).
Choose \(t>0\) so that the new mean
\(\operatorname{tr}(D_0\Sigma_t)\) is zero.
Theorem~\ref{thm:uniform-det-cap} shows that exactly one such \(t\) lies
in \(0<t<1/(2\lambda_{\max}(D_0))\), where the new Gaussian density
is integrable. To compute the \emph{original} probability, undo the
reweighting on the passing event.
\[
 q_D=m_t\,\mathbb E_t[e^{-tX}\mathbf1_{\{X\ge0\}}]
 =\sqrt{\det\Sigma_t}\,
   \mathbb E_w[e^{-Z}\mathbf1_{\{Z\ge0\}}],
 \qquad Z=\tfrac12w^\top(\Sigma_t-I_n)w .
\]
Here \(g=\Sigma_t^{1/2}w\) under the tilted law, \(w\sim N(0,I_n)\),
and \(tX=Z\). To relate the prefactor to the target spectrum, put
\(R_t=\Sigma_t^{-1/2}T\Sigma_t^{-1/2}\). Since
\(\operatorname{tr}(D_0T)=0\), its trace is
\(\operatorname{tr}[(I_n-2tD_0)T]=n\). Its positive eigenvalues
therefore have product at most one. Thus
\(\det R_t=\det T/\det\Sigma_t\le1\).
Theorem~\ref{thm:uniform-det-cap} controls the weighted probability
in \eqref{conf:tilt} near \(Z=0\) and obtains, uniformly in \(D\),
\[
 q_D\ge\frac{\sqrt{\det T}}{\min\{500n,\,800\sqrt{n\|T\|_{\rm op}}\}} .
\]
Using this lower bound as \(p\) in the halfspace comparison proves the
upper bound in \eqref{conf:uniform}.

\paragraph{The spectrum-dependent correction to the determinant bound.}
For the upper bound in \eqref{conf:sharp}, restrict the tilted expectation
to \(0<Z<1\), where \(e^{-Z}\ge e^{-1}\).
The matrix \(\Sigma_t-I_n\) has trace zero by the choice of \(t\).
A Gaussian rotation estimate \citep{bobkovvolzone} gives this interval probability a lower
bound proportional to \(1/(1+\|\Sigma_t-I_n\|_F)\). Consequently
\(\sqrt{\det T}/q_D\) is at most a universal constant times
\(\sqrt{\det R_t}(1+\|\Sigma_t-I_n\|_F)\).
Theorem~\ref{thm:frobenius-cap} proves the matrix comparison that bounds
this last quantity by a constant times \(1+\|T-I_n\|_F\). The
halfspace argument then gives the required density cap.

For the matching lower bound, fix any feasible \(f\) and weight the
standard Gaussian law by \(f(g/\|g\|_2)\). The independent radius
retains its second moment \(n\), so the weighted second moment of
\(g\) is \(nC=T\). Hence the test function
\(\psi(g)=1+g^\top(I_n-T^{-1})g\) has weighted mean one.
Its positive part satisfies \(1\le\|f\|_\infty\mathbb E_0[\psi(g)_+]\).
Substitution of \(g=T^{1/2}w\) in the last integral gives
\[
 \mathbb E_0[\psi(g)_+]
 =\sqrt{\det T}\,\mathbb E_w[(1+Y)_+e^{-Y/2}],
 \qquad Y=w^\top(T-I_n)w,\quad w\sim N(0,I_n).
\]
Only \(Y\ge-1\) contributes. Lemma~\ref{lem:one-sided-quadratic}
bounds the probability of each unit interval there by \(2048/(1+F)\).
On \(j-1\le Y<j\) the weight is at most \((j+1)e^{(1-j)/2}\), whose
sum over \(j\ge0\) is finite. Therefore every feasible \(f\) satisfies
\(\|f\|_\infty\sqrt{\det T}\ge c_0(1+F)\).
Proposition~\ref{prop:frobenius-lower} completes this estimate with an
explicit constant. Together with Theorem~\ref{thm:frobenius-cap},
it proves \eqref{conf:sharp}.

\subsection{A single weak direction in the threshold and the gain}\label{conf:outlier-proof}
For the spectrum in \eqref{conf:outlier}, \(\tau=1\). Write \(x=e^{-\gamma n}=a_1\) and \(a=(n-x)/(n-1)=a_2=\cdots=a_n\). The eigenvalues of \(T=nS\) are exactly the \(a_i\). Since \(a=1+(1-x)/(n-1)\), we have \(0\le(n-1)\log a\le1-x\). In the notation of Section~\ref{sec:questions},
\[
 n\log\rho_n=\log\det T=-\gamma n+(n-1)\log a\in[-\gamma n,-\gamma n+1].
\]
Equation~\eqref{conf:uniform} sandwiches \(n^{-1}\log B_{\min}\) between \(\gamma/2-1/(2n)\) and \(\gamma/2+\log(500n)/n\). The sharper estimate \eqref{conf:sharp} even gives \(\log B_{\min}=\gamma n/2+O(1)\), since \(\|T-I_n\|_F^2=(1-x)^2n/(n-1)\le2\).

For the second claim in \eqref{conf:outlier}, fix \(B_n=n\). Let \(V_n(a_n)\) denote the best unrestricted gain for this normalized spectrum, and \(V_n(\mathbf1)\) the same quantity for the flat reference. On the weak-direction spectrum the best polar gain is one. On the flat reference the polar step is globally optimal by \eqref{eq:selector}, and so the best unrestricted gain there is also one. This equality of benchmarks lets us use a comparison between two different spectra to estimate the gap on one spectrum:
\[
 \Delta_n=V_n(a_n)-V_{\mathrm{pol},n}(a_n)=V_n(a_n)-V_n(\mathbf1).
\]
To compare the two values, evaluate the same aligned update, with singular lengths \(q_i\), on both spectra. The penalty is identical, since the input distribution and update are identical. Only the linear reward changes, by at most
\[
 \left|\frac1n\sum_i(a_i-1)q_i\right|
 \le
 \underbrace{\left(\frac1n\sum_i(a_i-1)^2\right)^{1/2}}_{\|a_n-\mathbf1\|_2}
 \left(\frac1n\sum_i q_i^2\right)^{1/2}.
\]
For bounded mean-square spectra and \(B_n\to\infty\), the root-mean-square step length of a nonnegative-gain candidate is at most \(\sqrt{2\pi}\) for sufficiently large \(n\). Optimizing gives the gain stability estimate \eqref{eq:gain-stability} in Appendix~\ref{sec:continuum}. For our spectrum the first factor is explicit:
\[
 \|a_n-\mathbf1\|_2^2=\frac1n\left[(1-x)^2+(n-1)
       \left(\frac{1-x}{n-1}\right)^2\right]=\frac{(1-x)^2}{n-1}.
\]
Combining these bounds gives
\[
 0\le\Delta_n\le\sqrt{2\pi}\,\frac{1-e^{-\gamma n}}{\sqrt{n-1}} .
\]
Thus an order-one deviation from the mean in just one of \(n\) directions has root-mean-square size \(O(n^{-1/2})\). This yields an upper bound on the loss of gain at \(B_n=n\), even though exact optimality on this same spectrum requires an exponential slope.

\section{Gain Gaps for Matched Shell and Gaussian Inputs}
\label{conf:gains}
For a given penalty, how much could a layer gain by giving different singular
directions different step lengths? In the random-radius extension of Su's
model \citep{su,xu}, an input has the form $X=R\zeta$, where $R$ is its
length and $\zeta$ is an independent uniform direction on the unit sphere.
\citet[Theorem~1 and Corollary~3.2]{xu} showed that for atomless radii,
under the integrability conditions of that result, a positive polar step
can be exactly optimal only if the gradient has
equal singular values. For a nonuniform spectrum this leaves open how close
the best polar gain comes to the unrestricted gain. We compute their
difference for two radial laws that agree in their first two moments.

Fix input width $n$, a full-column-rank gradient $G_n$ with nuclear norm
one, and its polar factor $P_n$. A displacement of length $r=\|QX\|$
costs $H_n(r)=B_n(r-1)_+$, where $y_+=\max(y,0)$. Beyond length one,
each additional unit of displacement costs $B_n$. For each input law and
this prescribed slope, compare the best gain $V_n$ over all matrices with
the best gain $V_{\mathrm{pol},n}$ over $tP_n$, $t\ge0$. In particular,
\[
 V_{\mathrm{pol},n}=\sup_{t\ge0}\{t-B_n\,\mathbb E[(tR-1)_+]\},
 \qquad \Delta_n=V_n-V_{\mathrm{pol},n}.
\]
The scalar $t$ is optimized separately for each law. On the sphere $R=1$,
so $B_n>1$ gives $V_{\mathrm{pol},n}=1$ at $t=1$.

\subsection{Matching the inputs while keeping the gradient fixed}
We compare two inputs with the same second moment at each width.
\begin{itemize}
\item For $g\sim N(0,I_n)$, the normalized Gaussian input $X_{\rm G}=g/\sqrt n$ has
radius $R_{\rm G}=\|g\|/\sqrt n$ and an independent uniform direction.
Its radius can exceed any fixed length.
\item The shrinking-shell input $X_{\rm U}=R_{\rm U}\zeta$ has a
uniform radius on $[\mu_n-\delta_n,\mu_n+\delta_n]$, where
$\mu_n=\mathbb E R_{\rm G}$ and $\delta_n=\sqrt{3(1-\mu_n^2)}$.
The endpoints are positive for large $n$ and both tend to one.
\end{itemize}
The two radii have exactly the same mean $\mu_n$. Their variances also agree:
$\operatorname{Var}(R_{\rm G})=1-\mu_n^2$ since $\mathbb E R_{\rm G}^2=1$,
and $\operatorname{Var}(R_{\rm U})=\delta_n^2/3$. Both input vectors
have mean zero and covariance $\mathbb E[R^2]\,I_n/n=I_n/n$. Thus a
difference in gains can arise with the same direction distribution,
gradient, slope, and input covariance.

To vary the step lengths in a controlled way, split the right singular
directions into a fixed number $J\ge2$ of blocks. The $k_{j,n}$ singular
values in block $j$ are equal. Write
\[
 \theta_j=\lim_{n\to\infty}k_{j,n}/n,\qquad
 s_{j,n}=\sum_{i\in\mathcal I_{j,n}}\sigma_i(G_n),\qquad
 s_j=\lim_{n\to\infty}s_{j,n}.
\]
Each vector $\theta,s$ has positive components summing to one.
The ratio $s_j/\theta_j$ is the limiting mean singular value in the
block relative to the mean over all directions. For an angular energy
allocation $u$ (nonnegative fractions summing to one), define
\[
 I(u)=\tfrac12\sum_j\theta_j\log(\theta_j/u_j),
\]
with $I=+\infty$ on the boundary. Its minimum is $I(\theta)=0$.
We consider $B_n\to\infty$ and $n^{-1}\log B_n\to b<\infty$.
For example, $B_n=n$ has $b=0$, while $B_n=e^{nb}$ has positive rate for $b>0$.
The two optimal gains are finite at all sufficiently large widths
under these hypotheses (Appendices~\ref{sec:phase}--\ref{sec:gaussian}).

\begin{theorem}[Matched radial moments and different limiting gains]\label{conf:matched}
Under the preceding block and slope assumptions, suppose $s\ne\theta$.
Write $\Delta_{n,{\rm U}}$ and $\Delta_{n,{\rm G}}$ for the gaps under the
shrinking-shell and normalized Gaussian laws defined above.
At $b=0$, the best polar gain tends to one under both laws, while each
unrestricted gain tends to $(\sum_j s_j^2/\theta_j)^{1/2}>1$.
Write $b_*=I(s)$. At every finite $b>0$,
\[
 \lim_{n\to\infty}\Delta_{n,{\rm U}}=0
     \quad\Longleftrightarrow\quad b\ge b_*,
 \qquad
 \lim_{n\to\infty}\Delta_{n,{\rm G}}>0.
\]
\end{theorem}

In particular, at and above $b_*$ the best common step loses vanishing
gain on the shrinking shell, even though Xu's criterion precludes
exact optimality of a positive common step at each sufficiently large
finite width. The Gaussian gap remains positive at the same rate.
At $b=0$ both laws give the same positive limiting gap, since
$\sum_j s_j^2/\theta_j=1+\sum_j(s_j-\theta_j)^2/\theta_j$.
The assumptions here give every block a positive limiting dimension
and gradient share. The single weak direction of
Section~\ref{conf:outlier-proof} requires the separate estimate given there.

\subsection{Why the shell has a threshold and the Gaussian keeps a gap}
\paragraph{The unrestricted gain reduces to one step length per block.}
Write $x_j\ge0$ for the length in block $j$ and
$U_{j,n}=\|\zeta_j\|^2$ for the fraction of the input direction's squared
norm in that block. Projecting onto the column space of $G_n$ preserves
the linear gain and can only shorten displacements. Averaging the projected
matrix over paired left and right rotations within each equal-value block
also preserves its linear gain. Convexity and rotational invariance ensure
that its expected penalty does not increase. Consequently the full matrix
optimization has value
\[
 V_n=\sup_{x_j\ge0}
 \left\{\sum_j s_{j,n}x_j
       -B_n\,\mathbb E\!\left[\left(R\sqrt{\sum_j U_{j,n}x_j^2}-1\right)_+\right]\right\}.
\]
The polar ray is the choice $x_1=\cdots=x_J=t$. Allowing separate
lengths can increase the linear gain, but the input directions for which
the displacement exceeds one then change as well.

\paragraph{On the sphere, the threshold is the rate of the angular allocation $s$.}
On the unit sphere the squared displacement is $\sum_j U_{j,n}x_j^2$.
The fractions $U_{j,n}$ have the law of independent $\chi^2_{k_{j,n}}$
variables divided by their sum. They concentrate near $\theta$, and $I(u)$
is the exponential rate for seeing angular fractions near $u$
(Theorem~\ref{thm:hinge-phase}). If the induced displacement exceeds one
on a neighborhood with rate below $b$, its expected penalty there grows
without bound because its probability decays more slowly than $B_n$ grows.
A strict violation at a point with $I(u)=b>0$ persists on moving slightly
toward $\theta$ into the lower-rate region. Conversely, shrinking all feasible
lengths by a common factor $1-\varepsilon$ places any remaining violations
at rates strictly above $b$. At $b=0$ only the typical allocation
$\theta$ survives in the limiting constraint. These estimates yield
\[
 V_{\rm sph}(b)=\max_{x\ge0}\left\{s^\top x:
     \sum_j u_jx_j^2\le1\quad\text{whenever }I(u)\le b\right\}
 =\min_{I(u)\le b}\left(\sum_j s_j^2/u_j\right)^{1/2}.
\]
For the second equality, Cauchy--Schwarz bounds the gain by the
right-hand expression for every allowed $u$. If $u^*$ minimizes it,
the first-order condition for the minimum makes the lengths
$x_j^*=s_j/(u_j^* V_{\rm sph}(b))$ feasible and they attain the bound
(Appendix~\ref{sec:phase}). The minimum possible value is one,
attained precisely at $u=s$. Thus the gain falls to the common-step
value one exactly when $I(s)\le b$.

\paragraph{A shrinking shell inherits the two sphere gains.}
For any input with $0<\ell_n\le R\le h_n<\infty$, evaluate the same update at
$\ell_n$, $R$, and $h_n$. Its linear gain stays fixed, and the penalty is
monotone in the radius. For a deterministic radius $r$, rescaling the update
from $Q$ to $rQ$ gives $W_{r,n}(B)=W_{1,n}(rB)/r$ for either optimal
gain $W$ (free or polar). Hence
\[
 \frac{W_{1,n}(h_nB_n)}{h_n}\le W_{R,n}(B_n)
 \le \frac{W_{1,n}(\ell_nB_n)}{\ell_n}.
\]
The endpoints for $R_{\rm U}$ converge to one, since
$\mu_n=1-1/(4n)+O(n^{-2})$. Both outer slopes have the same exponential
rate $b$ as $B_n$. Applying the bound separately to the two gains gives
$\lim_n\Delta_{n,{\rm U}}=V_{\rm sph}(b)-1$. The comparison also holds
for other shrinking shells with uniform marginal direction, allowing
dependence between radius and direction (Theorem~\ref{thm:shell}).

\paragraph{Large Gaussian radii constrain the common step.}
For a Gaussian input the block energies
$Z_{j,n}=n^{-1}\sum_{i\in\mathcal I_{j,n}}g_i^2$ sum to the random
squared radius. Write an energy vector as $z=hu$, with squared radius
$h=\sum_j z_j$ and angular shares $u=z/h$. Its exponential rate is
\[
 L(hu)=\tfrac12(h-1-\log h)+I(u).
\]
The Gaussian optimization imposes $\sum_j z_jx_j^2\le1$ for all
$z$ with $L(z)\le b$. Let $h_b\ge1$ solve $h_b-1-\log h_b=2b$.
This is the largest squared radius in that set, attained only at
$z=h_b\theta$. A common length therefore cannot exceed $1/\sqrt{h_b}$,
which is also its limiting gain. The unrestricted limiting gain is
$\min_{L(z)\le b}(\sum_j s_j^2/z_j)^{1/2}$. At $z=hu$ the expression
being minimized is at least $1/\sqrt h\ge1/\sqrt{h_b}$.
Equality would require $u=s$ and $h=h_b$ simultaneously. The largest
radius already uses the entire rate $b$, forcing $u=\theta$ there.
For $s\ne\theta$ these conditions are incompatible, giving a strictly
positive gap. Theorem~\ref{thm:gaussian-phase} controls the unbounded
tail in passing to this limit and gives the optimizing block lengths.

Replacing the linear hinge by $B_n(r-1)_+^p$ with fixed $p>1$ preserves
the respective limiting gains for the sphere and for the Gaussian input
(Appendices~\ref{sec:phase} and~\ref{sec:gaussian}). For a shrinking
shell the deterministic-radius comparison follows by rescaling the same
power penalty at both endpoints. In this family $B_n$ is a coefficient and
the right derivative of the penalty at one is zero. The numerical results
below use the linear hinge, whose right derivative is $B_n$.

\subsection{Limiting predictions and finite-width gains for two equal-size blocks}
\begin{figure}[htbp]
\centering
\includegraphics[width=.98\linewidth]{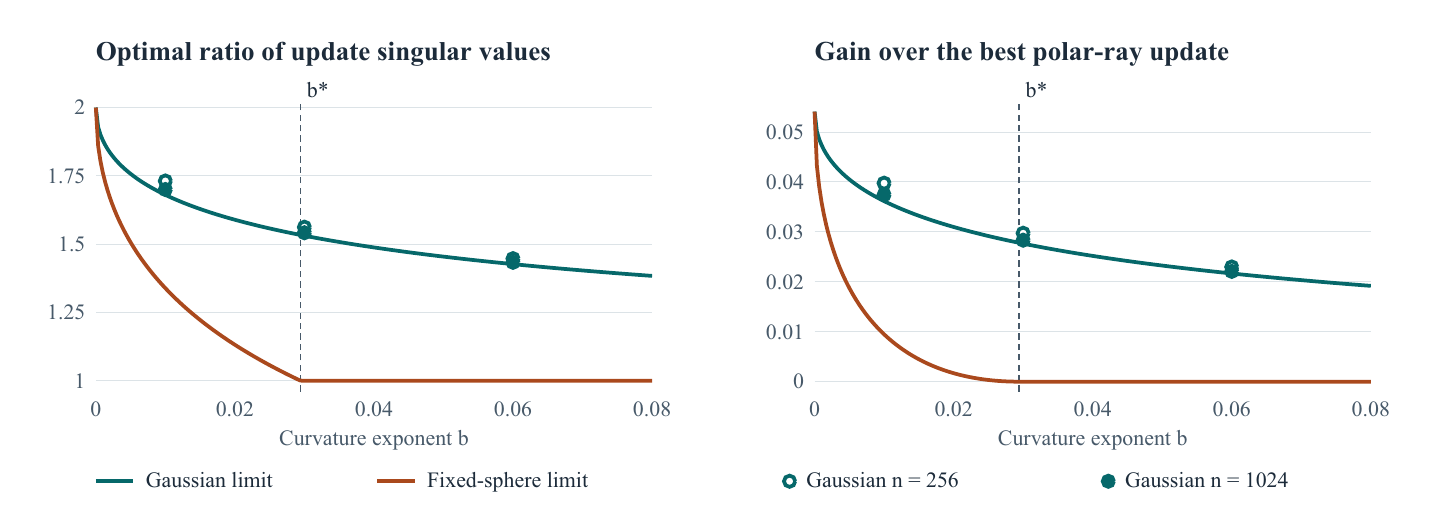}
\caption{Equal-size gradient blocks with singular values in ratio \(2:1\).
Curves show the sphere and Gaussian width limits of the optimal length
ratio \(x_1/x_2\) (left) and the absolute gain gap
\(\Delta(b)=\lim_n\Delta_n\) (right). Here \(x_1\) is the length in the
stronger block. The horizontal variable is the exponential rate
\(b=\lim_n n^{-1}\log B_n\), with \(B_n\to\infty\) also at \(b=0\).
The sphere and shell have the same limiting gains. The vertical dashed line is the shell threshold
\(b_*=\tfrac14\log(9/8)\). Open and filled circles are finite-width
Gaussian solutions at \(n=256,1024\), respectively, for \(b=0.01,0.03,0.06\).
The \(b=0.03\) markers lie just above \(b_*\).
Table~\ref{conf:finite-table} also evaluates \(b_*\) itself.}
\label{fig:radial}
\end{figure}

In Figure~\ref{fig:radial} each half of the gradient spectrum has a single
singular value, with the strong half twice the weak half. Thus
$\theta=(1/2,1/2)$, $s=(2/3,1/3)$, and $b_*=\tfrac14\log(9/8)\simeq0.02945$.
This choice holds the number of directions fixed between the two halves
while creating a clear incentive to allocate unequal lengths. Both update
families see this same gradient. Optimizing the lengths $x_1,x_2$
determines their ratio. For example, on the sphere
their limiting ratio decreases from $2$ at $b=0$ to $1$ at $b_*$
(Corollary~\ref{cor:two-block-phase}). The shell has the same limiting
gain as the sphere for every $b$ by Theorem~\ref{thm:shell}.

Theorem~\ref{conf:matched} concerns limiting objective values. A finite
threshold calculation shows the distinction at $b=b_*$. For this spectrum
on the sphere, Proposition~\ref{prop:beta} gives
$B_{\min}\sim\sqrt{\pi n}\,e^{nb_*}/(3e)$ along even widths. Setting
$B_n=e^{nb_*}$ therefore leaves $B_n/B_{\min}\sim3e/\sqrt{\pi n}\to0$.
The common step is strictly suboptimal at every sufficiently large even
width, yet its gain gap tends to zero. On the matched shell the absolute
gap likewise tends to zero, while on the Gaussian input it tends to a
positive value.

\paragraph{Computing the finite-width entries.}
For each radial law Table~\ref{conf:finite-table} optimizes both the free
block lengths and the common length using the same input distribution.
For two equal-size blocks $U_1\sim{\rm Beta}(n/4,n/4)$ and $U_2=1-U_1$.
Put $r_x=(U_1x_1^2+U_2x_2^2)^{1/2}$. At the positive optima the two
free-length equations and the separate common-length equation are
\[
 s_{j,n}=B_n x_j\,\mathbb E\!\left[\frac{R U_j}{r_x}
              \mathbf 1_{\{Rr_x>1\}}\right]\quad(j=1,2),\qquad
 1=B_n\,\mathbb E[R\mathbf 1_{\{Rt>1\}}].
\]
At either optimum, the linear gain cancels the displacement part of the
active hinge cost, leaving $B_n$ times the probability of crossing one.
Thus each entry is $B_n[\Pr(Rr_x>1)-\Pr(Rt>1)]$, evaluated at the two
separate optima. Appendix~\ref{sec:matched-radii} gives the radial
integrals, angular quadrature and numerical checks.

\begin{table}[htbp]
\caption{Absolute one-step gain gap on the equal-block \(2:1\) spectrum.
At each width, the two columns use radii with identical mean and variance.
A fixed slope of 16 gives a finite-width comparison. The choices \(B_n=n\) and
\(B_n=e^{nb}\) satisfy the diverging-slope assumptions of
Theorem~\ref{conf:matched}.}
\label{conf:finite-table}
\centering
\begin{tabular}{rlrr}
\toprule
$n$ & Slope $B_n$ & Uniform shell & Gaussian\\
\midrule
256  & $16$ & $0.0323102$ & $0.0390922$\\
1024 & $16$ & $0.0418919$ & $0.0457427$\\
256  & $n$  & $0.0146289$ & $0.0328547$\\
1024 & $n$  & $0.0251768$ & $0.0402433$\\
\midrule
256 & $e^{0.01n}$ & $0.0340780$ & $0.0397627$\\
256 & $e^{nb_*}$  & $0.0075060$ & $0.0299152$\\
256 & $e^{0.06n}$ & $0.0000657$ & $0.0229799$\\
\bottomrule
\end{tabular}
\end{table}

At $B_n=n$, both limiting gaps equal $\sqrt{10}/3-1\simeq0.0541$,
although at $n=1024$ the two values are still $0.0252$ and $0.0402$.
At $B_n=e^{nb_*}$ the shell gap at $n=256$ is $0.0075060$.
Above the threshold, at $b=0.06$ and $n=256$, the optimized length
ratio is $1.006608$ for the shell and $1.447020$ for the Gaussian,
with corresponding gaps of $0.0000657$ and $0.0229799$.
At $n=1024$ the Gaussian gap is $0.022087$, approaching its predicted
limit of $0.021673$ (Appendix~\ref{sec:matched-radii}).

The positive limiting Gaussian gap is a statement at fixed finite $b$.
Taking $n\to\infty$ first, then $b\to\infty$, the \emph{relative} gap
$V_{\rm G}(b)/V_{\mathrm{pol},\rm G}(b)-1$ decays as $1/b$
(Corollary~\ref{cor:large-budget}). Along a sequence
$b_n=n^{-1}\log B_n\to\infty$, both Gaussian gains are asymptotic
to $(2b_n)^{-1/2}$ (Proposition~\ref{prop:gauss-joint}).

\section{Step Allocation and Displacement Orientation at Frozen Checkpoints}
\label{sec:checkpoint-tests}
The surrogate suggests two experiments with distinct predictions. The block calculation motivates a candidate that gives stronger gradient directions longer steps. We can compare its measured loss with that of an equal-length polar step. For the second experiment, the radial penalty predicts that updates giving the same displacement length on every input will have equal costs after averaging opposite signs. We test both predictions at the same frozen weights.

We trained three 124M-parameter language models with different initializations on the same token stream, using Muon for the hidden-layer matrices. We froze each run at steps 192, 576 and 768, with step 768 prespecified as the primary endpoint. At each checkpoint we perturbed one of six \(768\times768\) matrices at a time, the key and value projections in blocks 0, 5 and 11. Each probe started at the saved weights.

\subsection{Constructing and calibrating the polar and two-block steps}

Within each of two disjoint document panels, we used 32 documents \(A\) to fit a direction, 64 further documents \(B\) to select its scalar amplitude and 128 held-out documents \(C\) to measure its loss. For each matrix \(W\), we computed the current gradient \(G_A=U\operatorname{diag}(\sigma_i)V^\top\) on \(A\), ordering the singular values from largest to smallest. The polar candidate \(D_{\rm pol}=UV^\top\) assigns every singular direction the same length. The two-block candidate retains \(U,V\) and replaces the first 384 singular values with their mean \(\bar\sigma_1\) and the remaining 384 with their mean \(\bar\sigma_2\). A raw-gradient reference retains the measured \(\sigma_i\).

These means serve two purposes. First, for any aligned update \(Q\) sharing a length \(x_j\) within half \(I_j\), its linear gain depends on the gradient in that half only through the mean,
\[
 \langle G_A,Q\rangle=\sum_{j=1}^2\sum_{i\in I_j}\sigma_i x_j
                       =\sum_{j=1}^2 |I_j|\bar\sigma_j x_j .
\]
Second, for a block spectrum on the sphere, Theorem~\ref{thm:hinge-phase} gives limiting optimal block lengths proportional to \(s_j/\theta_j\) when \(B_n\to\infty\) and \((\log B_n)/n\to0\). Here the fraction of nuclear norm in a half divided by its fraction of the dimensions is \(768\bar\sigma_j/\|G_A\|_*\). The common factor cancels in the ratio. We therefore use the measured \(\bar\sigma_1:\bar\sigma_2\) as a relative step allocation and test it on actual network loss.

\citet[Secs.~4--6]{wu_spectral} amplify momentum directions below the leading rank. At matched Frobenius norm, \citet[App.~E.2]{chaitanya_bulk} compare flat and two-band momentum updates on held-out loss, and \citet[Fig.~1]{sengupta_equal} find that reducing the step along the leading momentum directions improves one-step loss at some training checkpoints. Our two-block candidate gives the larger step to the upper half of the current gradient spectrum. A separate rotation experiment preserves the displacement length of every input, allowing us to test how output orientation affects the measured cost.

We normalized all three directions to \(\|D\|_F=\sqrt{768}\) and selected \(t_D\) separately for each direction by minimizing \(L_B(W-tD)\) over the same 16-point grid, which included zero. For example, in block 0's value projection at step 768 (seed 20261121, panel 1), the measured means had ratio \(9.52:1\). Both candidates selected \(t_D=0.01\). The polar step then had length \(0.01\) in every singular direction, while the two-block step had lengths \(0.01406\) and \(0.001477\). Thus the two allocations can differ even when their scalar amplitudes coincide. Appendix~\ref{app:checkpoints} gives the full grid and Table~\ref{tab:checkpoint-steps} records the selections.

\subsection{Why the polar step had lower held-out loss at step 768}

Let \(L_C\) denote held-out cross entropy in nats per predicted token. The measured gain over the unperturbed weights is \(g_D=L_C(W)-L_C(W-t_DD)\). The first endpoint compares \(g_{\rm two}-g_{\rm pol}\) on the same evaluation documents. It equals polar loss minus two-block loss, so a positive value favors two blocks. This signed contrast uses two fitted candidates on network loss. The theoretical \(\Delta_n\) in \eqref{conf:gap} optimizes over all matrices and over the entire polar ray in the surrogate.

At step 768, all three seed means favored the polar step. The paired mean was \(-8.63\) in units of \(10^{-6}\) nats per token (joint 95\% interval \([-13.91,-3.35]\), Figure~\ref{fig:frozen-probes}, left). Both directions had slightly higher mean loss than leaving the weights fixed. Their separate 95\% intervals for gain over zero included zero (Table~\ref{tab:checkpoint-audit}). The paired contrasts at the secondary checkpoints were \(-320.81\) at step 192 and \(-89.88\) at step 576, in the same units. The intervals quantify variation across the evaluation documents conditional on these three trained models, six matrices and calibrated steps.

To identify which part of the loss produced the ranking, we made a post hoc decomposition at each selected step. The held-out gradient gives the linear gain \(a_D\). Subtracting the observed gain from it gives the finite-step remainder \(u_D\). These terms satisfy
\[
 a_D=t_D\langle\nabla L_C(W),D\rangle,\qquad
 u_D=L_C(W-t_DD)-L_C(W)+a_D,\qquad g_D=a_D-u_D .
\]
The first term uses the gradient on the evaluation documents. The separate gradient on \(A\) was used only to construct the direction. A positive \(u_D\) measures how much less gain the step achieves than its local linear prediction.

\begin{table}[htbp]
\centering
\caption{Held-out gains and costs at the selected steps, in \(10^{-6}\) nats per token. All columns are averages over the same 36 step-768 fits (three seeds, two panels, six matrices). Means and contrasts are rounded independently.}
\label{tab:frozen-gain-components}
\begin{tabular}{@{}lrrr@{}}
\toprule
Quantity & Polar & Two blocks & Two blocks minus polar\\
\midrule
Linear gain \(a_D\)                & \(22.89\)  & \(25.27\)  & \(+2.38\)\\
Finite-step remainder \(u_D\)      & \(38.49\)  & \(49.49\)  & \(+11.00\)\\
Measured gain \(g_D=a_D-u_D\)      & \(-15.59\) & \(-24.22\) & \(-8.63\)\\
\bottomrule
\end{tabular}
\end{table}

The two-block direction had the larger linear gain, but its additional remainder was larger still (Table~\ref{tab:frozen-gain-components}). To estimate this remainder from local curvature, we measured two additional losses along each direction on the calibration documents at \(h=0.0025\). The resulting gain estimate was
\[
 \widehat g_D=a_D-\frac{t_D^2}{2}\,
 \frac{L_B(W+hD)+L_B(W-hD)-2L_B(W)}{h^2}.
\]
Its mean two-block minus polar gain was \(-7.59\). Across the 36 paired fits its mean absolute prediction error fell from \(29.04\) using the evaluation gradient alone to \(1.85\) with the calibration curvature term, in the same units. This is a retrospective calculation at the selected amplitudes.

\subsection{Equal-length displacements have different even costs}

To isolate the effect of output orientation, take the polar direction \(D\) and a signed permutation \(O\) of its rows. Since \(O^\top O=I\), each input \(x\) has exactly the same displacement length under \(tD\) and \(tOD\). The gradient can still score the two directions differently. Averaging losses at the positive and negative perturbations cancels the linear term in \eqref{eq:su}. The resulting even cost is
\[
 E_t(D)=\tfrac12\{L_C(W+tD)+L_C(W-tD)\}-L_C(W).
\]
Any length-only penalty therefore predicts \(E_t(OD)-E_t(D)=0\) for every input law. This equality holds at finite \(t\) and requires no fitted penalty slope.

\begin{figure}[htbp]
\centering
\includegraphics[width=.98\linewidth]{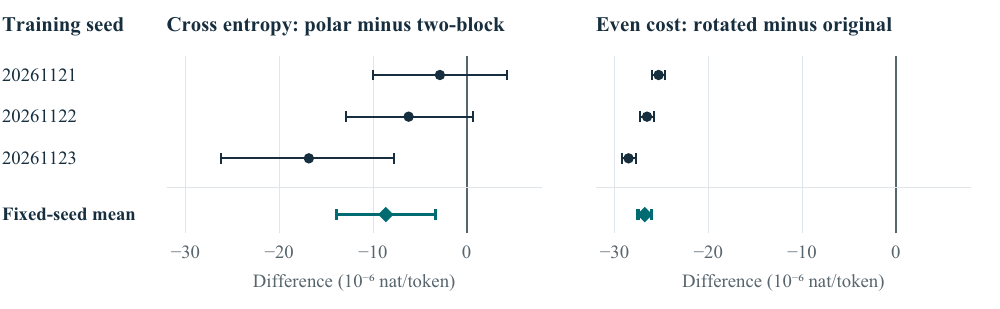}
\caption{One-step probes at training step 768, in \(10^{-6}\) nats per token. The left panel shows polar minus two-block cross entropy after separate calibration (positive favors two blocks). The right panel shows rotated minus original even cost at the fixed amplitude \(t=0.01\) (zero is the radial prediction). Each estimate averages six matrices and two document panels. The right panel also averages two signed permutations. Seed bars are pointwise 95\% paired document intervals. For the means over the three fixed seeds we show simultaneous 95\% document-bootstrap intervals for the two endpoints.}
\label{fig:frozen-probes}
\end{figure}

At the prespecified amplitude \(t=0.01\), the original and rotated even costs averaged \(31.46\) and \(4.68\) in units of \(10^{-6}\) nats per token. The paired difference was \(-26.78\) (joint 95\% interval \([-27.49,-26.08]\), Figure~\ref{fig:frozen-probes}, right). Any common predicted cost for the two orientations must therefore miss at least one observed mean by at least half their separation, \(13.39\). At the earlier checkpoints the same contrast averaged \(-110.48\) at step 192 and \(-33.93\) at step 576. Appendix~\ref{app:checkpoints} gives the matrix and amplitude comparisons (Table~\ref{tab:checkpoint-rotations}), cross-panel evaluation (Table~\ref{tab:checkpoint-panels}) and the paired-document uncertainty calculation.

This rotation comparison holds every input's displacement length fixed. The comparison between polar and two-block updates also changes the length distribution and can change the selected amplitude. In the step-768 fitting documents the top half of the gradient directions contained \(82.3\%\) to \(88.1\%\) of measured input energy across fits. At the same scalar amplitude the two-block direction produced \(1.280\) to \(1.325\) times the polar root-mean-square displacement on those inputs, despite the common Frobenius norm. The ranking therefore reflects the aggregate cost of the fitted steps. The signed permutations isolate the contribution of output orientation at fixed displacement lengths.

\section{Discussion}\label{sec:discussion}
\paragraph{What does a large threshold imply for a single update?}
The threshold is the least right slope for which the prescribed matrix attains the global minimum. Its size alone says little about how much better those alternatives can be. In the single-weak-direction example, one exponentially small relative singular value changes the determinant of the normalized spectrum by an exponential factor. At \(B_n=n\) the best polar step still achieves a net gain of one, within \(O(n^{-1/2})\) of the unrestricted optimum. For spectra with macroscopic strong and weak blocks, a positive fraction of the directions can instead profit from separate lengths. Even then, two inputs with identical covariance can assign very different costs to the same lengths. These conclusions concern the loss in one step at a fixed gradient. They explain why observing a nonuniform spectrum, or knowing that exact optimality fails by Xu's criterion, does not determine the size of the gain available from a change of update.

\paragraph{How should a slope in the model be interpreted?}
At a displacement kink \(c\), changing variables \(Q=c\widetilde Q\) and dividing the objective by \(c\tau\) gives the penalty \(H(cr)/(c\tau)\). Its right slope at \(r=1\) is \(H'_+(c)/\tau\): the marginal cost of excess displacement measured relative to the linear reward on the unit polar ray. A Hessian measures a second derivative. Estimating this slope in a network would require specifying a scale for the inputs and measuring the radial loss profile on both sides of a proposed kink. Su's measurements of approximate power-law growth on a finite range motivate attention to displacement length \citep[Fig.~1]{su}. The threshold calculation concerns local slopes at a kink; the hinge in our gain comparison specifies the cost at all lengths. The persistence of the limiting comparisons for the differentiable excess-length penalty \(B_n(r-1)_+^p\), at fixed \(p>1\), illustrates the distinction. In that model \(B_n\) is a coefficient and the derivative at \(1\) is zero (Appendices~\ref{sec:phase}--\ref{sec:gaussian}).

The exponential rate \(b\) measures how much rare inputs can contribute after their penalty has been multiplied by \(B_n\). At \(b=0.06\) and \(n=1024\), \(e^{nb}\) is about \(4.82\times 10^{26}\). Table~\ref{conf:finite-table} also reports ordinary finite slopes, where finite-width effects are visible. A useful benchmark at diverging subexponential slopes is the limiting unrestricted gain \(\lim_n\sqrt n\,\|G_n\|_F/\|G_n\|_*\) for macroscopic blocks, compared with limiting polar gain one. It follows from Theorem~\ref{conf:matched} by writing each block's mass in terms of its singular value. Writing the gradient's nuclear rank as \(\mathrm{nr}(G)=\|G\|_*^2/\|G\|_F^2\), the finite-width ratio here is \(\sqrt{n/\mathrm{nr}(G_n)}\). \citet[Sec.~1]{davis} use nuclear rank together with activation stable rank to compare descent guarantees for spectral and Euclidean steps.

\paragraph{Which assumptions determine the conclusions?}
Theorem~\ref{conf:threshold} uses a full-column-rank gradient and uniform unit inputs. Appendix~\ref{app:rank} gives a different answer for a compact polar factor of rank \(r<n\): a unit input then loses its component in the null space, so almost every displacement at \(cP_r\) falls strictly below \(c\). For the hinge, \(cP_r\) cannot be optimal for a nonzero gradient. Even with full rank, a finite sample of directions changes the moment test. Appendix~\ref{sec:sample} shows that some finite slope suffices exactly when the target \(C\) is in the convex hull of the sampled projectors. Its probability bound assumes independently sampled directions and a fixed gradient independent of these samples. A minibatch that also determines \(G\) can violate this independence.

The shell result extends beyond finitely many blocks. If the normalized spectral profiles \(a_n\) converge in mean square to \(a\), the critical rate is \(-\tfrac12\int_0^1\log a(u)\,du\) under the hypotheses of Theorem~\ref{thm:continuum} and the shrinking-shell comparison. For example, if half the spectrum tends to zero and half to two, the critical rate is infinite and the shell retains a positive gap at each finite rate. This contrasts with the vanishing fraction of weak directions in \eqref{conf:outlier}. For Gaussian inputs and fixed nonuniform positive block masses, the gap is positive at each finite rate. Its \emph{relative} size can nevertheless be small at large rates; Corollary~\ref{cor:large-budget} quantifies this after taking the width limit.

\paragraph{What did the frozen-layer tests establish?}
The orientation test gives the cleanest connection between the model and measured loss. An output rotation preserves every displacement length, so any purely radial penalty gives the same prediction for the two orientations. Their different measured even costs demonstrate a missing dependence on orientation at the tested layers and amplitudes. The ranking of a polar and a two-block update asks a different empirical question: the fitted candidates change the lengths as well as the output directions. At the final checkpoint the polar candidate had the smaller held-out loss, while both gains relative to leaving the weights fixed had intervals containing zero. These comparisons are conditional on the frozen models, matrices and evaluation protocol (Section~\ref{sec:checkpoint-tests}). The experiment measures neither the slope threshold of a Transformer layer nor a training-speed improvement.

A tractable extension adds the quadratic cost \(\frac{\varepsilon}{2}\mathbb E[(Qx)^\top M(\zeta)(Qx)]\), with \(0\preceq M(\zeta)\preceq I\). Proposition~\ref{prop:directional-cap} shows how its gradient shifts the exact moment condition. Even an arbitrarily weak orientation effect can disrupt the alignment needed for exact optimality. The value of the gain is more stable: under the block and radial assumptions at each finite rate, the absolute change in the gap is at most \(\varepsilon/2+o(1)\). For the specific block-symmetric family of Proposition~\ref{prop:directional-gaps}, a sufficiently weak perturbation moves the shell threshold continuously and preserves a positive Gaussian gap at each finite rate. The rotation data demonstrate sensitivity to orientation but do not identify this family. Estimating how much of the orientation residual can be predicted on fresh layers, and testing the resulting updates over complete training trajectories, remain empirical questions. Repeated updates also involve evolving gradients, momentum and approximate orthogonalization \citep{kovalev,kim}.

\paragraph{Research disclosure.}
Generative AI tools assisted with formulating the questions, comparing prior work, developing and revising the arguments and prose, writing numerical checks, designing and running checkpoint interventions, analyzing measurements, and preparing figures. Checks included separate derivations, numerical integration, and comparisons of objective values and stationarity equations. The finite-width calculations follow Appendix~\ref{sec:matched-radii}. The paired experimental contrasts were checked against the calibration records and aggregated by the resampling scheme in Appendix~\ref{app:checkpoints}.

\renewcommand{\bibsection}{\section*{\refname}\addcontentsline{toc}{section}{\refname}}
\bibliographystyle{abbrvnat}
\bibliography{references}

\clearpage
\appendix
\pdfbookmark[0]{Appendices}{appendices}
\section*{Guide to the Appendices}\label{guide:appendices}
\begin{description}
\item[Prescribed steps and spectral thresholds (A--D).]
Appendix~\ref{app:selector} supplies the analytic details of the known sphere criterion and its extensions to positive left slopes and direction-dependent costs. Appendix~\ref{sec:determinant} proves the explicit bounds in \eqref{conf:uniform} and computes the worst case at fixed condition number. Appendix~\ref{sec:frobenius} proves the matching bounds in \eqref{conf:sharp}. Appendix~\ref{sec:beta} gives an exact finite-width formula for a two-level spectrum and the prefactor in its large-width asymptotics.
\item[Gain at a specified penalty (E--I).]
Appendix~\ref{sec:phase} derives the optimal block lengths on the sphere. Appendix~\ref{sec:shell} transfers that result to a shrinking shell. Appendix~\ref{sec:gaussian} treats Gaussian radial tails. Appendix~\ref{sec:matched-radii} matches the two input laws, gives the equations for Table~\ref{conf:finite-table}, and quantifies the large-rate Gaussian gap. Appendix~\ref{sec:continuum} passes from finitely many blocks to a limiting spectrum and proves the stability estimate used in Section~\ref{conf:outlier-proof}.
\item[Departures from the population model (J--L).]
Appendix~\ref{app:rank} treats a gradient with a nontrivial null space. Appendix~\ref{sec:sample} studies finitely many independently sampled input directions. Appendix~\ref{app:checkpoints} provides the training, calibration and evaluation protocols, the loss decomposition, and sensitivity analyses for the frozen-checkpoint tests in Section~\ref{sec:checkpoint-tests}.
\end{description}

\section{The Optimality Condition for a Prescribed Polar Step}\label{app:selector}
Section~\ref{sec:criterion} obtained the sphere criterion \eqref{eq:selector} by testing every perturbation of \(cP\). We justify the limit under the expectation and the projection used there. We then extend the threshold to a positive left slope and examine how an orientation-dependent cost changes optimality.

Let \(\zeta\) be uniform on the unit sphere, let \(c>0\), and write \(A=H'_-(c)\), \(B=H'_+(c)\) for the one-sided slopes of the finite convex nondecreasing penalty. For fixed \(D\), the squared-norm identity in Section~\ref{sec:criterion} gives \((r_\varepsilon-c)/\varepsilon\to d_\zeta=\langle P\zeta,D\zeta\rangle\), with \(r_\varepsilon=\|(cP+\varepsilon D)\zeta\|_2\).
For \(d_\zeta\ne0\), the scalar secant slope \((H(r_\varepsilon)-H(c))/(r_\varepsilon-c)\) approaches the slope selected in Section~\ref{sec:criterion}. Multiplying by \((r_\varepsilon-c)/\varepsilon\) gives \(\max_{A\le a\le B}a d_\zeta\). When \(d_\zeta=0\), local Lipschitz continuity gives the same limit, zero. To integrate these limits, choose \(\varepsilon\) small enough that the reverse triangle bound \(|r_\varepsilon-c|\le\varepsilon\|D\|_{\rm op}\) places every radius in \([c/2,3c/2]\). If \(L\) is a Lipschitz constant of \(H\) on this interval, the absolute penalty difference quotient is bounded by \(L\|D\|_{\rm op}\), uniformly in \(\zeta\). Dominated convergence therefore gives
\[
 J'_{\rm sph}(cP;D)=\sup_{A\le h\le B}\langle M(h)-G,D\rangle_F,\qquad
 M(h)=\mathbb E[h(\zeta)P\zeta\zeta^\top].
\]
The pointwise selector \(h_D\) from Section~\ref{sec:criterion} attains this supremum.

The measurable selectors form a weak-star compact convex order interval in \(L^\infty\). Each entry of \(M(h)\) integrates \(h\) against a bounded function on the sphere, so their image \(\mathcal M\) is compact and convex. If \(G\notin\mathcal M\), let \(M_0\) minimize \(\|G-M\|_F\) over \(\mathcal M\), and put \(D=G-M_0\). For every \(M\in\mathcal M\), the segment \(M_0+s(M-M_0)\), \(0\le s\le1\), stays in the set. Its squared distance to \(G\) has a minimum at \(s=0\), giving \(-2\langle M-M_0,D\rangle_F\ge0\) by right differentiation. Hence
\[
 \langle M-G,D\rangle_F
 =\langle M-M_0,D\rangle_F-\|D\|_F^2
 \le-\|D\|_F^2<0 \qquad(M\in\mathcal M).
\]
The directional derivative is strictly negative. If \(G\in\mathcal M\), taking \(M=G\) instead gives a nonnegative supremum in every direction. Convexity gives \(J_{\rm sph}(cP+D)-J_{\rm sph}(cP)\ge J'_{\rm sph}(cP;D)\ge0\). Using \(G=PS\) and \(P^\top P=I_n\) now gives \eqref{eq:selector}.

\paragraph{How a positive left slope shifts the target moment.}
Taking traces in the moment condition of \eqref{eq:selector} gives \(\mathbb E h=\tau\). If \(0\le A<\tau\), put \(h=A+(\tau-A)f\). Then \(\mathbb E f=1\), and substitution into the same condition gives
\[
 S=\frac{A}{n}I_n+(\tau-A)\,\mathbb E[f\zeta\zeta^\top],
 \qquad C_A=\frac{S-(A/n)I_n}{\tau-A}=\mathbb E[f\zeta\zeta^\top].
\]
Here \(\mathbb E\zeta\zeta^\top=I_n/n\) by rotation symmetry and trace one. The restriction \(h\le B\) becomes \(f\le (B-A)/(\tau-A)\). For \(C_A\succ0\), the least feasible cap is attained (Appendix~\ref{sec:determinant}), giving \(B_{\min}=A+(\tau-A)K(C_A)\). A negative eigenvalue of \(C_A\) contradicts positivity of a second moment. For a zero eigenvalue with unit eigenvector \(v\), the equality \(\mathbb E[f(v^\top\zeta)^2]=0\) would concentrate the weighted probability on a great subsphere of surface measure zero. In either case no finite right slope suffices.

At \(A=\tau\), \(h\ge\tau\) and \(\mathbb E h=\tau\) force \(h=\tau\) almost surely. Thus a selector exists exactly when \(S=(\tau/n)I_n\), as witnessed by the constant selector. In this case every genuine kink \(B>A\) is feasible, with infimum \(\tau\). Allowing equal one-sided slopes attains \(B=A=\tau\). For \(A>\tau\), the lower bound on \(h\) contradicts \(\mathbb E h=\tau\).

\paragraph{Minimum density caps, zonoid depth and related moment problems.}
The reciprocal \(1/K(C)\) equals the zonoid depth of \(C\) for the random rank-one projector \(\zeta\zeta^\top\) \citep{zonoid}. For two spectral levels, write \(U\) for the squared input projection onto the higher level and \(s\) for its required share of spectral mass. If \(q_*\) satisfies \(\mathbb E[U\mid U\ge q_*]=s\), Proposition~\ref{prop:beta} gives \(1/K(C)=\Pr(U\ge q_*)\), the buffered exceedance probability at \(s\) \citep{buffered}. Minimizing relative entropy at a prescribed second moment leads to Bingham's quadratic exponential family \citep{bingham}. In radial isotropic position, the weights are given and the directions transformed \citep{barthe,radialposition}. In \eqref{eq:density}, the reference directions are given and the weights are chosen to realize \(C\). Comparisons between halfspace depth and the Cram\'er transform on full-dimensional convex bodies \citep{pafis} motivate the entropy calculation in Appendix~\ref{sec:determinant}. Here the projector law is supported on a lower-dimensional set.

Write \(\zeta=g/\|g\|_2\) for independent standard normal coordinates \(g_i\). The variables \(Y_i=g_i^2/2\) are independent gamma variables with shape \(1/2\) and unit scale. Normalization gives \(\zeta_i^2=Y_i/\sum_jY_j\), so the joint law of the squared coordinates is Dirichlet with all parameters \(1/2\) \citep{shadow}. Related weights occur in random spectral measures \citep{dirichlet}. At a simple extreme eigenvalue the corresponding shape is \(1/2\), below the endpoint hypotheses of \citet[Corollary~2]{weighted}. Our bounds include this case. Rare-event constraints in optimization have also been studied for linear chance-constrained programs \citep{chance}. Here large deviations enter through an expected hinge penalty, and the radius distribution determines which rare inputs contribute.
\subsection{An orientation-dependent displacement cost}\label{app:directional-cost}
Consider \(x=R\zeta\) and let \(M(\zeta)\) be a bounded measurable positive semidefinite \(m\times m\) matrix. At \(\varepsilon\ge0\) and \(\mathbb E R^2<\infty\), set
\begin{equation}\label{eq:directional-cost}
 J_\nu^\varepsilon(Q)=J_\nu(Q)+\frac{\varepsilon}{2}
   \mathbb E\big[(Qx)^\top M(\zeta)(Qx)\big].
\end{equation}
The added cost is convex in \(Q\). It allows the output curvature encountered by a displacement to depend on the input that produced it. The following results separate the symmetry needed for exact optimality from that needed for the gain comparison.

\begin{proposition}[Prescribed step with directional curvature]\label{prop:directional-cap}
On the unit sphere put \(N=\varepsilon c\,\mathbb E[M(\zeta)P\zeta\zeta^\top]\). Then \(cP\) minimizes \(J_{\rm sph}^\varepsilon\) if and only if
\[
 G-N=P\,\mathbb E[h(\zeta)\zeta\zeta^\top]\qquad\text{for some }A\le h\le B.
\]
In particular, if \(A=0\) and \(G-N=PS'\) for a symmetric \(S'\succ0\), Theorem~\ref{conf:threshold} applies with \(C'=S'/\operatorname{tr}S'\) and \(B_{\min}=(\operatorname{tr}S')K(C')\).
\end{proposition}
\begin{proof}
The gradient of the added quadratic at \(cP\) is \(N\). Adding \(\langle N,D\rangle\) to the directional derivative above gives the same separation argument with \(G-N\).
\end{proof}

Even arbitrarily weak directional curvature can preclude exact optimality for every finite right slope. Take \(n=m=2\), \(G=I_2/2\), \(P=I_2\), and let \(T\) be rotation by \(\pi/2\). Set \(M(\zeta)=ww^\top\) for \(w=(\zeta+T\zeta)/\sqrt{2}\). Then \(\|M(\zeta)\|_{\rm op}=1\) and \(N=(\varepsilon c/4)(I_2+T)\). For \(\varepsilon>0\), \(G-N\) has a nonzero skew-symmetric part, whereas every feasible moment is symmetric. At \(\varepsilon=A=0\) the constant selector \(h=1\) suffices for \(B\ge1\).

\paragraph{Bounds on the gain gap under directional curvature.}
First allow arbitrary \(0\preceq M_n(\zeta)\preceq I\), with no block symmetry. Under the finite-block and input assumptions of Theorem~\ref{conf:matched}, denote the new gap by \(\Delta_n^\varepsilon\). For fixed \(\varepsilon\),
\begin{equation}\label{eq:directional-gap-bound}
 \limsup_n|\Delta_n^\varepsilon-\Delta_n|\le\varepsilon/2 .
\end{equation}
Indeed, adding a nonnegative cost reduces each of the unrestricted and common-step gains. To bound each reduction from above, test an aligned optimizer of the corresponding radial problem. In the radial proofs its block scales \(y_{j,n}\) stay bounded and every accumulation point \(y\) is feasible for the limiting energy constraints. The typical energy vector \(\theta\) belongs to each rate sublevel, so \(\sum_j\theta_j y_j^2\le1\). Since \(\mathbb E[xx^\top]=I_n/n\), the quadratic costs at these test updates are at most \((\varepsilon/2)\sum_j(k_{j,n}/n)y_{j,n}^2\le(\varepsilon/2)(1+o(1))\). Both gain reductions thus lie in the same interval, which bounds their difference. The argument also gives \(\Delta_n^{\varepsilon_n}-\Delta_n\to0\) whenever \(\varepsilon_n\to0\). At fixed strength, a radial limiting gap greater than \(\varepsilon/2\) stays separated from zero.

For an exact comparison at fixed strength, let \(\Pi_j\) project onto the \(j\)th right singular subspace. An explicit family with different output costs uses fixed \(d_j\ge0\) and
\[
 w_j(\zeta)=\frac{P_n\Pi_j\zeta}{\|\Pi_j\zeta\|},\qquad
 M(\zeta)=\sum_j d_j w_j(\zeta)w_j(\zeta)^\top ,
\]
with \(w_j=0\) if the denominator vanishes. This family can distinguish equal-length displacements even for a square polar factor and uniform spherical inputs. For example, when all block sizes are even, choose an orthogonal output transformation \(O\) acting by quarter-turns in successive coordinate pairs of each left singular block. Each \(w_j(\zeta)\) is perpendicular to \(OP_n\zeta\). At amplitude \(t\), the rotated minus original sign-averaged cost under the unit-sphere model is therefore \(-(\varepsilon t^2/2)\sum_j d_j k_{j,n}/n\). This calculation illustrates a possible source of an orientation contrast. The checkpoint experiment does not identify \(M(\zeta)\).

\begin{proposition}[Limiting gaps with blockwise directional curvature]\label{prop:directional-gaps}
Use the sphere, matched shell or Gaussian law of Section~\ref{conf:gains}, and the block curvature above. Suppose \(\varepsilon\theta_jd_j<s_j\) for every block and define
\[
 p_j(t)=\frac{s_j-\varepsilon t\theta_jd_j}{1-\varepsilon t\sum_i\theta_i d_i}
 \qquad(0\le t\le1).
\]
For the unit sphere and matched shell, the limiting gap vanishes exactly when \(b\ge I(p(1))\). For the Gaussian law put \(t_b=h_b^{-1/2}\), where \(h_b-1-\log h_b=2b\), \(h_b\ge1\). The limiting Gaussian gap vanishes exactly when \(p(t_b)=\theta\).
\end{proposition}
\begin{proof}
Projection onto the column space of \(P_n\) preserves the quadratic cost. For each orthogonal input map \(T\) preserving the blocks, extend \(P_nTP_n^\top\) by the identity on the orthogonal complement to obtain an orthogonal output map \(\tilde T\). Then \(w_j(T\zeta)=\tilde T w_j(\zeta)\). Joint block rotations preserve the objective, so the convex averaging in Theorem~\ref{thm:hinge-phase} still reduces the problem to block scales \(y_j\). On these updates the quadratic cost is unchanged by taking absolute values, and equals \((\varepsilon/2)\sum_j d_j(k_{j,n}/n)y_j^2\) since \(\mathbb E[xx^\top]=I_n/n\) for all three laws.

Consequently the limiting maximand on the original feasible set is
\[
 f(y)=s^\top y-\tfrac{\varepsilon}{2}\sum_j d_j\theta_j y_j^2 .
\]
Indeed, the nonnegative added penalty preserves the boundedness of scales with nonnegative gain. On bounded sets the displayed finite-width quadratic converges uniformly. The diverging hinge cost at any strictly violating limit and the vanishing hinge cost at \((1-\delta)y\) for feasible \(y\), \(0<\delta<1\), give the same upper and lower limit arguments as in \eqref{eq:phase-primal} and \eqref{eq:gauss-primal}. For the shell, sandwich its hinge between those at its two endpoint radii, which both tend to one.

On the common-step ray the limiting feasible interval ends at \(t=1\) for sphere and shell and at \(t=t_b\) for the Gaussian law. The derivative of \(f(t\mathbf1)\) is \(1-\varepsilon t\sum_j\theta_jd_j>0\) throughout. At the respective endpoint \(t\mathbf1\), the gradient of \(f\) is a positive scalar multiple of \(p(t)\). By concavity and convexity of the feasible set, \(t\mathbf1\) maximizes \(f\) exactly when it maximizes the linear form \(p(t)^\top y\) on that set. Theorems~\ref{thm:hinge-phase} and~\ref{thm:gaussian-phase} give the two criteria upon replacing the mass vector by \(p(t)\).
\end{proof}
In particular, for \(s\ne\theta\) the shell threshold moves continuously with \(\varepsilon\) near zero. The Gaussian gap stays positive at every finite rate for all sufficiently small fixed \(\varepsilon\), because \(p(t)\to s\) uniformly on \([0,1]\). These conclusions use the joint block symmetry of the directional cost. A general input-dependent curvature need not permit the reduction to one length per block.

For the equal-block \(2:1\) spectrum of Table~\ref{conf:finite-table}, set \(\varepsilon=1\) and \(d=(1/3,0)\). Then \(p(1)=(3/5,2/5)\). The shell threshold falls from \(\tfrac14\log(9/8)\) to \(\tfrac14\log(25/24)\), and \(p_1(t)>1/2\) for \(0\le t\le1\), so the Gaussian gap stays positive. Raising \(d_1\) to \(2/3\) gives \(p(1)=\theta\) and closes even the Gaussian gap at \(b=0\).

\section{Determinant Bounds for the Critical Slope}\label{sec:determinant}
For \(A=0\), the critical slope satisfies \(B_{\min}/\tau=K(C)\). Lemma~\ref{conf:halfspace} converts a uniform lower bound on the probability of a halfspace into an upper bound on this density cap. Theorems~\ref{thm:det-cap} and~\ref{thm:uniform-det-cap} give the determinant estimate and the explicit finite-width factors in \eqref{conf:uniform}. Theorem~\ref{thm:specht-cap} then identifies the worst exponential rate at a fixed condition number. Appendix~\ref{sec:frobenius} gives the two-sided correction \eqref{conf:sharp} for an individual spectrum, and Proposition~\ref{prop:beta} gives exact two-level examples.

\label{app:attainment}
Throughout, $\sigma$ is uniform probability on $\mathbb S^{n-1}$, $n\ge2$. A feasible density $f\ge0$ integrates to one under $\sigma$ and has $\int vv^\top f(v)\,d\sigma(v)=C$. We set $K(C)=+\infty$ if no bounded density satisfies these constraints. When $K(C)<\infty$, the sets of feasible densities bounded by $K(C)+1/j$ are nested, nonempty, weak-star closed subsets of a weak-star compact ball in $L^\infty(\sigma)$. Indeed, the mass and moment constraints integrate against the bounded test functions $1$ and $v_i v_j$, while inequalities on $f$ can be tested against nonnegative $L^1(\sigma)$ functions. The intersection therefore contains a feasible $f_*$ with $\|f_*\|_\infty=K(C)$.

For the random projector $vv^\top$, the zonoid depth of $C$ equals $1/K(C)$ \citep{zonoid}. To pass from directions to projectors, replace $f(v)$ by $(f(v)+f(-v))/2$. Antipodal symmetry preserves both mass and moment, and averaging preserves the cap. The resulting even density is a function of $vv^\top$, whose two preimages are $v$ and $-v$.

\begin{lemma}[Halfspace comparison]\label{conf:halfspace}
If $0<p\le1$ and for each real symmetric matrix $D$ the event
$E_D=\{v:v^\top Dv\ge\operatorname{tr}(DC)\}$ has probability $\sigma(E_D)\ge p$, then $K(C)\le1/p$.
\end{lemma}
\begin{proof}
Let $\mathcal M_p$ be the image of the normalized densities bounded by $1/p$ under the moment map $f\mapsto\int vv^\top f\,d\sigma$. Convexity and weak-star compactness show that $\mathcal M_p$ is compact and convex. For any $D$, the conditional density $f_D=\mathbf1_{E_D}/\sigma(E_D)$ belongs to this class since $\sigma(E_D)\ge p$. Its moment $M_D$ satisfies $\operatorname{tr}(DM_D)\ge\operatorname{tr}(DC)$. If $C\notin\mathcal M_p$, strict separation in the space of symmetric matrices would supply a $D$ for which every $M\in\mathcal M_p$ has $\operatorname{tr}(DM)<\operatorname{tr}(DC)$. Taking $M=M_D$ contradicts that conclusion. Hence $C$ lies in the moment set.
\end{proof}

\begin{theorem}[Determinant law]\label{thm:det-cap}
Let $C$ be positive semidefinite of trace one and put $T=nC$. If $C$ is positive definite, then
\[
 \frac{1}{\sqrt{\det T}}
 \ \leq\ K(C)\ \leq\
 \frac{120}{\sqrt{\det T}}
 \left(1+\frac{\sqrt{240}\,\|T-I\|_F}{\sqrt{n(n+2)}}\right)^{n/2}.
\]
In particular,
\[
 0\leq \log K(C)+\tfrac12\log\det(nC)
 \leq \log 120+\sqrt{60}\,\|nC-I\|_F .
\]
If $C$ has a zero eigenvalue, $K(C)=+\infty$. If $C=I/n$, $K(C)=1$.
\end{theorem}

\begin{proof}
Take $g\sim N(0,I_n)$ and $y=g/\|g\|_2\sim\sigma$. The vector $T^{1/2}g\sim N(0,T)$ has direction
$\Phi(y)=T^{1/2}y/\sqrt{q(y)}$, where $q(y)=y^\top Ty$. Its directional density relative to $\sigma$ is
\[
 \rho(v)=(\det T)^{-1/2}w(v)^{-n/2},\qquad
 w(v)=v^\top T^{-1}v .
\]
Indeed, in the Gaussian density at $rv$ put $s=r\sqrt{w(v)}$ in the radial integral of $r^{n-1}\exp(-r^2w(v)/2)$. Relative to $T=I$, this yields the factor $w(v)^{-n/2}$. For any measurable $E$, integrate $\mathbf{1}_E/\rho$ against the law with density $\rho$ and use $w(\Phi(y))=1/q(y)$. This gives
\begin{equation}\label{eq:det-change}
 \sigma(E)=\sqrt{\det T}\int {\bf1}_E(\Phi(y))q(y)^{-n/2}\,d\sigma(y).
\end{equation}
Since $\int \rho\,d\sigma=1$, $\int w^{-n/2}\,d\sigma=\sqrt{\det T}$. The moment constraint under $f\,d\sigma$ gives $\mathbb E_f w=\operatorname{tr}(T^{-1}C)=1$. Convexity on $(0,\infty)$ now gives
\[
 1=(\mathbb E_f w)^{-n/2}\le\mathbb E_f[w^{-n/2}]
   \le\|f\|_\infty\int w^{-n/2}\,d\sigma=\|f\|_\infty\sqrt{\det T}.
\]
Taking the infimum over feasible densities gives the lower bound.

For the upper bound fix a symmetric test matrix $D$. Put
$D_0=D-\operatorname{tr}(DC)I$ and $J_0=T^{1/2}D_0T^{1/2}$.
Then $\operatorname{tr}J_0=n\operatorname{tr}(D_0C)=0$. We first show
\begin{equation}\label{eq:det-chaos-sign}
 \sigma\{y:y^\top J_0y\geq0\}\geq1/60 .
\end{equation}
It suffices to consider $J_0\ne0$. Diagonalize $J_0$ with eigenvalues
$a_i$ and write $X=g^\top J_0g=\sum_i a_i(g_i^2-1)$ in that basis.
For $X_i=a_i(g_i^2-1)$, the Gaussian moments give $\mathbb EX_i=0$,
$\mathbb EX_i^2=2a_i^2$ and $\mathbb EX_i^4=60a_i^4$. Put $V=\mathbb EX^2=2\sum_i a_i^2$. Independence and centering give
\[
 \mathbb E X^4
 =\sum_i\mathbb E X_i^4+6\sum_{i<j}\mathbb E X_i^2\,\mathbb E X_j^2
 =3V^2+48\sum_i a_i^4\le15V^2.
\]
H\"older's inequality gives $\mathbb EX^2\le(\mathbb E|X|)^{2/3}(\mathbb EX^4)^{1/3}$, whence
$\mathbb E|X|\geq V^{3/2}/(\mathbb E X^4)^{1/2}\geq\sqrt{V/15}$.
Since $\mathbb E X=0$, $\mathbb E X_+=\mathbb E|X|/2$.
Cauchy--Schwarz gives
$(\mathbb E X_+)^2\leq V\mathbb P(X>0)$. Normalizing $g$ preserves
the sign, proving \eqref{eq:det-chaos-sign}.

The random variable $q=y^\top Ty$ has mean $1$ and variance
$2\|T-I\|_F^2/[n(n+2)]$. Expand in an eigenbasis and use
$\mathbb Ey_i^4=3/[n(n+2)]$ and
$\mathbb Ey_i^2y_j^2=1/[n(n+2)]$ for $i\ne j$.
Write $d=\sqrt{240}\,\|T-I\|_F/\sqrt{n(n+2)}$.
Chebyshev's inequality gives $\sigma(q>1+d)\leq1/120$ when $d>0$.
The same bound holds when $d=0$, as $q=1$ identically.
Thus $y^\top J_0y\geq0$ and $q\leq1+d$ hold simultaneously with
probability at least $1/120$. On their intersection $\Phi(y)^\top D_0\Phi(y)=y^\top J_0y/q(y)\ge0$ and $q(y)^{-n/2}\ge(1+d)^{-n/2}$. Substituting in \eqref{eq:det-change} gives
\[
 \sigma(E_D)\geq p,\qquad E_D=\{v:v^\top D_0v\geq0\},\qquad
 p=\sqrt{\det T}(1+d)^{-n/2}/120.
\]
Lemma~\ref{conf:halfspace} gives $K(C)\le1/p$. Taking logarithms and using $\log(1+d)\le d$ gives the stated error bound.

Finally, if a unit vector $w$ lies in the kernel of $C$, feasibility would imply $\int(w^\top v)^2 f(v)\,d\sigma(v)=w^\top Cw=0$. The probability $f\,d\sigma$ would have to be supported on $\{v:w^\top v=0\}$, a set of $\sigma$-measure zero.
For $C=I/n$, $f=1$ attains the lower bound.
\end{proof}

For example, if $\lambda_{\max}(C)/\lambda_{\min}(C)\leq\kappa$
with fixed $\kappa$, then $\|nC-I\|_F=O_\kappa(\sqrt n)$. Thus
\[
 \frac{\log K(C)}n=
 -\frac1{2n}\sum_{i=1}^n\log\big(n\lambda_i(C)\big)+O_\kappa(n^{-1/2}).
\]
The Frobenius remainder stays bounded for many spiked spectra. A second
bound gives logarithmic accuracy even for arbitrarily ill-conditioned spectra.

\begin{theorem}[Sharp order of the uniform factor]\label{thm:uniform-det-cap}
For every $n\geq2$ and every positive definite, trace-one $C$, put
$T=nC$ and $M=\|T\|_{\rm op}$. Then
\[
 1\leq K(C)\sqrt{\det T}\leq
 U_n(M):=\min\{500n,\,800\sqrt{nM}\}.
\]
The worst possible factor over all such $C$ has order $n$.
If $M$ stays bounded, the upper bound has order $\sqrt n$,
also attained up to constants by fixed nontrivial two-level profiles.
More precisely, there exist absolute $c_0,C_0>0$ such that for every spectrum
\[
 c_0(1+\|T-I\|_F)\le K(C)\sqrt{\det T}\le C_0(1+\|T-I\|_F).
\]
The last bounds are proved in Appendix~\ref{sec:frobenius} (Theorem~\ref{thm:frobenius-cap} and Proposition~\ref{prop:frobenius-lower}).
\end{theorem}

\begin{proof}
Center a symmetric test matrix at $C$ by setting
$D_0=D-\operatorname{tr}(DC)I$. The separation argument above reduces
the problem to lower bounding $\mathbb P(g^\top D_0g\geq0)$ for a
standard Gaussian $g$, since its positive radius preserves the sign of the quadratic form, and its direction is uniform. If $D_0=0$ the event is certain.
For nonzero $D_0$ with $\operatorname{tr}D_0\geq0$, diagonalize $D_0$ and write $g^\top D_0g-\operatorname{tr}D_0=\sum_i\lambda_i(g_i^2-1)$. The centered fourth-moment calculation in the proof of \eqref{eq:det-chaos-sign} gives this sum a positive sign with probability at least $1/60$. On that event $g^\top D_0g>0$. Thus $\mathbb P(g^\top D_0g\ge0)\ge1/60\ge\sqrt{\det T}/U_n(M)$, since $\det T\le1$ and $U_n(M)\ge60$.

Now suppose $\operatorname{tr}D_0<0$. Since $\operatorname{tr}(D_0 C)=0$
and $C\succ0$, $D_0$ has a positive eigenvalue because a nonzero negative semidefinite $D_0$ would give $\operatorname{tr}(C^{1/2}D_0C^{1/2})<0$. For
$0\leq t<1/(2\lambda_{\max}(D_0))$ put $\Sigma_t=(I-2tD_0)^{-1}$.
The function $\operatorname{tr}(D_0 \Sigma_t)$ is strictly increasing from a
negative value to $+\infty$, as seen in an eigenbasis. Choose its unique
zero $t>0$. Indeed, for eigenvalues $\lambda_i$ of $D_0$ the function is $\sum_i\lambda_i/(1-2t\lambda_i)$ and its derivative is $\sum_i2\lambda_i^2/(1-2t\lambda_i)^2>0$. Write $L=\|\Sigma_t\|_{\rm op}$.
The identity $\Sigma_t-I=2tD_0\Sigma_t$ gives
$\operatorname{tr}\Sigma_t=n$. Its eigenvalues are positive and sum to $n$, so $1\le L<n$ and every eigenvalue of $\Sigma_t-I$ lies in $(-1,n-1)$.
Let $m=\sqrt{\det\Sigma_t}$. The positive matrix
$R_t=\Sigma_t^{-1/2}T\Sigma_t^{-1/2}$ has $\operatorname{tr}R_t=\operatorname{tr}[(I-2tD_0)T]=n$, using $\operatorname{tr}(D_0T)=0$, and hence determinant
at most one. For a unit eigenvector $e_L$ of $\Sigma_t$ with eigenvalue $L$, we have
$e_L^\top R_t e_L=e_L^\top T e_L/L\le M/L$. Thus $\lambda_{\min}(R_t)\le M/L$.
The remaining $n-1$ eigenvalues have sum at most $n$. By AM--GM their product is at most $(n/(n-1))^{n-1}\le e$, so $\det R_t\le eM/L$. Therefore
\[
  m/\sqrt{\det T}=(\det R_t)^{-1/2}
       \ge \max\{1,\sqrt{L/(eM)}\}.
\]

Write $\mathbb P_0$ for the standard Gaussian law of $g$ and $X=g^\top D_0g$.
Define the tilted law by $d\mathbb P_t/d\mathbb P_0=e^{tX}/m$. Completing the quadratic term in the density gives $-g^\top(I-2tD_0)g/2$, so $m=\mathbb E_0 e^{tX}=\sqrt{\det\Sigma_t}$ and $g\sim N(0,\Sigma_t)$ under $\mathbb P_t$. This choice makes $\mathbb E_t X=\operatorname{tr}(D_0\Sigma_t)=0$.
Represent $g=\Sigma_t^{1/2}w$ under the tilted law using a fresh $w\sim N(0,I_n)$. Since $\Sigma_t$ commutes with $D_0$, the identity $2tD_0\Sigma_t=\Sigma_t-I$ gives $tX=Z=w^\top(\Sigma_t-I)w/2$. Returning to the original measure multiplies the integrand by the inverse density, so
\begin{equation}\label{conf:tilt}
 p_D:=\mathbb P_0(X\ge0)
 =m\,\mathbb E_t[e^{-tX}{\bf1}_{\{X\ge0\}}]
 =m\,\mathbb E_w[e^{-Z}{\bf1}_{\{Z\ge0\}}].
\end{equation}
Let $J_0=\Sigma_t-I$, $u=w/\|w\|_2$, and $Q(u)=u^\top J_0u$. The squared radius $\|w\|_2^2\sim\chi_n^2$ is independent of $u$. On $0<Q(u)<1/n$, $Z=\|w\|_2^2Q(u)/2>0$ and integration over the radius gives $\mathbb E[e^{-Z}\mid u]=(1+Q(u))^{-n/2}\ge e^{-1/2}$. A lower bound on the probability of this band will therefore lower bound $p_D$. Since $J_0$ is nonzero and traceless, \eqref{eq:det-chaos-sign} applied to $J_0$ and $-J_0$ gives each strict sign of $Q$ probability at least $1/60$. The zero set of a nonzero quadratic form has Gaussian measure zero.

We estimate the measure $\delta=\sigma(0<Q<1/n)$ of the narrow band.
For each eigenvalue $\lambda\in[-1,L-1]$ of $J_0$, the product
$(\lambda+1)(\lambda-(L-1))$ is nonpositive. Expanding gives
$\lambda^2\le(L-2)\lambda+L-1$. Applying this in the eigenbasis yields
$J_0^2\preceq (L-2)J_0+(L-1)I$, where $\preceq$ means the difference of the right and left sides is positive semidefinite. In the band,
$u^\top J_0^2u\le L-1+(L-2)_+/n\le L$. Thus the Lipschitz function
$\varphi(u)=\max\{0,\min\{1,nQ(u)\}\}$ has spherical gradient of norm at
most $2n\sqrt L$ there, since $\nabla_{\mathbb S}Q(u)=2(J_0u-Q(u)u)$. The gradient vanishes almost everywhere outside the band.
For independent uniform $u,v$ we have
\[
 \mathbb E|\varphi(u)-\varphi(v)|
 \le \frac{\pi}{2\sqrt{n-1}}\,\mathbb E\|\nabla_{\mathbb S}\varphi(u)\|_2 .
\]
Indeed, write $v=u\cos\phi+v_\perp\sin\phi$, where $v_\perp$ is uniform on the
unit tangent sphere given $u$. The angle $\phi$ is independent of the
orthonormal pair $(u,v_\perp)$ and has mean $\pi/2$ by symmetry. Put $u_s=u\cos s+v_\perp\sin s$. Integrating along the arc bounds $|\varphi(v)-\varphi(u)|$ by
\[
 \int_0^\pi {\bf1}_{\{\phi\ge s\}}
   |\langle\nabla_{\mathbb S}\varphi(u_s),\dot u_s\rangle|\,ds .
\]
For fixed $s$, $(u_s,\dot u_s)$ has the original uniform frame law and is independent of $\phi$. Given $u_s$, the uniform unit tangent $\dot u_s$ has covariance $I/(n-1)$ on the tangent space. Cauchy--Schwarz therefore bounds the conditional mean of the absolute inner product by $\|\nabla_{\mathbb S}\varphi(u_s)\|_2/\sqrt{n-1}$. Taking expectations and integrating $\mathbb P(\phi\ge s)$ gives $\mathbb E\phi=\pi/2$, as required also for $n=2$.

The events $\varphi(u)=0$ and $\varphi(u)=1$ have probabilities at least $1/60$
and $1/60-\delta$, respectively. Keeping only the contributions of
these events gives a lower bound of $p_0\,\mathbb E \varphi(v)+p_1(1-\mathbb E \varphi(v))\ge 1/60-\delta$, where $p_i=\mathbb P(\varphi(u)=i)$. Substituting $\mathbb E\|\nabla_{\mathbb S}\varphi\|\le2n\sqrt L\,\delta$ in the rotation bound gives $1/60-\delta\le\pi n\sqrt{L/(n-1)}\,\delta$, and hence
\[
 \delta\ge [60(1+\pi n\sqrt{L/(n-1)})]^{-1}.
\]
Equation~\eqref{conf:tilt} gives $p_D\ge m\delta/\sqrt e$ for every test with negative original mean. We bound $\sqrt{\det T}/p_D$ uniformly in the test matrix.

First use $m\ge\sqrt{\det T}$ and $L\le n$, then use
$m/\sqrt{\det T}\ge\max\{1,\sqrt{L/(eM)}\}$. The two resulting bounds
for $\sqrt{\det T}$ divided by this probability are, respectively,
\[
 60\sqrt e(1+\pi n\sqrt{n/(n-1)})\le 500n,\qquad
 60\sqrt e(1+\pi n\sqrt{eM/(n-1)})\le 800\sqrt{nM}.
\]
Separation proves the upper bound on $K$ and Theorem~\ref{thm:det-cap}
gives the lower bound.

For sharpness take two equal upper eigenvalues of $C$ carrying total
mass $s\in(2/n,1)$ and equal remaining eigenvalues, with $n>2$.
In Proposition~\ref{prop:beta}, $U\sim{\rm Beta}(1,(n-2)/2)$, so
$\mathbb E[U\mid U\ge q]=q+(1-q)2/n$. Substituting
$1-q=(1-s)/(1-2/n)$ into $\mathbb P(U\ge q)=(1-q)^{(n-2)/2}$
gives $K(C)\sqrt{\det T}=ns/2$ exactly. The eigenvalues of $T$ are $ns/2$ twice and $n(1-s)/(n-2)$ in the other $n-2$ coordinates. Fixing $s$ proves the linear
order. For fixed $0<\theta<s<1$ and $k/n\to\theta$, the prefactor in
\eqref{eq:prefactor} yields order $\sqrt n$ while $M$ stays bounded.
\end{proof}

For comparison with statistical depth, let $Y=uu^\top$, $W=gg^\top$. For a random matrix $X$ define
\[
 I_X(M)=\sup_{H=H^\top}\{\operatorname{tr}(HM)-
              \log\mathbb E e^{\operatorname{tr}(HX)}\}.
\]
The halfspace depth $d_{\rm H}(C)$ is $\inf_H\mathbb P\{\operatorname{tr}[H(Y-C)]\ge0\}$ over symmetric $H$. For $T=nC\succ0$, it satisfies
\begin{equation}\label{eq:depth-determinant}
 I_W(T)\le I_Y(C)\le\log K(C)\le-\log d_{\rm H}(C)
       \le I_W(T)+\log U_n(M),\qquad M=\|T\|_{\rm op}.
\end{equation}
For $W$, the Gaussian log moment generating function is $-\frac12\log\det(I-2H)$ on $I-2H\succ0$ and infinite outside. In that domain the objective has derivative $\operatorname{tr}(L[T-(I-2H)^{-1}])$ in a symmetric direction $L$. Its stationary point $H=(I-T^{-1})/2$ belongs to the domain and maximizes the concave objective. Substitution gives $I_W(T)=(\operatorname{tr}T-n-\log\det T)/2=-\frac12\log\det T$. Conditional Jensen over the independent squared radius of mean $n$ gives $\mathbb E e^{g^\top Hg}\ge\mathbb E e^{n u^\top H u}$. Use $T=nC$ and substitute $nH$ in the supremum for $Y$ to obtain $I_W(T)\le I_Y(C)$. For any bounded feasible $f$, apply Jensen's inequality to $\log$ under $f\,d\sigma$, with $0\log0=0$ in the entropy. This yields
\[
 \operatorname{tr}(HC)-\int f\log f\,d\sigma
 =\mathbb E_f\log\frac{e^{u^\top Hu}}{f(u)}
 \le\log\int_{\{f>0\}}e^{u^\top Hu}\,d\sigma
 \le\log\mathbb E e^{u^\top Hu}.
\]
Taking the supremum over symmetric $H$ gives $I_Y(C)\le\int f\log f\,d\sigma$. Since $f\le\|f\|_\infty$ and $\int f\,d\sigma=1$, the entropy is at most $\log\|f\|_\infty$. Taking the infimum over $f$ gives $I_Y(C)\le\log K(C)$. Lemma~\ref{conf:halfspace} gives the next inequality, and the bound on every $p_D$ above gives the last.

The geometric mean of the normalized eigenvalues thus gives a
spectrum-specific exponential rate. A condition number alone specifies
a different quantity, the worst rate among spectra sharing a range.

\subsection{The Worst Spectrum at a Prescribed Condition Number}
\begin{theorem}[Worst-case exponent]\label{thm:specht-cap}
For fixed $\kappa>1$ let
$K_n^{\max}(\kappa)=\sup\{K(C):\operatorname{tr}C=1,\ C\succ0,\
\lambda_{\max}(C)\leq\kappa\lambda_{\min}(C)\}$.
Put $\theta_*=(\log\kappa)^{-1}-(\kappa-1)^{-1}$ and
$s_*=\kappa\theta_*/[1+(\kappa-1)\theta_*]$. Then
\[
 K_n^{\max}(\kappa)\sim
 \frac{(s_*-\theta_*)\sqrt{\pi n}}{e\sqrt{\theta_*(1-\theta_*)}}\,
 S(\kappa)^{n/2},\qquad S(\kappa)=
 \frac{\kappa-1}{e\log\kappa}\,\kappa^{1/(\kappa-1)} .
\]
Here $S$ is the classical reverse arithmetic--geometric mean constant
of Specht \citep{specht}. At $\kappa=1$ the cost is identically one.
\end{theorem}

\begin{proof}
Orthogonal invariance of $\sigma$ makes $K$ depend only on the
eigenvalues. The convex combination of two feasible densities has the mixed moment and an essential upper bound no larger than the same convex combination of the two bounds. Thus $K$ is convex in $C$.
The eigenvalue polytope has constraints $\sum_i\lambda_i=1$ and
$\lambda_i\leq\kappa\lambda_j$ for all $i,j$. At an extreme point the
ratio must equal $\kappa$, since strict inequalities leave a relative
neighborhood feasible. If any component lies strictly between the minimum
and maximum, vary the minimum group by $t$, the maximum group by
$\kappa t$ and one interior component by the amount needed to keep the
sum fixed. With $r$ minimal entries and $\ell$ maximal entries this last change is $-(r+\kappa\ell)t$. Both signs of small $t$ are feasible. Thus every extreme
point has two levels, $\kappa$ and $1$ up to a common factor. Conversely
these points are extreme, since their tight ratio equalities and the trace
equation determine them. A convex function on this polytope is bounded
by its largest value at the vertices.

At a vertex let $k\in\{1,\ldots,n-1\}$ be the multiplicity of the
upper level, $\theta=k/n$, and $s=\kappa\theta/[1+(\kappa-1)\theta]$ its
total mass. The two eigenvalues of $nC$ are $s/\theta$ and $(1-s)/(1-\theta)$, with multiplicities $n\theta$ and $n(1-\theta)$. Taking the logarithm of their product with these multiplicities gives
\[
 -\tfrac1n\log\det(nC)=
 \theta\log\frac{\theta}{s}+(1-\theta)\log\frac{1-\theta}{1-s}
 =D(\theta\Vert s).
\]
Theorem~\ref{thm:uniform-det-cap} then gives $\log K(C)=\tfrac n2D(\theta\Vert s)+O(\log n)$ uniformly over $k$.

Finally $D(\theta\Vert s)=\log(1+(\kappa-1)\theta)-\theta\log\kappa$.
Its derivative $(\kappa-1)/(1+(\kappa-1)\theta)-\log\kappa$ is strictly decreasing and vanishes at the unique maximum
$\theta_*=(\log\kappa)^{-1}-(\kappa-1)^{-1}\in(0,1)$, with value
$\log S(\kappa)$. For fixed $\kappa$ a nearest mesh point $k/n$ loses
$O_\kappa(n^{-2})$ in this value by Taylor's theorem for large $n$.
Outside any fixed neighborhood of $\theta_*$ the exponent $D(\theta\Vert s(\theta))$ loses at least some $\delta>0$ from its strict maximum. The resulting loss $n\delta/2$ exceeds the uniform $O(\log n)$ error for all large $n$. Comparing with the nearest mesh point therefore shows that every maximizing vertex has $k/n\to\theta_*$. To recover the multiplicative constant, apply Proposition~\ref{prop:beta} at both the nearest and the maximizing vertices: $\theta_*\in(0,1)$ and $\theta_*<s_*<1$ ensure both multiplicities diverge, while $s-\theta$ and $1-s$ stay bounded away from zero. In \eqref{eq:prefactor} the prefactor divided by $\sqrt n$ tends to the same positive constant on both sequences. The nearest mesh point gives the lower bound since its exponential term is $(1+o(1))S(\kappa)^{n/2}$. At the maximizing vertex the term is at most $S(\kappa)^{n/2}$, giving the matching upper bound.
\end{proof}

A single spike with eigenvalue ratio $\kappa$ has $\|nC-I\|_F=O_\kappa(1)$
and $-\log\det(nC)=O_\kappa(1)$, so its cost stays bounded as $n$ grows.
A two-level spectrum with a fraction approaching $\theta_*$ of upper eigenvalues
and the same ratio has exponentially growing cost. This distinction persists
even when both spectra have full rank and the same condition number.
At $\kappa=2$ and $n=100$ the worst-case leading log term
$(n/2)\log S(2)$ is $2.983$, whereas the uniform additive error bound
$\log(500n)$ is $10.820$. The leading term first exceeds that
conservative bound at $n=411$. The beta equation in
Proposition~\ref{prop:beta} gives finite-width values directly.

\section{The Two-Sided Frobenius Correction for Each Spectrum}\label{sec:frobenius}
After the determinant accounts for the leading cost, how large a factor remains in \(K(C)\)? Theorem~\ref{thm:frobenius-cap} bounds \(K(C)\sqrt{\det T}\) from above, and Proposition~\ref{prop:frobenius-lower} bounds it from below. Together these prove \eqref{conf:sharp}. In particular, the correction stays bounded independently of width whenever \(\sum_i(\lambda_i(T)-1)^2\) stays bounded.

\begin{theorem}\label{thm:frobenius-cap}
For every positive definite $C$ of trace one, with $T=nC$ and $n\ge2$,
\[
 1\le K(C)\sqrt{\det T}\le 2\cdot10^5\,(1+\|T-I\|_F).
\]
\end{theorem}
For a two-level spectrum with upper block of relative size $\theta$ and
mass $s$, the norm in this bound equals
$\sqrt n(s-\theta)/\sqrt{\theta(1-\theta)}$, by substituting the two normalized eigenvalues $s/\theta$ and $(1-s)/(1-\theta)$ with multiplicities $n\theta$ and $n(1-\theta)$.
In particular, for fixed $0<\theta<s<1$, the asymptotic equivalent in
Proposition~\ref{prop:beta} is exactly
$K(C)\sqrt{\det T}\sim(\sqrt{\pi}/e)\|T-I\|_F$.

\begin{proof}
The lower bound is Theorem~\ref{thm:det-cap}. For the upper bound we first
control the tilted expectation in \eqref{conf:tilt} on $0<Z<1$, where $e^{-Z}\ge e^{-1}$. We establish a one-sided estimate for any nonzero traceless symmetric matrix
$D$. Write $g\sim N(0,I_n)$, $d=\|D\|_F$,
$Z=g^\top Dg/2$, and $\delta=\mathbb P(0<Z<1)$. We claim
\begin{equation}\label{eq:frob-small-slab}
 \delta\ge \frac{0.014}{1+51d}.
\end{equation}
Let $[a]_0^1=\max(0,\min(a,1))$, $r=\|Dg\|_2/d$, and
$\alpha=[2-r/20]_0^1$, $\beta=[Z]_0^1$, and $\varphi(g)=\alpha\beta$. The second factor changes only in the slab $0<Z<1$. The first equals one for $r\le20$ and vanishes for $r\ge40$. Since $\mathbb E r^2=1$,
$\mathbb P(r>20)\le1/400$. Both strict signs of $Z$ have probability at
least $1/60$ by the fourth-moment calculation in
Theorem~\ref{thm:det-cap}. The event $Z\le0$ forces $\varphi=0$, and $Z\ge1,r\le20$ forces $\varphi=1$. Consequently $\mathbb P(\varphi=0)\ge1/60$ and
$\mathbb P(\varphi=1)\ge1/60-\delta-1/400$. For an independent copy $g'$,
conditioning on $\varphi(g)=u$ bounds $\mathbb E[|u-\varphi(g')|]$ below by
$u\mathbb P(\varphi=0)+(1-u)\mathbb P(\varphi=1)$. It follows that
$\mathbb E|\varphi(g)-\varphi(g')|\ge1/60-\delta-1/400$.

For Lipschitz $\varphi$, the Gaussian rotation inequality follows by integrating along $g_t=g\cos t+g'\sin t$, $0\le t\le\pi/2$ \citep[Theorem~1.1]{bobkovvolzone}. The pair $(g_t,\dot g_t)$ is an orthogonal image of $(g,g')$ and thus consists of independent standard Gaussians. Given $g_t$, the conditional mean of $|\nabla \varphi(g_t)^\top\dot g_t|$ is $\sqrt{2/\pi}\|\nabla \varphi(g_t)\|_2$. Integration over an interval of length $\pi/2$ yields $\mathbb E|\varphi(g)-\varphi(g')|\le\sqrt{\pi/2}\,\mathbb E\|\nabla \varphi(g)\|_2$.

Because $\|D\|_{\rm op}\le d$, $r$ is $1$-Lipschitz and
$\|\nabla\alpha\|_2\le\mathbf1_{\{20<r<40\}}/20$ almost everywhere.
Where $\alpha>0$, $\|\nabla Z\|_2=\|Dg\|_2\le40d$, so
$\alpha\|\nabla\beta\|_2\le40d\,\mathbf1_{\{0<Z<1\}}$ almost everywhere.
The product is Lipschitz (the local gradient bound is uniform), and the product rule with $0\le\beta\le1$ gives
\[
 \mathbb E\|\nabla \varphi\|_2\le
 40d\,\delta+\tfrac1{20}\mathbb P(r>20)
 \le40d\,\delta+20^{-3}.
\]
Consequently
\[
 \frac{1}{60}-\delta-\frac{1}{400}\le\sqrt{\pi/2}\,(40d\,\delta+20^{-3}).
\]
The constant left after moving $\sqrt{\pi/2}\,20^{-3}$ is greater than $0.014$ and $40\sqrt{\pi/2}<51$, giving \eqref{eq:frob-small-slab}.

Fix a symmetric test matrix $D_{\rm test}$, let $D_0=D_{\rm test}-\operatorname{tr}(D_{\rm test}C)I$ and put $p_D=\mathbb P(g^\top D_0g\ge0)$. If $D_0=0$ then $p_D=1$. For $\operatorname{tr}D_0\ge0$ the fourth-moment bound gives $p_D\ge1/60$. In the remaining case choose the tilt $\Sigma_t=(I-2tD_0)^{-1}$ with $\operatorname{tr}(D_0\Sigma_t)=0$ as in Theorem~\ref{thm:uniform-det-cap}. Then $\Sigma_t-I=2tD_0\Sigma_t$ is nonzero and traceless, so it qualifies for the slab estimate. Put $R_t=\Sigma_t^{-1/2}T\Sigma_t^{-1/2}$. We have
\[
 \operatorname{tr}\Sigma_t=\operatorname{tr}T=n,\qquad
 \operatorname{tr}R_t=\operatorname{tr}[(I-2tD_0)T]=n
\]
since $\operatorname{tr}(D_0T)=0$. In \eqref{conf:tilt} take $Z=g^\top(\Sigma_t-I)g/2$ with a fresh standard Gaussian. Restricting this expectation to the slab yields
\[
 p_D\ge e^{-1}\sqrt{\det\Sigma_t}\,\delta,\qquad
 \frac{\sqrt{\det T}}{p_D}\le\frac{e}{0.014}\sqrt{\det R_t}\,
      (1+51\|\Sigma_t-I\|_F),
\]
where $\det T/\det\Sigma_t=\det R_t$.

To express this in terms of $T$, take an eigenbasis of $\Sigma_t$ and let $x_i$
be its eigenvalues, $y_i$ the diagonal entries of $T$ and $v_i=y_i/x_i$
those of $R_t$. Here $T$ and $R_t$ may have off-diagonal entries in the chosen basis. Hadamard's inequality gives $\det R_t\le\prod_i v_i$, while $\sum_i(y_i-1)^2\le\sum_{ij}(T_{ij}-\delta_{ij})^2=\|T-I\|_F^2$. Each of $x,y,v$ sums to $n$. With $a_i=x_i-1$,
$b_i=v_i-1$ and $c_i=y_i-1$, we have $c_i=a_i+b_i+a_i b_i$ and
$\sum_i a_i b_i=0$. Since $a_i=(c_i-b_i)/v_i$, on the indices with $v_i\ge1/2$,
$|a_i|\le2(|c_i|+|b_i|)$. If $v_i<1/2$ and $a_i<0$, then
$|a_i|\le1\le2|b_i|$. The remaining indices satisfy
\[
 \sum_{\{v_i<1/2,\,a_i\ge0\}} a_i
 \le 2\sum_{\{a_i b_i<0\}}|a_i b_i|
 =2\sum_{\{a_i b_i>0\}}a_i b_i
 \le2\|c\|_2\|b\|_2 .
\]
For the last inequality, $a_i b_i>0$ implies $|a_i|\le |c_i|$,
since $x_i$ and $v_i$ are on the same side of one and $y_i=x_i v_i$.
On the remaining set $E=\{i:v_i<1/2,\ a_i\ge0\}$, the displayed estimate gives $\|a_E\|_2\le\sum_{i\in E}a_i\le2\|c\|_2\|b\|_2$. Combining the three $\ell^2$ bounds gives
$\|a\|_2\le2\|c\|_2+4\|b\|_2+2\|c\|_2\|b\|_2$.

Write $J=\sum_i(v_i-1-\log v_i)$. Since $\sum_i(v_i-1)=0$, $J=-\sum_i\log v_i$. Hadamard's inequality gives
$\sqrt{\det R_t}\le\prod_i\sqrt{v_i}=e^{-J/2}$. Integrating
$1-1/v$ from $1$ to $v$ yields
$v-1-\log v\ge(v-1)^2/(2\max(1,v))$. Hence
$\|b\|_2^2\le2(1+\|b\|_2)J$ and
$\|b\|_2\le2J+\sqrt{2J}$. Maximizing the two terms separately gives
$e^{-J/2}\|b\|_2\le4/e+\sqrt{2/e}<3$. We obtain
\[
 \sqrt{\det R_t}\,\|\Sigma_t-I\|_F
 =\sqrt{\det R_t}\,\|a\|_2
 \le 12+8\|c\|_2\le12+8\|T-I\|_F .
\]
Write $F=\|T-I\|_F$. In the negative-trace case, substituting the last estimate and $\det R_t\le1$ into the bound on $\sqrt{\det T}/p_D$ gives
\[
 \frac{\sqrt{\det T}}{p_D}\le\frac{e}{0.014}\,[1+51(12+8F)]
 \le 2\cdot10^5(1+F).
\]
In the nonnegative-trace case $p_D\ge1/60$ and $\det T\le1$ give a ratio at most $60$. The bound thus holds for every test, and Lemma~\ref{conf:halfspace} gives the same bound on $K(C)\sqrt{\det T}$.
\end{proof}

\paragraph{A matching lower bound for each spectrum.}
We first record the one-sided interval estimate that enters the
lower bound. Global density bounds and L\'evy concentration bounds control interval probabilities over all locations \citep{quadraticdensity,levyconcentration}. The test in Proposition~\ref{prop:frobenius-lower} uses only intervals with left endpoint $x\ge-1$. For a single large coefficient $a>0$, the unit interval starting at the singular endpoint $-a$ has probability $\mathbb P(|g|\le a^{-1/2})$ of order $a^{-1/2}$. An interval starting at $x\ge-1$ is at distance at least $a-1$ from that endpoint. With a residual sum $S$, the conditional endpoint becomes $S-a$. On $S\le a/2$ the distance is still at least $a/2-1$, giving the $a^{-1}$ bound from the conditional density. The proof controls $S>a/2$ by exponential tilting.

\begin{lemma}\label{lem:one-sided-quadratic}
Let $g_i$ be independent $N(0,1)$ variables. Let $a_i\ge-1$, $F=(\sum_i a_i^2)^{1/2}$ and
$Z=\sum_i a_i(g_i^2-1)$. Then
\[
 \sup_{x\ge-1}\mathbb P(x\le Z\le x+1)
 \ \le\ \frac{2048}{1+F}.
\]
\end{lemma}
\begin{proof}
Under any of the laws below, a density bounded by $L$ assigns probability at most $L$ to each unit interval. We first prove a density estimate for sums with well-spread coefficients.
If $V^2=\sum_i b_i^2>0$ and $\max_i |b_i|\le V/4$, partition
the indices into three groups, each with squared sum at least $V^2/4$.
To do so, accumulate terms until the squared sum first reaches $V^2/4$,
twice, and use the remaining indices as the last group.
Each of the first two groups consumes at most $5V^2/16$.
For the characteristic function $\phi(t)=\mathbb E e^{it\sum_i b_i(g_i^2-1)}$ the modulus is $\prod_i(1+4b_i^2t^2)^{-1/4}$. For each of the three groups $I$, $\prod_{i\in I}(1+4b_i^2t^2)\ge1+4t^2\sum_{i\in I}b_i^2\ge1+V^2t^2$. Hence $|\phi(t)|\le(1+V^2t^2)^{-3/4}$. The Fourier inversion density bound is $(2\pi)^{-1}\int|\phi(t)|\,dt\le1/V$, since $\int_{\mathbb R}(1+t^2)^{-3/4}\,dt\le 2+4<2\pi$. This also bounds the mass of any unit interval.

Translations preserve this density bound.

For $F\le256$ the lemma follows from the probability bound one.
For $F>256$, the preceding estimate applies if $\max_i|a_i|\le F/4$.
Otherwise the largest coefficient in absolute value is positive since
all negative coefficients are at least $-1$. Relabel so that
$a=a_1>F/4>64$. Let $S=\sum_{i>1}a_i(g_i^2-1)$ and
$r^2=\sum_{i>1}a_i^2$, so $r<4a$. We split according to the size of $S$. When $S\le a/2$ the singular endpoint of the leading summand stays below the interval of interest. When $S>a/2$ an exponential tilt of the residual supplies a small weight. For the first case, fix the value of $S\le a/2$. Under this conditioning,
the density of $Z$ at any $z\ge-1$ is bounded by
\[
 [2\pi a(z+a-S)]^{-1/2}\le 1/a ,
\]
since $z+a-S\ge a/2-1\ge a/4$. This treats $r=0$ as well.

For $r>0$, tilt the law of $S$ alone by the density
$e^{tS}/\mathbb E e^{tS}$ with $t=1/(8r)$, leaving $g_1$ independent
and standard. Independence and the Gaussian square moment generating function give
\[
 \log\mathbb E e^{tS}
 =\tfrac12\sum_{i>1}[-\log(1-2ta_i)-2ta_i]
 \le 2t^2r^2=1/32.
\]
Here $|2t a_i|\le1/4$ and $-\log(1-u)-u\le u^2$ for $|u|\le1/4$. Thus $\mathbb E e^{tS}\le e$. The reverse density $(\mathbb E e^{tS})e^{-tS}$ converts probabilities back to the original law. For any interval $I$ of
length one,
\[
 \mathbb P(S>a/2,\ Z\in I)
 \le e\,e^{-a/(16r)}\,\mathbb P_t(Z\in I),
\]
where $\mathbb P_t$ is the tilted product law. If $r\le16$, this is
at most $e\,e^{-a/256}\le256/a$.

If $r>16$, the residual coordinates under the tilt are independent, and $g_i=\xi_i/\sqrt{1-2ta_i}$ in law, with independent $\xi_i\sim N(0,1)$. Thus $S$ has the law of $\sum_{i>1}b_i(\xi_i^2-1)+\sum_{i>1}(b_i-a_i)$, where $b_i=a_i/(1-2ta_i)$. The deterministic shift only translates the interval.
Write $r'=(\sum_{i>1}b_i^2)^{1/2}\ge4r/5$. If some $a_i\ge r/8$
for $i>1$, then $b_i\ge r/8$. Under the tilted law the density of $ag_1^2+b_i\xi_i^2$, with independent standard $g_1,\xi_i$, is everywhere bounded
by $1/(2\sqrt{ab_i})$. Indeed, bounding the exponential factors by
one in the two densities leaves the integral
$\int_0^y [z(y-z)]^{-1/2}\,dz=\pi$ for $y>0$.
Adding the independent remaining summands preserves the bound.
It is at most $3/r$ since $r<4a$.
In the other case every residual coefficient satisfies $|a_i|\le r/8$,
using $a_i\ge-1$ and $r>16$ for the negative ones. Therefore
$\max_{i>1}|b_i|\le r/6\le r'/4$, and the three-group density bound
on the residual gives $\mathbb P_t(Z\in I)\le 1/r'\le 3/r$.
In both cases the contribution on $S>a/2$ is at most
$3e\,e^{-a/(16r)}/r\le48/a$, using $y e^{-y}\le1/e$.
Together the estimates bound every interval starting at $x\ge-1$ by
$257/a<1028/F\le2048/(1+F)$ in the large-coefficient case.
\end{proof}

\begin{proposition}\label{prop:frobenius-lower}
For $T=nC\succ0$ with $\operatorname{tr}T=n$,
\[
 K(C)\sqrt{\det T}\ \ge\
 \frac{(1-e^{-1/2})^2}{2048\sqrt e}\,(1+\|T-I\|_F).
\]
\end{proposition}
\begin{proof}
Write $\mathbb E_0$ for expectation under $g\sim N(0,I_n)$ and $\mathbb E_f$ for the law with density $f(u)$ relative to that Gaussian, where $u=g/\|g\|_2$ and $f$ is feasible. Independence of $u$ and $\|g\|_2$ under the original law gives
\[
 \mathbb E_f[gg^\top]=\mathbb E_0\|g\|_2^2\,\mathbb E_0[f(u)uu^\top]=nC=T.
\]
Use the test $\psi(g)=1+g^\top(I-T^{-1})g$. Its quadratic term has mean $\operatorname{tr}[(I-T^{-1})T]=\operatorname{tr}(T-I)=0$ under every feasible weighting. The constant gives $\mathbb E_f\psi=1$, while the same matrix yields $T-I$ under the Gaussian substitution below. Thus
\[
 1\le\mathbb E_f \psi_+\le\|f\|_\infty\mathbb E_0 \psi_+ .
\]
In the Gaussian integral for the last expectation substitute $g=T^{1/2}w$. The Jacobian is $\sqrt{\det T}$ and $\|g\|_2^2=\|w\|_2^2+Z$, where $Z=w^\top(T-I)w$. Since $\psi(T^{1/2}w)=1+Z$, this gives
\[
 \mathbb E_0 \psi(g)_+=\sqrt{\det T}\,
 \mathbb E_{w\sim N(0,I_n)}[(1+Z)_+e^{-Z/2}].
\]

In an orthonormal eigenbasis of $T$, the coordinates $w_i$ remain independent standard normals by rotation invariance. Put $a_i=\lambda_i(T)-1$. Since $\operatorname{tr}T=n$, $\sum_i a_i=0$ and $Z=\sum_i a_i(w_i^2-1)$. Positive definiteness gives $a_i>-1$, and $(\sum_i a_i^2)^{1/2}=\|T-I\|_F$. On the disjoint intervals $j-1\le Z<j$, $j=0,1,\ldots$, the integrand
is at most $(j+1)e^{(1-j)/2}$, and it vanishes for $Z<-1$.
Apply Lemma~\ref{lem:one-sided-quadratic} and sum the geometric
series $\sum_{j\ge0}(j+1)e^{-j/2}=(1-e^{-1/2})^{-2}$. This gives
\[
 \mathbb E[(1+Z)_+e^{-Z/2}]
 \le \frac{2048\sqrt e}{(1-e^{-1/2})^2(1+\|T-I\|_F)}.
\]
It follows that
\[
 1\le\|f\|_\infty\sqrt{\det T}\,
 \frac{2048\sqrt e}{(1-e^{-1/2})^2(1+\|T-I\|_F)} .
\]
Dividing by the positive fraction and taking the infimum over feasible $f$ proves the claim.
\end{proof}

Combining Theorem~\ref{thm:frobenius-cap} and Proposition~\ref{prop:frobenius-lower} proves \eqref{conf:sharp}. Returning to the zero-left-slope model gives $C=S/\tau$ and $K(C)=B_{\min}/\tau$. Taking logarithms shows that
\[
 \log(B_{\min}/\tau)=-\tfrac12\log\det T+\log(1+\|T-I\|_F)+O(1).
\]
The additive error lies between $\log c_0$ and $\log C_0$, independent of both width and spectrum.

\section{Exact Critical Slope for a Two-Level Gradient Spectrum}
\label{sec:beta}
\begin{figure}[htbp]
\centering
\includegraphics[width=.93\linewidth,trim=0 0 0 38,clip]{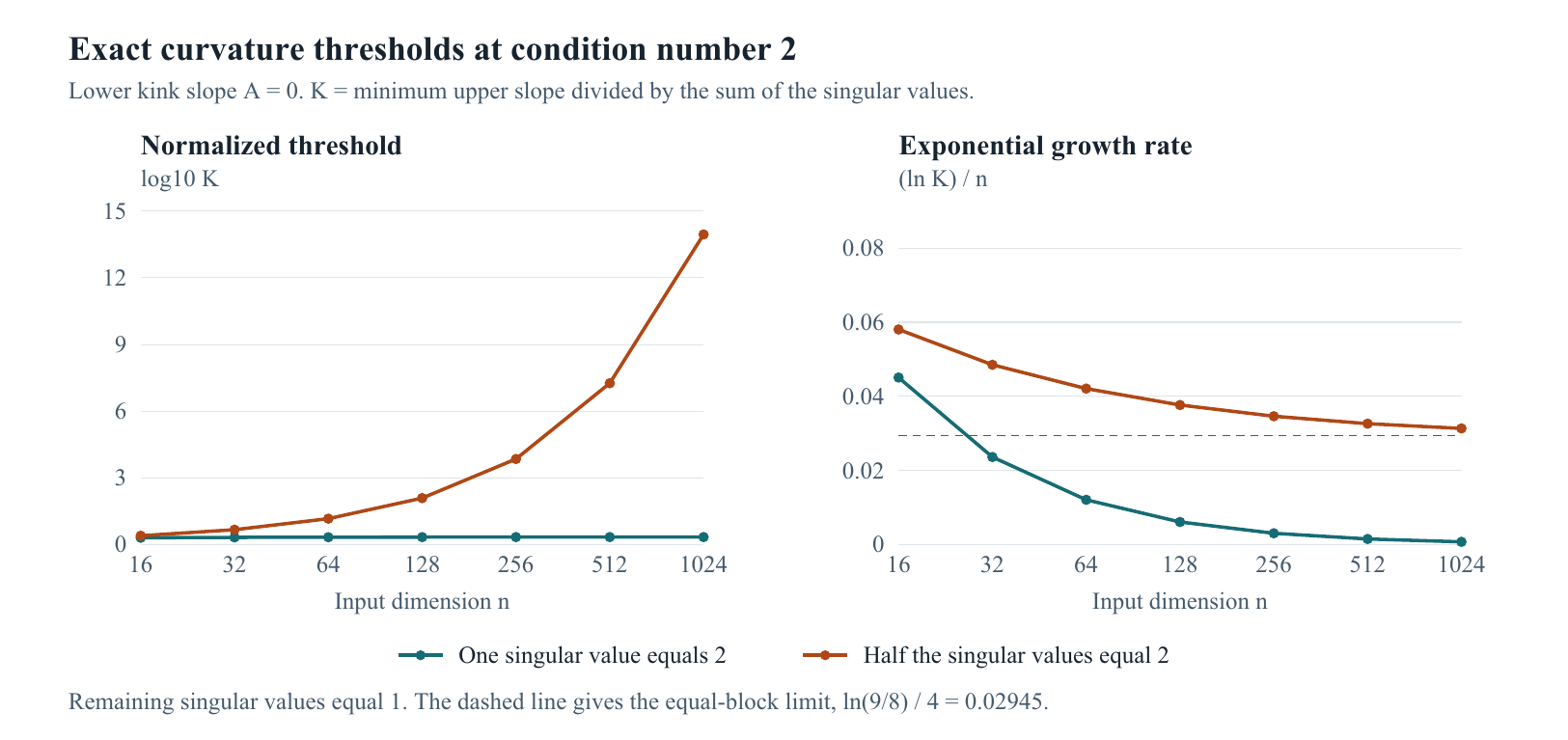}
\caption{Changing the multiplicity of the upper singular value at fixed condition number two. Left, $\log_{10}K$, where $K=B_{\min}/\tau$. Right, $(\ln K)/n$.
The dashed line marks the equal-block rate $\tfrac14\ln(9/8)$ for $\theta=1/2$, $s=2/3$. With one upper singular value $K$ stays bounded, while with half the singular values at the upper level it grows exponentially. Both curves use the beta equation in Proposition~\ref{prop:beta}.}
\label{fig:threshold}
\end{figure}

For two distinct gradient singular values, the density problem of Section~\ref{sec:criterion} reduces to a scalar constraint. Proposition~\ref{prop:beta} computes the exact threshold from a beta tail and gives its asymptotic prefactor when both block dimensions grow and the target share stays separated from the dimension share and from one. Figure~\ref{fig:threshold} compares the exact thresholds for one upper singular value and for an upper block occupying half the dimensions.

Let $S/\tau$ have two eigenvalue levels with ratio $\kappa>1$. The upper
eigenspace has dimension $1\le k<n$ and dimension fraction $\theta=k/n$.
Its share of the trace is $s=\kappa k/(\kappa k+n-k)$, so $\theta<s<1$.
If $\Pi$ projects onto this space and $\zeta=g/\|g\|$, its angular
energy $U=\|\Pi\zeta\|^2=X/(X+Y)$ has the $\mathrm{Beta}(k/2,(n-k)/2)$ law.
Here $X$ and $Y$ are independent
sums of squares of respectively $k$ and $n-k$ standard normal variables.
A sum of $m$ such squares has the $\chi_m^2$ law. The target block
mass is $s=\operatorname{tr}(\Pi S/\tau)$. Multiplying
$\mathbb E[f\zeta\zeta^\top]=S/\tau$ by $\Pi$ and taking the trace gives
$\mathbb E[fU]=s$.
The density of $U$ is
$u^{a-1}(1-u)^{d-1}/\mathrm B(a,d)$ for $0<u<1$, with $a=k/2$,
$d=(n-k)/2$ and $\mathrm B(a,d)=\int_0^1v^{a-1}(1-v)^{d-1}\,dv$.
\begin{proposition}[Two-level threshold]\label{prop:beta}
There is a unique $q\in(0,s)$ with $\mathbb E[U\mid U\ge q]=s$, and
\[
 K(S/\tau)=\frac1{\mathbb P(U\ge q)} .
\]
Suppose $k,n-k\to\infty$ and, for some $\varepsilon>0$, $\theta+\varepsilon\le s\le1-\varepsilon$. The parameters $s$ and $\theta$ may vary with $n$ under these separation conditions. Writing $I_\theta(s)=\tfrac12[\theta\log(\theta/s)+(1-\theta)\log((1-\theta)/(1-s))]$, the normalized critical slope $B_{\min}/\tau=K(S/\tau)$ for $A=0$ satisfies
\begin{equation}\label{eq:prefactor}
 K(S/\tau)=(1+o(1))\,
 \frac{(s-\theta)\sqrt{\pi n}}{e\sqrt{\theta(1-\theta)}}\,
 \exp\{n I_\theta(s)\}.
\end{equation}
\end{proposition}
\begin{proof}
For a trial cap $K\ge1$, define $q_K\in[0,1)$ by
$\Pr(U\ge q_K)=1/K$. The density $f_K=K\mathbf1_{\{U\ge q_K\}}$ has
mean one. Any other density $f\le K$ obeys $(f-f_K)(U-q_K)\le0$
pointwise. Integrating cancels the $q_K$ term because both densities
have mean one, giving $\mathbb E[fU]\le\mathbb E[f_K U]=\mathbb E[U\mid U\ge q_K]$.
Write $m(t)=\mathbb E[U\mid U\ge t]$ and $p_n$ for the beta density.
For $0<t<1$, $m(t)>t$ and
$m'(t)=p_n(t)(m(t)-t)/\Pr(U\ge t)>0$. Since $m(0)=\theta$ and
$m(1^-)=1$, the equation $m(q)=s$ has a unique solution and $q<s$.
If $K<1/\Pr(U\ge q)$, then $q_K<q$, so no density
under this cap can realize $s$. At equality $q_K=q$: within each block
the law weighted by $f_K$ is rotation invariant and reflections annul
cross moments. The block traces are $s$ and $1-s$, and the diagonal
entries are $s/k$ and $(1-s)/(n-k)$, giving the required matrix.

The beta density at $t$ is, by Stirling's formula,
\[
 (1+O(k^{-1}+(n-k)^{-1}))\frac{\sqrt{n\theta(1-\theta)}}{2\sqrt\pi\, t(1-t)}
 e^{-nI_\theta(t)} .
\]
On $s-\varepsilon/2\le t\le s$, both $t$ and $1-t$ stay bounded away from zero and $t-\theta\ge\varepsilon/2$. Thus $I'_\theta(t)=(t-\theta)/(2t(1-t))$ has a positive lower bound depending only on $\varepsilon$. For fixed $L$ and $0\le z\le L$, Taylor expansion gives
\[
 \log\frac{p_n(t+z/n)}{p_n(t)}
 =-zI'_\theta(t)+O_{\varepsilon,L}(n^{-1}).
\]
Since $k,n-k\to\infty$, the coefficients of $\log t$ and $\log(1-t)$
in $\log p_n$ are eventually nonnegative. Thus $\log p_n$ is concave,
and $(\log p_n)'(t)=-nI_\theta'(t)+O_\varepsilon(1)\le-c_\varepsilon n$
on this interval. Its tangent bounds the density ratio by $e^{-c_\varepsilon z}$ throughout the remaining support. This bound is integrable both alone and after multiplication by $z$. Set $D_n(z)=p_n(t+z/n)/p_n(t)$ and extend it by zero outside the support.
The two integrals that determine the cutoff are
\[
 \mathbb P(U\ge t)=\frac{p_n(t)}n\int_0^\infty D_n(z)\,dz,\qquad
 n\,\mathbb E[U-t\mid U\ge t]=
 \frac{\int_0^\infty zD_n(z)\,dz}{\int_0^\infty D_n(z)\,dz}.
\]
On $[0,L]$ use the uniform Taylor estimate above. For $z>L$, both
integral errors are bounded by a constant multiple of
$\int_L^\infty(1+z)e^{-c z}\,dz$ for some $c>0$ depending only on
$\varepsilon$, using the tangent bound and the positive lower bound on
$I'_\theta(t)$. Sending $L\to\infty$ yields uniformly in $t$
\[
 \mathbb P(U\ge t)\sim
 \frac{\sqrt{\theta(1-\theta)}}{(t-\theta)\sqrt{\pi n}}e^{-nI_\theta(t)},
 \qquad n\,\mathbb E[U-t\mid U\ge t]=\frac1{I'_\theta(t)}+o(1).
\]
At $t=s-\varepsilon/2$ the conditional mean is $t+O(n^{-1})<s$, and at $t=s$ it strictly exceeds $s$. Hence $q$ lies in this interval. Since $s-q=\mathbb E[U-q\mid U\ge q]$, the second estimate first gives $s-q=O(n^{-1})$ and then, since $I''_\theta$ is bounded there, $q=s-1/(nI'_\theta(s))+o(n^{-1})$. Thus $n I_\theta(q)=n I_\theta(s)-1+o(1)$. Taking the reciprocal of the tail probability gives the factor $1/e$ in \eqref{eq:prefactor}.
\end{proof}

For example, fix $0<s<1$ and let $k\to\infty$ with $k/n\to0$. The largest eigenvalue of $T=nS/\tau$ is $M=s/\theta$ and $\det(T)^{-1/2}=\exp(n I_\theta(s))$. Equation~\eqref{eq:prefactor} gives
\[
 \frac{K(S/\tau)\sqrt{\det T}}{\sqrt{nM}}
 =(1+o(1))\,\frac{\sqrt{\pi}(s-\theta)}{e\sqrt{s(1-\theta)}}
 \ \longrightarrow\ \frac{\sqrt{\pi s}}{e}.
\]
Thus this family requires a prefactor of order $\sqrt{nM}$ in
Theorem~\ref{thm:uniform-det-cap} even as the upper block's dimension fraction vanishes.

For fixed $k$ and fixed $\kappa>1$, $s\to0$ and the separation
assumption of \eqref{eq:prefactor} fails. Here $nU=X/[(X+Y)/n]$,
where $X\sim\chi_k^2$ and $(X+Y)/n\to1$ in probability. Hence $nU$
converges in law to $\chi_k^2$. Also $\mathbb E[(nU)^2]=nk(k+2)/(n+2)\le k(k+2)$.
The chi-square law has no atom at any fixed cutoff, and the
second-moment bound gives uniform integrability. Both the upper-tail
probability and the truncated first moment therefore converge, and their
ratio gives the limiting conditional mean. Here $s=\kappa k/(\kappa k+n-k)$
so $ns\to\kappa k$. Strict monotonicity of the conditional mean traps $nq$ between thresholds on either side of the unique $q_\infty$ solving $\mathbb E[\chi_k^2\mid\chi_k^2\ge q_\infty]=\kappa k$. It follows that $K\to1/\mathbb P(\chi_k^2\ge q_\infty)$. Figure~\ref{fig:threshold} computes the finite-dimensional thresholds directly.

\section{Optimal Block Lengths and One-Step Gains on the Unit Sphere}
\label{sec:phase}
The sphere case supplies the reference gain for the thin shell in Theorem~\ref{conf:matched}. Theorem~\ref{thm:hinge-phase} gives the limiting gain at a specified slope growth rate and identifies the optimal block lengths. The two-block calculation below makes the transition to a common length explicit.

A failure of exact stationarity can coexist with a vanishing objective gap. We quantify the gap for the hinge profile $H_n(r)=B_n(r-1)_+$, where $(t)_+=\max(t,0)$. Let $G_n\in\mathbb R^{m_n\times n}$ have full column rank and $\|G_n\|_*=1$. Define the optimal one-step gain
\begin{equation}
 V_n=\sup_Q\left\{\langle G_n,Q\rangle-B_n\,\mathbb E(\|Q\zeta\|_2-1)_+\right\},\qquad
 \zeta\sim\mathrm{Unif}(\mathbb S^{n-1}).
 \label{eq:hinge-gain}
\end{equation}
On the ray $Q=tP$ the gain is $t-B_n(t-1)_+$, with slope one below
$t=1$ and slope $1-B_n$ above it. Thus $B_n>1$ gives a best polar gain of one.

\begin{theorem}[Optimal gain and direction for a block spectrum]\label{thm:hinge-phase}
Fix $J\ge2$. At each $n$, partition all $n$ right singular directions as
$\{1,\ldots,n\}=\bigsqcup_{j=1}^J\mathcal I_{j,n}$. The $k_{j,n}$
singular values in block $j$ are equal, with mass
$s_{j,n}=\sum_{i\in\mathcal I_{j,n}}\sigma_i(G_n)$. Thus
$\sum_j k_{j,n}=n$ and $\sum_j s_{j,n}=1$. Assume
$k_{j,n}/n\to\theta_j>0$, $s_{j,n}\to s_j>0$, $B_n\to\infty$ and
$n^{-1}\log B_n\to b\in[0,\infty)$. On the probability simplex $\Delta_J=\{u\ge0:\sum_j u_j=1\}$, put
\[
 I(u)=\tfrac12\sum_{j=1}^J\theta_j\log(\theta_j/u_j),\qquad
 K_b=\{u\in\Delta_J:I(u)\le b\},
\]
with $I=+\infty$ on the boundary. Then
\begin{align}
 \lim_{n\to\infty} V_n=V(b)
 &=\max_{x\ge0}\left\{s^\top x:\ \sum_j u_jx_j^2\le1\ \text{for every }u\in K_b\right\} \label{eq:phase-primal}\\
 &=\min_{u\in K_b}\left(\sum_j s_j^2/u_j\right)^{1/2}. \label{eq:phase-dual}
\end{align}
If $u^*$ minimizes \eqref{eq:phase-dual}, the unique optimal block scales in \eqref{eq:phase-primal} are $x_j^*=s_j/(u_j^*V(b))$. In particular, $V(b)=1$ exactly when $b\ge I(s)$. At $b=0$ the limiting block scales are proportional to $s_j/\theta_j$, the relative singular values of the gradient.
\end{theorem}

\begin{proof}
Write $G_n=U\Sigma V^\top$ in a thin singular value decomposition.
The projection $UU^\top Q=UYV^\top$, with $Y=U^\top QV$ square,
preserves the linear gain and cannot increase any displacement length.
For $O=\operatorname{diag}(O_1,\ldots,O_J)$ with orthogonal blocks,
$O\Sigma O^\top=\Sigma$, because the singular values within each block
are equal. Thus $O^\top YO$ has the same linear gain as $Y$. The
rotation-invariant input law also gives it the same expected penalty.
By Jensen's inequality, the penalty of the rotation average is at most
the average penalty, with the linear gain unchanged. Independent
block sign flips annul the cross blocks, and averaging $O_j^\top
Y_{jj}O_j$ over uniform rotations and reflections gives
$(\operatorname{tr}Y_{jj}/k_{j,n})I$. Replacing any negative block
scale by its absolute value preserves the penalty and improves the linear gain.
Thus the supremum is attained or approached among aligned matrices with nonnegative block scales $x_j$. Their linear gain is $s_n^\top x$. Their squared radius is $\sum_j U_{j,n} x_j^2$, where $U_{j,n}$ is the squared norm of the $j$th block of $\zeta$.

From $\zeta=g/\|g\|$, $U_n$ is the vector of independent chi-square
block sums divided by their total. Its Dirichlet density in the first
$J-1$ coordinates (with the last fixed by their sum) is
\[
 f_n(u)=\frac{\Gamma(n/2)}{\prod_j\Gamma(k_{j,n}/2)}
              \prod_j u_j^{k_{j,n}/2-1}.
\]
At speed $n$ the rate $I$ gives the following bounds on the simplex:
\[
 \liminf_n n^{-1}\log\mathbb P(U_n\in O)\ge-\inf_O I
 \quad(O\text{ open}),\qquad
 \limsup_n n^{-1}\log\mathbb P(U_n\in F)\le-\inf_F I
 \quad(F\text{ closed}).
\] Indeed, Stirling's formula applied to its density gives $n^{-1}\log f_n(u)\to-I(u)$ uniformly on compact subsets of the simplex interior. Integrating over small interior balls gives the lower bound for open sets. For the upper bound one can discard the boundary strip $\min_j u_j\le\varepsilon$ at arbitrarily high exponential rate as $\varepsilon\downarrow0$. Each coordinate has a beta marginal, whose density on $(0,\varepsilon)$ is bounded above for large $n$ by $\exp(O(n))u_j^{k_{j,n}/2-1}$. Integrating and summing over the fixed $J$ coordinates gives
\[
 \Pr(\min_j U_{j,n}\le\varepsilon)
 \le e^{Cn}\sum_j\frac{\varepsilon^{k_{j,n}/2}}{k_{j,n}/2}.
\]
Since $k_{j,n}/n\ge\min_i\theta_i/2$ eventually, the upper limit of the log probability divided by $n$ tends to $-\infty$ as $\varepsilon\downarrow0$. On the remaining compact interior, uniform convergence of the log density gives the closed-set upper bound.

Write $M=\max_j x_j$ and $a_n=\min_j\mathbb E\sqrt{U_{j,n}}\to\min_j\sqrt{\theta_j}>0$. The gain is at most $M-B_n(a_nM-1)$. The zero update earns zero. Every candidate with nonnegative gain satisfies
$M\le B_n/(a_nB_n-1)$ once $a_nB_n>1$. Maximizing the continuous objective on this closed box gives existence. Along a subsequence realizing $\limsup_n V_n$, take a further subsequence on which these optimal scales converge to $x$. If $\sum_j u_jx_j^2>1$ at some $u\in K_b$ with $b>0$, put $u_t=(1-t)u+t\theta$. For sufficiently small $t>0$ the strict violation persists by continuity, while $I(u_t)\le(1-t)I(u)<b$. On a neighborhood of that point, the positive hinge is bounded below away from zero for all large $n$. If $c_0>0$ is that lower bound on an open ball $O$ about $u_t$, then
for $0<\eta<b-I(u_t)$ the scaled hinge costs at least
$c_0 B_n\Pr(U_n\in O)\ge c_0\exp\{n[b-I(u_t)-\eta+o(1)]\}\to\infty$. For $b=0$, a violation at $\theta$ likewise gives divergence since $U_n\to\theta$ in probability and $B_n\to\infty$. Either contradicts optimality. Thus $x$ is feasible in \eqref{eq:phase-primal}. The penalties are nonnegative, so $\limsup_n V_n\le s^\top x\le\max_{y\ {\rm feasible}}s^\top y$.

Conversely, fix any feasible $y$ and $0<\delta<1$. For $(1-\delta)y$, the closed set
$F_\delta=\{u\in\Delta_J:\sum_j u_j[(1-\delta)y_j]^2\ge1\}$ is disjoint from $K_b$. If nonempty, it has minimum rate strictly larger than $b$ by compactness and lower semicontinuity. The hinge is bounded, so the closed-set upper bound gives a vanishing penalty after multiplication by $B_n$. Testing these scales gives $\liminf_n V_n\ge(1-\delta)s^\top y$. Send $\delta\downarrow0$ and maximize over feasible $y$ to obtain \eqref{eq:phase-primal}.

For every $u\in K_b$, Cauchy--Schwarz bounds the gain in \eqref{eq:phase-primal} by $(\sum_j s_j^2/u_j)^{1/2}$. For equality, let $u^*$ minimize $F(u)=\sum_j s_j^2/u_j$ on the compact convex set $K_b$. For each $u\in K_b$ the segment from $u^*$ to $u$ is feasible.
Differentiating there gives $\nabla F(u^*)^\top(u-u^*)\ge0$.
Substitution of $\partial_j F=-s_j^2/u_j^2$ yields
\[
 \sum_j u_j(s_j/u_j^*)^2\le \sum_j u_j^*(s_j/u_j^*)^2=F(u^*)\qquad(u\in K_b).
\]
Thus $x_j=s_j/(u_j^*\sqrt{F(u^*)})$ is feasible, and $s^\top x=(\sum_j s_j^2/u_j^*)/\sqrt{F(u^*)}=\sqrt{F(u^*)}$. Equality in Cauchy--Schwarz at $u^*$ requires precisely these scales. Since $F(u)\ge(\sum_j s_j)^2=1$ with equality only for $u=s$, the threshold is $I(s)$. At $b=0$, $K_b=\{\theta\}$.
\end{proof}

\begin{corollary}[Two spectral blocks]\label{cor:two-block-phase}
Write $\theta=(\theta_1,1-\theta_1)$ and $s=(s_1,1-s_1)$ with $s_1>\theta_1$. Let $t_b\in[\theta_1,1)$ be the upper solution of $I((t_b,1-t_b))=b$, and set $t=\min(t_b,s_1)$. Then
\[
 V(b)=\left(\frac{s_1^2}{t}+\frac{(1-s_1)^2}{1-t}\right)^{1/2},\qquad
 \frac{x_1^*}{x_2^*}=\frac{s_1(1-t)}{t(1-s_1)}.
\]
\end{corollary}

\begin{proof}
The function $s_1^2/u_1+(1-s_1)^2/(1-u_1)$ decreases up to $u_1=s_1$ and increases thereafter. The feasible interval contains $\theta_1<s_1$ and has upper endpoint $t_b$.
Its minimum is therefore reached at $t=\min(t_b,s_1)$.
\end{proof}

For equally sized blocks with gradient singular values in ratio $2:1$, one has $\theta_1=1/2$ and $s_1=2/3$. When $\log B_n=o(n)$ and $B_n\to\infty$, the optimum gain approaches $\sqrt{10}/3$, compared with $1$ on the polar ray. The gap is about $0.0541$ in the normalized one-step objective. Between $b=0$ and $I(s)=\tfrac14\log(9/8)$ the best ratio decreases continuously from $2$ to $1$. At the threshold this statement concerns the limiting gain and scales. Finite-dimensional exact optimality on the threshold depends on the subexponential factor in $B_n$.

\begin{proposition}[Finite-width gains at moderate slopes]\label{prop:moderate}
Use the hinge in \eqref{eq:hinge-gain} at slope $B>1$. At the current
width, write $\theta_j=k_j/n$ and $s_j$ for the dimension fractions and
nuclear masses of the blocks. Define the trial scales $x$ and their maximum $M$ by
\[
 C_n=\Big(\sum_j s_j^2/\theta_j\Big)^{1/2},\qquad
 x_j=\frac{s_j}{\theta_j C_n},\qquad M=\max_j x_j,\qquad
 d=M\sqrt{2\log B/n}.
\]
Write $\mu_k=\mathbb E[\chi_k/\sqrt k]$ for the normalized mean of a chi
variable and $a=\min_j\mu_{k_j}$. Let $V_n(B)$ denote the unrestricted
gain for the chosen input law. Set $\rho=\mu_n$ for the unit sphere
and $\rho=1$ for $g/\sqrt n$ with $g\sim N(0,I_n)$. If $B\ge C_n\rho/a$,
\[
 \frac{\rho C_n-M\sqrt{\pi/(2n)}}{1+d}
 \ \le\ V_n(B)\ \le\ \frac{\rho C_n}{a}.
\]
The polar-ray gain satisfies the same inequalities with
$C_n=M=1$ and $a=\mu_n$. If the block fractions stay bounded away from
zero, $B\to\infty$ and $\log B=o(n)$, the difference between the two
gains is $C_n-1+O(\sqrt{\log B/n})$, with $C_n$ evaluated at the current width.
\end{proposition}
\begin{proof}
Both input laws are rotation invariant, so the block reduction in
Theorem~\ref{thm:hinge-phase} applies. For nonnegative scales $y_j$ let
$m^2=\sum_j\theta_j y_j^2$. Their linear gain is at most $C_n m$.
For $m>0$ put $\lambda_j=\theta_j y_j^2/m^2$ and let $Y_j$ be the
length of the $j$th Gaussian block divided by $\sqrt{k_j}$.
The weights sum to one, so concavity of the square root gives, for each realization,
\[
 \Big(\sum_j\lambda_j Y_j^2\Big)^{1/2}\ge\sum_j\lambda_j Y_j.
\]
The left side is the Gaussian displacement divided by $m$. Taking expectations
gives the lower bound $\sum_j\lambda_j\mu_{k_j}\ge a$ for this ratio.
The Gaussian input is a uniform direction multiplied by an independent
radius of mean $\mu_n$, so the bound for the chosen law is $ma/\rho$.
Jensen's inequality bounds the gain by $C_n m-B(ma/\rho-1)_+$, whose slope above $m=\rho/a$ is $C_n-Ba/\rho\le0$. Its
supremum over $m\ge0$ is $C_n\rho/a$ under the stated slope condition.

For the lower bound, first maximize the linear gain over
$\sum_j\theta_j y_j^2=1$. Weighted Cauchy--Schwarz gives $s^\top y\le C_n$
with equality at the displayed $x$. We shrink this candidate below to control
its penalty. Let $A$ be its Gaussian displacement before shrinking. For $i$ in block $j$ set $w_i=x_j^2/n$.
Then $A^2=\sum_i w_i g_i^2$, $\sum_iw_i=\sum_j\theta_j x_j^2=1$,
and $0\le w_i\le l=M^2/n$. For $0<2\lambda l<1$, the convex
function $f(w)=-\log(1-2\lambda w)$ obeys $f(w_i)\le w_if(l)/l$.
Independence gives
$\mathbb E e^{\lambda A^2}=\prod_i(1-2\lambda w_i)^{-1/2}
\le(1-2\lambda l)^{-1/(2l)}$.
Taking $\lambda=[1-(1+v)^{-2}]/(2l)$ in Markov's bound yields
$\mathbb P(A\ge1+v)\le
 \exp(-[(1+v)^2-1-2\log(1+v)]/(2l))\le e^{-v^2/(2l)}$ for $v\ge0$.
Integrating from $d$ and using $(d+u)^2\ge d^2+u^2$ and $d^2/(2l)=\log B$ gives
$B\,\mathbb E(A-1-d)_+\le M\sqrt{\pi/(2n)}$.
Choose scales $\rho x/(1+d)$. For the unscaled update $Q$ put
$A=R\|Q\zeta\|$, where the Gaussian radius $R$ is independent of
$\zeta$ and has mean $\mu_n$. At fixed $\zeta$, Jensen gives
$(\mu_n\|Q\zeta\|-1-d)_+\le\mathbb E_R(R\|Q\zeta\|-1-d)_+$.
Averaging over $\zeta$ and dividing by $1+d$ gives the same expected
hinge bound for the sphere and the Gaussian.
Their linear gain is $\rho C_n/(1+d)$. For a common scale the computation
uses a single block of size $n$ and total nuclear mass one, irrespective
of the gradient spectrum. Finally $\mu_k=1-O(k^{-1})$ by Stirling's
formula and $M\le1/\sqrt{\min_j\theta_j}$, giving the stated error.
\end{proof}

\paragraph{Computing two block lengths at finite width.}
Let $U_1\sim{\rm Beta}(k_1/2,k_2/2)$, $U_2=1-U_1$ and
$r_x=(U_1 x_1^2+U_2x_2^2)^{1/2}$. For the tabulated interior optima
$x_1\ne x_2$, the continuous Beta law gives $\mathbb P(r_x=1)=0$.
The coordinate difference quotients of $(r_x-1)_+$ are bounded by
$\sqrt{U_j}\le1$, by the reverse triangle inequality. Differentiation
under the expectation, using $\partial r_x/\partial x_j=U_jx_j/r_x$,
yields the two stationarity equations
\begin{equation}\label{eq:finite-foc}
 s_{j,n}= B_n\,x_j\,\mathbb E\left[\frac{U_j}{r_x}\mathbf1_{\{r_x>1\}}\right],
 \qquad j=1,2 .
\end{equation}
At any positive interior solution of both equations with $x_1\ne x_2$,
the gradient of the concave objective vanishes, proving global optimality.
At $x_1=x_2=1$ every sample lies on the kink, so use the selector in \eqref{eq:selector}.
Subtract one from the reported total gain to obtain the absolute gap.

\begin{table}[htbp]
\centering
\caption{Optimal update ratio $x_1/x_2$ and unrestricted gain for two equal blocks with singular values in ratio $2:1$ on the unit sphere. The best polar gain is one in both columns. Each finite-width row uses deterministic integration of the beta density and the first-order equations \eqref{eq:finite-foc}. The final row uses Corollary~\ref{cor:two-block-phase} at $b=0$ for
$B_n=n$ and at $b=0.01$ for $B_n=e^{0.01n}$.}
\smallskip
\begin{tabular}{r cc cc}
\toprule
&\multicolumn{2}{c}{$B_n=n$}&\multicolumn{2}{c}{$B_n=\exp(0.01n)$}\\
\cmidrule(lr){2-3}\cmidrule(lr){4-5}
$n$ & $x_1/x_2$ & unrestricted gain $V_n$ & $x_1/x_2$ & unrestricted gain $V_n$\\
\midrule
512   & 1.34292 & 1.00968 & 1.40242 & 1.01275\\
2048  & 1.60122 & 1.02477 & 1.36295 & 1.01068\\
8192  & 1.77035 & 1.03658 & 1.34707 & 1.00989\\
Limit & 2       & 1.05409 & 1.33885 & 1.00948\\
\bottomrule
\end{tabular}
\end{table}

\paragraph{Hinge powers above one preserve the limiting gains.}
The limiting gain and block scales in Theorem~\ref{thm:hinge-phase} also hold for $H_n(r)=B_n(r-1)_+^p$ with a fixed $p>1$. For compactness one may use $(r-1)_+^p\ge r-2$ for all $r\ge0$. On a compact set of block scales the penalty before multiplication by $B_n$ is bounded, has a positive lower bound on a small neighborhood of each strictly violating point, and vanishes under exactly the same constraints. These are precisely the properties used in the upper and lower limit arguments. In finite dimension this profile is differentiable at every positive radius. At $cP$, $c>0$, the penalty gradient for each unit input is
$H_n'(c)P\zeta\zeta^\top$. Averaging yields $H_n'(c)P/n$, which
cannot match $G=PS$ for nonscalar $S$. On the polar ray, the derivative
above one vanishes at $t=1+(pB_n)^{-1/(p-1)}$. Substitution into
$t-B_n(t-1)_+^p$ gives the best gain  $1+(1-1/p)(pB_n)^{-1/(p-1)}\to1$. At and above $b=I(s)$, the unrestricted and polar-ray gains therefore both tend to one even for this differentiable penalty.

\section{A Thin Shell Preserves Limiting One-Step Gains}
\label{sec:shell}
The shell half of Theorem~\ref{conf:matched} follows from the finite-width bounds \eqref{eq:shell-sandwich}. They compare both gains with those on spheres of two deterministic radii. The argument allows dependence between the radius and direction as long as the direction has a uniform marginal law.

Replace $\zeta$ by $R_n\zeta$ in \eqref{eq:hinge-gain}, keeping
$G_n$ deterministic with full column rank and nuclear norm one. For
$B\ge0$ write $V_{R,n}(B)$ for the unrestricted supremum and
$V_{\mathrm{pol},R,n}(B)$ for the supremum on the nonnegative polar
ray. Each may be $+\infty$. For instance, on the unit sphere with
$0<B<1$ the gain at $tP_n$ equals $(1-B)t+B$ for $t\ge1$, so both
suprema are infinite. Whenever both are finite, define the absolute gap
\[
 \Delta_{R,n}(B)=V_{R,n}(B)-V_{\mathrm{pol},R,n}(B).
\]
Use the subscript $1$ for the unit sphere and $r$ for a deterministic
radius $r>0$. The following comparison holds in the extended reals.

\begin{theorem}[Comparison at two deterministic radii]\label{thm:shell}
Suppose $0<\ell_n\le R_n\le h_n<\infty$ almost surely and the marginal law of $\zeta$ is uniform. For each $B\ge0$, let $W$ denote either gain, with the same choice on both sides. Then
\begin{equation}
 \frac{W_{1,n}(h_n B)}{h_n}\ \le\ W_{R,n}(B)\
 \le\ \frac{W_{1,n}(\ell_n B)}{\ell_n}.
 \label{eq:shell-sandwich}
\end{equation}
In particular, assume the block and budget hypotheses of Theorem~\ref{thm:hinge-phase}, and $\ell_n,h_n\to1$. Then
\[
 V_{R,n}(B_n)\longrightarrow V(b),\qquad
 V_{\mathrm{pol},R,n}(B_n)\longrightarrow 1,\qquad
 \Delta_{R,n}(B_n)\longrightarrow V(b)-1.
\]
Thus the limiting gap is zero exactly for $b\ge I(s)$. Under the stated budget assumptions, the shell endpoints may approach one at any rate.
\end{theorem}

\begin{proof}
At each input and for each $Q$,
\[
 (\ell_n\|Q\zeta\|-1)_+\le (R_n\|Q\zeta\|-1)_+
            \le(h_n\|Q\zeta\|-1)_+.
\]
Subtracting $B$ times these costs from the same linear gain reverses the
inequalities. Taking the expectation and the supremum therefore gives
$W_{h_n,n}(B)\le W_{R,n}(B)\le W_{\ell_n,n}(B)$. For a deterministic radius $r>0$ put $Y=rQ$. Then
\[
 \langle G_n,Q\rangle-B\mathbb E(r\|Q\zeta\|-1)_+
 =r^{-1}\big[\langle G_n,Y\rangle-rB\mathbb E(\|Y\zeta\|-1)_+\big].
\]
The substitution is bijective both on all matrices and on the polar ray,
giving $W_{r,n}(B)=W_{1,n}(rB)/r$. This proves
\eqref{eq:shell-sandwich} without independence of radius and direction.
For $r_n=\ell_n$ or $h_n$, $r_n\to1$ gives $r_nB_n\to\infty$ and
$n^{-1}\log(r_nB_n)=n^{-1}\log B_n+n^{-1}\log r_n\to b$.
The compactness estimate of Theorem~\ref{thm:hinge-phase} makes
both upper bounds eventually finite. Taking the two limits separately,
using $V_{\mathrm{pol},1,n}(rB_n)=1$ once $rB_n>1$, now justifies
subtracting the gains.
\end{proof}

For example, take $R_n$ uniform on $[1-\delta_n,1+\delta_n]$ independently of $\zeta$, with $0<\delta_n<1$ and $\delta_n\to0$. For nonscalar spectra Xu's atomless-radius result excludes an exact polar minimizer at any positive step for each finite $n$ \citep[Corollary~3.2]{xu}. The comparison theorem determines when the best polar step nevertheless attains the limiting optimal value. For $B_n\to\infty$ with $b<I(s)$ and $s\ne\theta$, the limiting gap is $V(b)-1>0$.

\section{Normalized Gaussian Inputs Retain an Objective Gap}
\label{sec:gaussian}
Theorem~\ref{thm:gaussian-phase} supplies the Gaussian half of Theorem~\ref{conf:matched}. It computes the limiting unrestricted gain and the polar-ray gain separately. Proposition~\ref{prop:gauss-joint} also treats slope sequences whose exponential growth rate diverges.

The shrinking-shell comparison used a deterministic upper endpoint.
A Gaussian radius exceeds every finite endpoint with positive probability.
Here we control that tail after multiplying the hinge by $B_n$.
Replace the unit direction in \eqref{eq:hinge-gain} with
$g/\sqrt n$, where $g\sim N(0,I_n)$. The radius $\|g\|_2/\sqrt n$
converges to one in probability. At finite exponential slope rates,
rare radii can nevertheless affect both the optimal learning rate
on the polar ray and the gain obtainable by changing the update spectrum.
Use the subscript $\mathrm{G}$ for the Gaussian input law. Write
$V_{\mathrm{G},n}$ for the unrestricted gain and $V_{\mathrm{pol},\mathrm{G},n}$
for the supremum over $Q=tP_n$, $t\ge0$, where $G_n=P_n S_n$.

\begin{theorem}[Gaussian gain and polar-ray gap]\label{thm:gaussian-phase}
At each width assume $G_n$ has full column rank and nuclear norm one.
Partition its right singular directions into a fixed number $J$ of blocks
of sizes $k_{j,n}$ exhausting the $n$ coordinates. Within each block all
singular values are equal at that width. Assume $k_{j,n}/n\to\theta_j>0$
and $s_{j,n}=\sum_{i\in j}\sigma_i(G_n)\to s_j>0$, as in
Theorem~\ref{thm:hinge-phase}.
Let $B_n\to\infty$ with $n^{-1}\log B_n\to b\in[0,\infty)$.
The energies $Z_{j,n}=n^{-1}\sum_{i\in j}g_i^2$ sum to the random
squared radius $H_n=\|g\|^2/n$, whereas the sphere energies $U_{j,n}$
sum to one. For $z\in(0,\infty)^J$ put
\[
 L(z)=\frac12\sum_{j=1}^J
 \left[z_j-\theta_j-\theta_j\log\!\left(\frac{z_j}{\theta_j}\right)\right],
 \qquad \mathcal Z_b=\{z:L(z)\le b\}.
\]
Extend $L$ by $+\infty$ outside the positive orthant.
For every finite $b\ge0$, the limiting optimal gain is
\begin{align}
 V_{\mathrm{G}}(b):=\lim_n V_{\mathrm{G},n}
 &=\max_{x\ge0}\left\{s^\top x:\ 
                  \sum_j z_jx_j^2\le1 \text{ for every } z\in\mathcal Z_b\right\}
 \label{eq:gauss-primal}\\
 &=\min_{z\in\mathcal Z_b}\left(\sum_j s_j^2/z_j\right)^{1/2}.
 \label{eq:gauss-dual}
\end{align}
There is a unique minimizing $z^*$. Every sequence of aligned block
scales whose gains approach the optimum to within $o(1)$ converges
to $x_j^*=s_j/(z_j^*V_{\mathrm{G}}(b))$. For $b=0$, $z^*=\theta$.
For $b>0$, there is a unique $\alpha>0$ such that $L(z^*)=b$ and
\begin{equation}\label{eq:gauss-kkt}
 z_j^*=\tfrac12\left(\theta_j+\sqrt{\theta_j^2+4\alpha s_j^2}\right).
\end{equation}
Let $h_b$ be the solution in $[1,\infty)$ of $h_b-1-\log h_b=2b$.
The optimal gain on the polar ray satisfies
\begin{equation}\label{eq:gauss-polar}
 V_{\mathrm{pol},\mathrm{G}}(b):=\lim_n V_{\mathrm{pol},\mathrm{G},n}=h_b^{-1/2}.
\end{equation}
Consequently $V_{\mathrm{G}}(b)>V_{\mathrm{pol},\mathrm{G}}(b)$ for every finite $b\ge0$
whenever $s\ne\theta$. Equality holds for $s=\theta$.
\end{theorem}

The equal-value hypothesis within each block matters even at $b=0$.
For $n$ divisible by four, divide the directions into two equal groups,
each containing equal counts of singular values $(1+\rho)/n$ and
$(1-\rho)/n$, with $0<\rho<1$. Each group then carries half the nuclear
norm. Grouping the directions instead by their actual singular values
gives $\theta=(1/2,1/2)$ and $s=((1+\rho)/2,(1-\rho)/2)$.
The unrestricted gain tends to $\sqrt{1+\rho^2}$, whereas treating the
original heterogeneous groups as equal-value blocks would give one.

The positive block fractions enter through the following estimate. A
block with $k_{j,n}=o(n)$ has a different boundary behavior at $z_j=0$.

\begin{lemma}[Gaussian block energies]\label{lem:gauss-ldp}
Let $nZ_{j,n}\sim\chi^2_{k_{j,n}}$ independently, with
$\sum_j k_{j,n}=n$ and $k_{j,n}/n\to\theta_j>0$ for fixed $J$.
With $L$ as in Theorem~\ref{thm:gaussian-phase}, for open $O$ and closed
$F$ in $\mathbb R^J$,
\[
 \liminf_n n^{-1}\log\Pr(Z_n\in O)\ge-\inf_O L,\qquad
 \limsup_n n^{-1}\log\Pr(Z_n\in F)\le-\inf_F L.
\]
The finite sublevel sets of $L$ are compact.
\end{lemma}
\begin{proof}
The density of a coordinate at $z>0$ is
\[
 \frac{(n/2)^{k/2}}{\Gamma(k/2)}z^{k/2-1}e^{-nz/2},
 \qquad k=k_{j,n}.
\]
On an interior rectangle $[\eta,D]^J$, Stirling's formula for the
joint density $f_n$ gives $n^{-1}\log f_n(z)=-L_n(z)+O_{\eta,D,J}(\log n/n)$,
where $L_n$ replaces each $\theta_j$ in $L$ by $k_{j,n}/n$.
Since $\log z_j$ is bounded on this rectangle, $L_n\to L$ uniformly there. Hence $\sup_{[\eta,D]^J}|n^{-1}\log f_n+L|\to0$. Integrating
over any small ball about an interior point of an open set gives the
open-set lower bound by continuity and then shrinking the ball.
For the upper bound, write $q_{j,n}=k_{j,n}/n$. For $0<\eta<q_{j,n}$,
use $\mathbb E e^{t\chi_k^2}=(1-2t)^{-k/2}$ and
a negative tilt $t$. On $\{nZ_{j,n}\le n\eta\}$ one has
$e^{tnZ_{j,n}}\ge e^{tn\eta}$, hence Markov's bound is
$e^{-tn\eta}(1-2t)^{-k_{j,n}/2}$. Taking $t=(1-q_{j,n}/\eta)/2<0$ gives
$\mathbb P(Z_{j,n}\le\eta)\le
  \exp\{-n[\eta-q_{j,n}-q_{j,n}\log(\eta/q_{j,n})]/2\}$.
The bracket tends uniformly to infinity as $\eta\downarrow0$ for large
$n$, since the $q_{j,n}$ are bounded away from zero. At the upper end,
$\sum_jZ_{j,n}\sim\chi_n^2/n$ gives, with $t=(1-1/D)/2>0$ for $D>1$,
\begin{equation}\label{eq:gauss-trunc}
 \mathbb P\big(\textstyle\sum_j Z_{j,n}>D\big)
 \le \exp[-n(D-1-\log D)/2].
\end{equation}
Thus the complement of $[\eta,D]^J$ has arbitrarily large exponential
cost by a union bound. On the intersection of the rectangle and a closed
set, bound the density by its supremum and multiply by the rectangle's volume. Sending
the complement cost to infinity gives the closed-set upper bound.
Finally $L$ diverges at the coordinate boundaries and at infinity,
so its sublevel sets are compact.
\end{proof}

\begin{lemma}[Limits of near-optimal scales]\label{lem:gauss-near}
Fix $p\ge1$. Under the hypotheses of Theorem~\ref{thm:gaussian-phase},
let $\mathcal D$ be either all block scales $[0,\infty)^J$ or the ray
$\{t\mathbf1:t\ge0\}$. For penalty $B_n(r-1)_+^p$, the optimal
gain over $\mathcal D$ converges to
\[
 \max\{s^\top x:x\in\mathcal D,\ \textstyle\sum_j z_jx_j^2\le1
                   \text{ for every }z\in\mathcal Z_b\}.
\]
Every sequence in $\mathcal D$ within $o(1)$ of its finite-width optimum
is bounded for large $n$, and every limit point attains this maximum on $\mathcal D$.
\end{lemma}
For example, at $b=0$ with two equal blocks in gradient singular-value
ratio $2:1$, the sole energy constraint is $(x_1^2+x_2^2)/2\le1$.
The maximum of $(2x_1+x_2)/3$ is $\sqrt{10}/3$ over all scales and
one on the ray $x_1=x_2$.
\begin{proof}
Write $r(z,x)=(\sum_j z_j x_j^2)^{1/2}$,
$M=\max_j x_j$, and $a_n=\min_j\mathbb E\sqrt{Z_{j,n}}$.
Since $Z_{j,n}\to\theta_j$ in $L^2$, $a_n\to\min_j\sqrt{\theta_j}>0$.
The inequalities $(r-1)_+^p\ge r-2$ and
$\mathbb E r(Z_n,x)\ge a_n M$ bound the gain by
$M-B_n(a_nM-2)$. The zero update gives zero gain. A near-maximizer
therefore has $M\le (2B_n+o(1))/(a_n B_n-1)$ for large $n$.

Take a convergent subsequence of such scales with limit $x$.
If $r(z,x)>1$ at some $z\in\mathcal Z_b$ and $b>0$, put $z_t=(1-t)z+t\theta$.
For all sufficiently small $t>0$, $r(z_t,x)>1$ by continuity and
$L(z_t)\le(1-t)L(z)<b$. There is an open ball about $z_t$ on
which the penalty before multiplication by $B_n$ is uniformly positive
along the subsequence (shrink the ball to have compact closure).
Let $\kappa>0$ be the resulting lower bound on this ball and choose
$\eta>0$ with $L(z_t)+2\eta<b$. Lemma~\ref{lem:gauss-ldp} and the
budget rate give a scaled cost of at least
$\kappa\exp\{n[b-L(z_t)-2\eta]\}\to\infty$.
For $b=0$ the sole constraint is at $\theta$, and concentration there
gives divergence since $B_n\to\infty$. Either case contradicts gain
at least $-o(1)$. The limit point is feasible, and its linear gain
$s^\top x$ bounds the subsequential upper limit.

For any feasible $y$ in the chosen domain, set $y'=(1-\delta)y$ with $0<\delta<1$.
The closed set $\{z\ge0:r(z,y')\ge1\}$ has rate infimum strictly above
$b$ by compactness of the sublevels and $r(z,y')\le1-\delta$ on
$\mathcal Z_b$. On $H_n=\sum_jZ_{j,n}\le D$ the penalty before scaling
is bounded by $(\|y'\|_\infty\sqrt D)^p$, so the closed-set bound makes
this part vanish after scaling. On $H_n>D$, Cauchy--Schwarz bounds its
expectation by
$\|y'\|_\infty^p(\mathbb E H_n^p)^{1/2}\mathbb P(H_n>D)^{1/2}$.
For $H_n=n^{-1}\sum_i g_i^2$, Jensen gives
$\mathbb E H_n^p\le\mathbb E|g_1|^{2p}<\infty$. Choose $D$ with
$(D-1-\log D)/4>b$. The denominator is four because the probability
in \eqref{eq:gauss-trunc} appears under a square root. This exponent grows without bound with $D$, so the choice covers each finite $b$. The scaled tail also
vanishes. Sending $\delta\downarrow0$ gives a lower limit of at least $s^\top y$ for each feasible $y\in\mathcal D$. The upper and lower bounds therefore coincide with the maximum on $\mathcal D$. Both domains are closed and preserved by multiplication by $1-\delta$, so the argument also applies to sequences constrained to the ray.
\end{proof}

\begin{proof}[Proof of Theorem~\ref{thm:gaussian-phase}]
Rotation invariance and convexity reduce the optimization to aligned
nonnegative block scales as in Theorem~\ref{thm:hinge-phase}. Their
squared displacement is $\sum_jZ_{j,n}x_j^2$ with the independent
energies of Lemma~\ref{lem:gauss-ldp}. Lemma~\ref{lem:gauss-near} with
$p=1$ gives \eqref{eq:gauss-primal}.
The compact convex set $\mathcal Z_b$ lies inside the positive orthant.
It has a unique minimizer of the strictly convex $F(z)=\sum_j s_j^2/z_j$.
Cauchy--Schwarz gives $s^\top x\le\sqrt{F(z)}$ for every feasible $x$
and $z\in\mathcal Z_b$. At the minimizer $z^*$, $\partial_jF(z^*)=-s_j^2/(z_j^*)^2$ and
$\nabla F(z^*)^\top(z-z^*)\ge0$ along every feasible segment. Rearranging gives
$\sum_j z_j(s_j/z_j^*)^2\le F(z^*)$ for every $z\in\mathcal Z_b$.
Thus $x_j^*=s_j/(z_j^*\sqrt{F(z^*)})$ satisfies every energy
constraint, and $s^\top x^*=\sqrt{F(z^*)}$.
Equality in Cauchy--Schwarz makes $x^*$ unique, which, together with
Lemma~\ref{lem:gauss-near}, gives convergence of the aligned near-maximizers. If $b>0$,
the constraint is active since $F$ strictly decreases in every coordinate.
The point $\theta$ is strictly feasible, so the multiplier $\lambda$
for $L(z)\le b$ exists. Stationarity gives
$s_j^2/(z_j^*)^2=(\lambda/2)(1-\theta_j/z_j^*)$ and $\lambda>0$.
Setting $\alpha=2/\lambda$ gives \eqref{eq:gauss-kkt}. Each coordinate
in that formula increases from $\theta_j$ to infinity with $\alpha$,
and $L$ strictly increases along this path, giving uniqueness.

Finally write $h=\sum_j z_j$ and $u=z/h$. Then
\[
 2L(z)=h-1-\log h+\sum_j\theta_j\log(\theta_j/u_j).
\]
The relative entropy in the sum is nonnegative, with equality precisely
at $u=\theta$. Hence $\max_{z\in\mathcal Z_b}\sum_j z_j=h_b$, attained
only at $z=h_b\theta$. The polar constraint on a common scale $t$ is
$t^2h_b\le1$, proving \eqref{eq:gauss-polar}.
Since $F(z)\ge1/h\ge1/h_b$ with equality throughout only if
$u=s$ in the first inequality and $z=h_b\theta$ in the second,
compact attainment gives the strict gap when $s\ne\theta$. If $s=\theta$,
the candidate $z=h_b\theta$ has $L(z)=b$ and $F(z)=1/h_b$, attaining equality also at $b=0$.
\end{proof}

The same reduction applies to every fixed $p>1$ since
$r\mapsto(r-1)_+^p$ is convex and nondecreasing.
Lemma~\ref{lem:gauss-near} therefore gives the same limiting
gains and block scales for these differentiable penalties.

\paragraph{Uniform convergence over bounded exponential slope rates.}
Write $V_{\mathrm{G},n}(B)$ for the gain at slope $B$. For every fixed $C>0$
and every $L_n\to\infty$ with $\log L_n=o(n)$,
\[
 \sup_{L_n\le B\le e^{nC}}
 \big|V_{\mathrm{G},n}(B)-V_{\mathrm{G}}(n^{-1}\log B)\big|\longrightarrow0 .
\]
The assertion holds also on the polar ray. Indeed, along any sequence
of slopes in this range, including one on which the supremal error stays above a positive constant if uniformity failed, the log rates have a convergent subsequence,
and Theorem~\ref{thm:gaussian-phase} applies on that subsequence.
The limiting formulas are continuous in $b$ by \eqref{eq:gauss-kkt}
with $\alpha\downarrow0$ at $b=0$, and by \eqref{eq:gauss-polar}.
Consequently the error must tend to zero on this subsequence. This proves uniform convergence without a quantitative rate.

The scale ratios also follow a monotone interpolation. Put
$r_j=s_j/\theta_j$. Equation~\eqref{eq:gauss-kkt} gives
$s_j/z_j^*=2r_j/(1+\sqrt{1+4\alpha r_j^2})$.
Denote the right side by $f_\alpha(r_j)$. Its logarithmic derivative
in $r$ is $\partial_{\log r}\log f_\alpha(r)=1/\sqrt{1+4\alpha r^2}>0$.
Thus $r_i>r_j$ gives $x_i^*/x_j^*=f_\alpha(r_i)/f_\alpha(r_j)>1$.
Moreover,
\[
 \partial_{\log\alpha}\log f_\alpha(r)=
 -\tfrac12+\frac1{2\sqrt{1+4\alpha r^2}}.
\]
This derivative is smaller at $r_i$ than at $r_j$, so the ratio decreases strictly with $\alpha$ and hence with $b=L(z^*(\alpha))$.
The ratio tends to $r_i/r_j$ as $b\downarrow0$ and to one as
$b\to\infty$, taking the width limit first. For fixed-radius inputs
the optimal scales become equal above a finite limiting rate by
Theorem~\ref{thm:hinge-phase}. In the Gaussian model a pair with
$r_i>r_j$ retains strict separation at every finite rate.

The Gaussian gain also admits a direct expression in terms of the
fixed-sphere gain $V$ in \eqref{eq:phase-dual}. With
$\psi(h)=(h-1-\log h)/2$, it is
\begin{equation}\label{eq:gauss-angular}
 V_{\mathrm{G}}(b)=\min_{h>0:\,\psi(h)\le b}\frac{V\big(b-\psi(h)\big)}{\sqrt h}.
\end{equation}
Here $h$ is the squared input radius. Writing $z=hu$ in \eqref{eq:gauss-dual} gives
$L(hu)=\psi(h)+I(u)$, and the reciprocal-square objective
$\sum_j s_j^2/(hu_j)$ is $h^{-1}\sum_j s_j^2/u_j$.
At $h=\sum_j z_j^*$, the sphere optimizer at rate $b-\psi(h)$
therefore has the same ratios of update scales as the Gaussian optimizer
at rate $b$. This identity shows exactly how a radial deviation uses
part of the available exponential rate.

\begin{proposition}[Joint superexponential limit]\label{prop:gauss-joint}
Let $G_n$ have full column rank and $\|G_n\|_*=1$, and use the Gaussian
input and hinge penalty of Theorem~\ref{thm:gaussian-phase}. If
$b_n=n^{-1}\log B_n\to\infty$, then for every sequence of gradient spectra
\[
 V_{\mathrm{G},n}\sim V_{\mathrm{pol},\mathrm{G},n}\sim(2b_n)^{-1/2}.
\]
\end{proposition}
\begin{proof}
Fix $0<\varepsilon<1$ and write $M_0=(1+\varepsilon)/\sqrt{2b_n}$.
For any update let $M=\|Q\|_{\rm op}$. Its linear gain is at most $M$
by nuclear norm duality, and its displacement is at least
$M|g_1|/\sqrt n$ in a right singular vector basis whose first vector corresponds to $M$. If $M\ge M_0$,
the event $|g_1|\ge(1+\varepsilon/2)\sqrt n/M_0$ gives hinge at least
$(1+\varepsilon/2)M/M_0-1\ge(\varepsilon/2)M/M_0$. The probability $p_n$ of this event satisfies
\[
 \frac{\log p_n}{n b_n}\longrightarrow
 -\frac{(1+\varepsilon/2)^2}{(1+\varepsilon)^2}>-1 ,
\]
as follows from the standard normal tail. At $u_n=(1+\varepsilon/2)\sqrt n/M_0$, it gives $\log\Pr(|g_1|\ge u_n)=-u_n^2/2+O(\log u_n)$. Substituting $M_0$ yields the displayed coefficient, with $\log u_n=o(nb_n)$. Since $\log B_n=nb_n$, it follows that $B_n p_n/M_0\to\infty$
and their gain is at most $M[1-(\varepsilon/2)B_np_n/M_0]<0$ eventually. For $M<M_0$
the gain is at most $M_0$.

For a lower bound take the polar step $tP_n$ with
$t=(1-\varepsilon)/\sqrt{2b_n}$ and write $R_n=\|g\|_2/\sqrt n$.
Its gain is $t-B_n\mathbb E(tR_n-1)_+$. For $q\ge2$,
H\"older's inequality bounds the expected hinge by
$t(\mathbb E R_n^q)^{1/q}\mathbb P(R_n^2>t^{-2})^{1-1/q}$.
Jensen bounds the moment by $\mathbb E|g_1|^q$ uniformly in $n$.
In \eqref{eq:gauss-trunc} the exponent per dimension at $D=t^{-2}$ is
$(D-1-\log D)/2\sim b_n/(1-\varepsilon)^2$. Choose fixed $q$ so that
$1-1/q>(1-\varepsilon)^2$. The logarithm of
$B_n\mathbb E(tR_n-1)_+/t$, divided by $nb_n$, is at most
$1-(1-1/q)/(1-\varepsilon)^2+o(1)$, eventually negative by a fixed
margin. Thus the scaled penalty is $o(t)$. The unrestricted optimum dominates this polar gain.
Letting $\varepsilon\downarrow0$ gives both equivalents.
\end{proof}

\section{Matched Radial Moments and Finite-Width Gain Calculations}
\label{sec:matched-radii}
Here we construct the two input laws of Theorem~\ref{conf:matched} on the same angular distribution. The formulas at the end of this appendix give a common method for computing the two gain gaps at finite width.

The difference between the shell and the Gaussian model persists after matching the mean and variance of the input radius \emph{exactly} at each sufficiently large width. Let $R_n^{\rm G}=\|g\|_2/\sqrt n$ for $g\sim N(0,I_n)$. Define
\[
 \mu_n=\sqrt{2/n}\,\frac{\Gamma((n+1)/2)}{\Gamma(n/2)},\qquad
 d_n=\sqrt{3(1-\mu_n^2)},\qquad
 R_n^{\rm U}\sim{\rm Unif}[\mu_n-d_n,\mu_n+d_n].
\]
Use independent uniform angles in both models. The chi density gives $\mathbb E[R_n^{\rm G}]=\mu_n$ and $\mathbb E[(R_n^{\rm G})^2]=1$. The uniform radius has mean $\mu_n$ and variance
$(2d_n)^2/12=1-\mu_n^2$, so its second moment also equals one. The gamma ratio gives $\mu_n=1-1/(4n)+O(n^{-2})$, so the uniform radius is positive for large $n$ and lies on a shrinking shell. Independence, $\mathbb E\zeta=0$ and
$\mathbb E\zeta\zeta^\top=I_n/n$ give
$\operatorname{Cov}(R\zeta)=\mathbb E R^2\,I_n/n=I_n/n$ under both laws.

\begin{corollary}[Matched radial moments]\label{cor:matched-radii}
Under the block hypotheses of Theorem~\ref{thm:hinge-phase}, take $s\ne\theta$ and $B_n\to\infty$ with $\log B_n/n\to b<\infty$. For every $b\ge I(s)$ the best polar-ray gain has a vanishing gap for $R_n^{\rm U}$ and a strictly positive limiting gap for $R_n^{\rm G}$. Both laws are atomless. At $b=0$ both limiting gaps equal $\sqrt{\sum_j s_j^2/\theta_j}-1$.
\end{corollary}

\begin{proof}
The comparison theorem applies with $\ell_n=\mu_n-d_n$, $h_n=\mu_n+d_n$. The Gaussian variational formula above gives strict inequality between optimal and polar-ray gains for $s\ne\theta$ at every finite $b$. At $b=0$ the nonnegative rates $I$ and $L$ vanish only at $\theta$.
Thus the energy sets in the sphere and Gaussian dual problems
\eqref{eq:phase-dual} and \eqref{eq:gauss-dual} each reduce to
$\{\theta\}$. Substitution gives limiting unrestricted gain
$(\sum_j s_j^2/\theta_j)^{1/2}$ in both laws. The shell comparison and
\eqref{eq:gauss-polar} with $h_0=1$ give common-step gains tending to one.
\end{proof}

For equal block dimensions and singular values in ratio $2:1$, the shell transition is $I(s)=\tfrac14\log(9/8)\simeq0.02945$. Take $B_n=e^{0.06n}$. At $n=256$ the optimal update ratio $x_1^*/x_2^*$ (larger-gradient block over smaller) is $1.006608$ for the moment-matched uniform shell and $1.447020$ for Gaussian inputs. The corresponding gains over the best polar steps are $0.00006566$ and $0.022980$. The shell values use analytic integration over the radius at each angle. At $n=1024$ the Gaussian gain difference is $0.022087$, close to its predicted limit $0.021673$. At this budget the shell limiting gap is zero. Figure~\ref{fig:radial} shows the dependence on the slope budget.

\paragraph{Quadrature for the shell and Gaussian gain gaps.}
Let $R$ be independent of $U=(U_1,1-U_1)$ as in \eqref{eq:finite-foc}.
For the finite-width nuclear shares $s_{j,n}$, define the angular
displacement by $r_x^2=\sum_j U_j x_j^2$. The input displacement has
length $Rr_x$. At positive scales the atomless independent radius gives
$\Pr(Rr_x=1)=0$ by conditioning on $U$. Differentiating the radial hinge and
solving separately for the common scale $t$ yields
\[
 s_{j,n}= B_n x_j\,\mathbb E\!\left[\frac{R U_j}{r_x}
                 \mathbf1_{\{Rr_x>1\}}\right],\qquad
 1=B_n\,\mathbb E[R\mathbf1_{\{Rt>1\}}].
\]
Multiply the first stationarity equations by $x_j$ and sum to obtain
$\sum_j s_{j,n}x_j=B_n\mathbb E[Rr_x\mathbf1_{\{Rr_x>1\}}]$. Expanding
$(Rr_x-1)_+$ in the objective then gives gain $B_n\Pr(Rr_x>1)$.
The ray optimum has gain $B_n\Pr(Rt>1)$ by the second equation, with
$t$ chosen separately. Their difference gives Table~\ref{conf:finite-table}.
For a uniform radius on $[\ell,h]$, put
$v=\min(h,\max(\ell,1/r_x))$ at a fixed angle. Let $E=\{Rr_x>1\}$. Then
$\Pr(E\mid U)=(h-v)/(h-\ell)$ and
$\mathbb E[R\mathbf1_E\mid U]=(h^2-v^2)/(2(h-\ell))$. The latter
is the truncated first moment used in the step equations. The remaining angular integral uses
Gauss--Legendre quadrature for $U_1\sim\mathrm{Beta}(n/4,n/4)$.
For the shell, analytic integration over $R$ and subdivision of the angular
interval at the radial endpoints allow checks with 112 and 176 nodes per
subinterval. For the Gaussian, the objective and stationarity at the
112-node solution are checked with 176 angular nodes and an independent
gamma-tail evaluation. Across Table~\ref{conf:finite-table}, the largest discrepancy in
these checks is below $2\times10^{-12}$. These are numerical consistency
checks at the reported parameters.

One can also quantify how the strict Gaussian gap closes at large exponential budgets. Write $V_{\mathrm{G}}(b)$ and $V_{\mathrm{pol},\mathrm{G}}(b)$ for the two limiting gains. The order of limits here is first $n\to\infty$ at fixed $b$, then $b\to\infty$.

\begin{corollary}[Large-budget relative gap]\label{cor:large-budget}
For fixed positive $\theta,s$ with $s\ne\theta$, let $D=\sum_j\theta_j\log(\theta_j/s_j)$. Then
\[
 \frac{V_{\mathrm{G}}(b)}{V_{\mathrm{pol},\mathrm{G}}(b)}-1
   =\frac{D}{2h_b}+O_{\theta,s}(h_b^{-2})
   =\frac{D}{4b}+O_{\theta,s}\!\left(\frac{\log b}{b^2}\right).
\]
\end{corollary}
\begin{proof}
Let $z_j$ be the optimal Gaussian block energies and put $m=\sum_jz_j$, $u_j=z_j/m$. The Gaussian rate decomposes as
\[
 2b=m-1-\log m+\sum_j\theta_j\log(\theta_j/u_j).
\]
From the optimality equations $z_j^2-\theta_jz_j=\alpha s_j^2$ for a common $\alpha>0$. At $b\to\infty$, $\alpha\to\infty$, so $z_j=\sqrt{\alpha}\,s_j+\theta_j/2+O_{\theta,s}(\alpha^{-1/2})$. Hence $u=s+O_{\theta,s}(m^{-1})$ componentwise. Strict positivity of the fixed masses gives $D(\theta\Vert u)=D+O_{\theta,s}(m^{-1})$. Write $h=h_b$. The difference of $t-1-\log t$ at $h$ and $m$ is $D(\theta\Vert u)$. The mean-value theorem gives $(h-m)(1-1/\xi)=D+O_{\theta,s}(m^{-1})$ for some $\xi\in[m,h]$. Since $h\ge m\to\infty$, the derivative factor eventually exceeds $1/2$, so $h-m=O_{\theta,s}(1)$ and $h/m\to1$. Dividing by the derivative factor now gives $h-m=D+O_{\theta,s}(h^{-1})$. Also $h=2b+1+\log h=2b+O(\log b)$. The gain formula yields
\[
 \frac{V_{\mathrm{G}}(b)}{V_{\mathrm{pol},\mathrm{G}}(b)}
 =\sqrt{\frac h m}\sqrt{\sum_j s_j^2/u_j}.
\]
The second factor is $1+O(m^{-2})$ by $\sum_j s_j^2/u_j=1+\sum_j(s_j-u_j)^2/u_j$. The radial factor is
\[
 \sqrt{h/m}=\left(1-\frac{h-m}{h}\right)^{-1/2}
     =1+\frac{D}{2h}+O_{\theta,s}(h^{-2}).
\]
Multiplication by the angular factor gives the expansion.
\end{proof}

Thus the divergence $D$ that fixes the shell transition at $b=D/2$ also controls the leading relative loss in the Gaussian problem far above that transition. For the $2:1$ example and $b=100$, the limiting relative gap is $1.431\times10^{-4}$ and $D/(2h_b)$ is $1.427\times10^{-4}$. The expansion concerns the limiting gains. The joint superexponential result in Section~\ref{sec:gaussian} determines the leading gains along $b_n\to\infty$. The coefficient of their relative difference in that joint limit remains open.

\section{Gain Stability and Limits Beyond Fixed Block Spectra}
\label{sec:continuum}
Section~\ref{conf:outlier-proof} uses the finite-width stability estimate \eqref{eq:gain-stability} to control the contribution of a single weak direction. The same estimate allows finite blocks to approximate a distribution of singular values, leading to Theorem~\ref{thm:continuum}.

The block calculus extends to spectra approximable in mean square.
At each width take a full-column-rank gradient with $\|G_n\|_*=1$.
Order its singular values in decreasing order and put
$a_n(t)=n\sigma_i(G_n)$ for $(i-1)/n<t\le i/n$. Then $\int a_n=1$. All integrals in this section use Lebesgue probability
measure on $[0,1]$. Let $a^{(\ell)}$ be the average of $a$ on each of
the $2^\ell$ equal dyadic intervals.

\begin{theorem}[Gain for a limiting spectrum]\label{thm:continuum}
Assume $a_n\to a$ in $L^2[0,1]$, where $a\ge0$ almost everywhere, and
let $B_n\to\infty$ with $(\log B_n)/n\to b\in[0,\infty)$.
Then $\int a=1$, since $|\int(a_n-a)|\le\|a_n-a\|_2$, where
$\|a_n-a\|_2^2=\int_0^1|a_n-a|^2$. Both the fixed-sphere and the Gaussian optimal gains
converge. In either model the limit is the increasing limit of the
corresponding block gains for $a^{(\ell)}$, with each block fraction
equal to $2^{-\ell}$ and each block mass equal to the integral of $a$
over that block.

For the fixed sphere the limiting gain equals the optimal polar-ray gain
$1$ if and only if
\[
 b\ \ge\ -\frac12\int\log a,\qquad
 -\int\log a\in[0,+\infty].
\]
We set $-\log 0=+\infty$. If $a$ vanishes on a set of positive measure,
or if its logarithm has a nonintegrable negative part, the threshold is
infinite and the limiting sphere gap is positive at every finite $b$.
For Gaussian inputs at $b=0$ the limiting optimal gain is $\|a\|_2$.
For $b>0$, let $w_\alpha=(1+\sqrt{1+4\alpha a^2})/2$, where
$\alpha>0$ is the unique solution of
\[
 \frac12\int (w_\alpha-1-\log w_\alpha)=b .
\]
The limiting gain is $(\int a^2/w_\alpha)^{1/2}$. For every finite
$b\ge0$ it exceeds the limiting Gaussian polar-ray gain $1/\sqrt{h_b}$
precisely when $a$ is nonconstant up to null sets.
\end{theorem}

\begin{proof}
The penalty under either isotropic input depends only on the singular
values of $Q$. Von Neumann's trace inequality bounds the linear term
by $\sum_i\sigma_i(G_n)\sigma_i(Q)$, achieved when the singular bases
are aligned. Let $x_i\ge0$
be its singular values, treated as a step function, and put $m=\|x\|_2$.
The linear gain is $\int a_n x$. For $m>0$, let
$p_i=x_i^2/(nm^2)$. Concavity of the square root gives
\[
 \mathbb E\|\operatorname{diag}(x)g/\sqrt n\|_2
 =m\,\mathbb E\sqrt{\textstyle\sum_i p_i g_i^2}
 \ \ge\ m\sum_i p_i\mathbb E|g_i|
 =\sqrt{2/\pi}\,m .
\]
For the same update $D_x=\operatorname{diag}(x)$, the Gaussian displacement
satisfies $\mathbb E\|D_xg/\sqrt n\|_2=\mathbb E R_n\,\mathbb E\|D_x\zeta\|_2$
by radial independence. Since $\mathbb E R_n\le1$, the sphere displacement also has mean at least $\sqrt{2/\pi}\,m$.
Writing $c_0=\sqrt{2/\pi}$ and using $(r-1)_+\ge r-1$, any update of
nonnegative gain satisfies
\[
 0\le \|a_n\|_2 m-B_n(c_0m-1),\qquad
 m\le \frac{B_n}{c_0B_n-\|a_n\|_2}\le\frac2{c_0}
\]
for large $n$, since $B_n\to\infty$ and $\|a_n\|_2$ stays bounded.
Block averaging contracts the $L^2$ norm, so this bound also holds for
each averaged spectrum. For an averaging $\widetilde a_n$, test an
optimizer for $a_n$ on $\widetilde a_n$. Its penalty is unchanged, and
the linear term changes by at most
$\|a_n-\widetilde a_n\|_2 m$ by Cauchy--Schwarz. Testing in the other
direction bounds the absolute difference between the optimal gains:
\begin{equation}\label{eq:gain-stability}
 |V_n(a_n)-V_n(\widetilde a_n)|
 \le \sqrt{2\pi}\,\|a_n-\widetilde a_n\|_2.
\end{equation}
The same argument applies to any two aligned profiles with uniformly
bounded $L^2$ norms and to near-maximizers of positive gain.

For fixed $\ell$, partition the indices at $\lfloor nj/2^\ell\rfloor$,
$j=0,\ldots,2^\ell$, and average $a_n$ over each resulting group.
All groups are nonempty for sufficiently large $n$ at fixed $\ell$.
The resulting profile tends in $L^2$ to $a^{(\ell)}$.
Indeed, averaging contracts $\|a_n-a\|_2$. At fixed $\ell$, each rounded interval differs from its dyadic interval on a set of length at most $2/n$. Cauchy--Schwarz bounds the difference of the integrals of $a$ there by $\|a\|_2\sqrt{2/n}$. Since each length tends to $2^{-\ell}$, the block averages converge. There are finitely many blocks, so the averages are uniformly bounded and the total length of the displaced boundaries tends to zero. This gives convergence in $L^2$. 
Some limiting block masses can be zero. To obtain the finite-block
value in this case, replace the averaged profile $v_n$ by
$(1-\varepsilon)v_n+\varepsilon$ for $0<\varepsilon<1$.
All limiting masses are then positive. The estimate just proved bounds
the change in finite-width gain by
$\sqrt{2\pi}\,\varepsilon\|1-v_n\|_2$, uniformly for large $n$.
Both limiting dual formulas minimize
$(\sum_j s_j^2/z_j)^{1/2}$ on a compact set with coordinates bounded
away from zero for fixed $\ell$ and finite $b$. The objective converges
uniformly there as the smoothed masses tend to the original masses.
Taking $n\to\infty$ and then $\varepsilon\downarrow0$ gives the
same dual value for nonnegative masses. On the sphere this value is one
exactly when $s$ belongs to the rate sublevel, by equality in
Cauchy--Schwarz. For the Gaussian at $b>0$ the formula
\eqref{eq:gauss-kkt} extends by continuity as well. The budget increases
strictly with $\alpha$ whenever some $s_j>0$, and bracketing its root
at two fixed values gives convergence under smoothing.

For a scale $x_J$ constant on an index
group $J$, its contribution to the linear gain of the original spectrum
is $n^{-1}\sum_{i\in J}a_{n,i}x_J=(|J|/n)\bar a_{n,J}x_J$.
The penalty is unchanged, so the two spectra agree on every
block-constant update. Block-rotation averaging gives an optimal
block-constant update for the averaged spectrum
(Theorem~\ref{thm:hinge-phase}). Its unrestricted gain therefore equals
the gain for the original spectrum optimized over block-constant scales.
The index partitions are nested.
Write $V_{n,\ell}$ for this block gain and $v_\ell=\lim_n V_{n,\ell}$
for the fixed-partition limit. Equation~\eqref{eq:gain-stability} gives
\[
 v_\ell\le\liminf_n V_n\le\limsup_n V_n
          \le v_\ell+\sqrt{2\pi}\,\|a-a^{(\ell)}\|_2 .
\]
First take $n\to\infty$ at fixed $\ell$ as in this display. Then take
$\ell\to\infty$: dyadic averages converge in $L^2$ and $v_\ell$ is
nondecreasing, hence $\lim_n V_n=\lim_\ell v_\ell$.

For the sphere the threshold at level $\ell$ is
$-\tfrac12\int\log a^{(\ell)}=\tfrac12\int\phi(a^{(\ell)})$,
where $\phi(t)=t-1-\log t$ is nonnegative and convex, with $\phi(0)=+\infty$. These integrals
increase under refinement, and each is at most $\tfrac12\int\phi(a)$
by conditional Jensen. Conversely $a^{(\ell)}\to a$ almost everywhere
by dyadic differentiation, so Fatou yields
$\int\phi(a)\le\liminf_\ell\int\phi(a^{(\ell)})$. This also admits
the value $+\infty$. If $b$ is below this limiting threshold, some finite
level has gain strictly above one, and that strict inequality persists
under refinement. At or above the threshold every level has gain one.

For the Gaussian model the block formula \eqref{eq:gauss-kkt} is
precisely $z_j^*/\theta_j=w_\alpha$ evaluated at the block average
$a_j$, since $s_j=\theta_j a_j$ in that formula.
For fixed $\alpha>0$, the map $t\mapsto (1+\sqrt{1+4\alpha t^2})/2$
on $[0,\infty)$ is $\sqrt\alpha$-Lipschitz. Also $\phi$ is
$1$-Lipschitz on $[1,\infty)$, so the block budgets converge to
$F(\alpha)=\tfrac12\int\phi(w_\alpha)$. For $0\le q\le Q$,
$0\le\phi(w_q)\le w_q-1\le\sqrt Q\,a$, so dominated convergence gives
continuity of $F$. Since $\int a=1$, the set $\{a>0\}$ has positive measure. On that set
the integrand increases strictly from zero to infinity with $q$, and
elsewhere it is zero. Monotone convergence gives $F(q)\to\infty$. For $b>0$ there is therefore a unique $\alpha$
with $F(\alpha)=b$. Write $F_\ell(q)$ for the budget using $a^{(\ell)}$.
For $0<q_-<\alpha<q_+$, $F(q_-)<b<F(q_+)$. Convergence at the two
fixed parameters gives $F_\ell(q_-)<b<F_\ell(q_+)$ for large $\ell$,
so each block root lies between $q_-$ and $q_+$. Taking both endpoints
toward $\alpha$ proves convergence of the roots.
The block optimal squared gain converges to $\int a^2/w_\alpha$
by the identity $a^2/w_\alpha=(w_\alpha-1)/\alpha$ and the same
bracketing applied to $\int w_\alpha$. At $b=0$ the direct formula
$\|a^{(\ell)}\|_2\to\|a\|_2$ suffices.

Finally $h=\int w_\alpha$ satisfies $\phi(h)\le2b$ by Jensen, hence
$h\le h_b$. Cauchy--Schwarz gives $1=(\int a)^2\le h\int a^2/w_\alpha$.
If $a$ is nonconstant and $b>0$, then $w_\alpha$ is nonconstant
and strict Jensen gives $\phi(h)<2b$. Since $\phi$ is strictly
increasing on $[1,\infty)$, $h<h_b$ and the squared gain exceeds
$1/h_b$. At $b=0$, $\|a\|_2^2-1=\int(a-1)^2>0$ for nonconstant $a$.
If $a=1$ almost everywhere, the inequalities are equalities.
\end{proof}

The fixed-sphere limit in this theorem also gives the gain on any shrinking bounded shell. Apply \eqref{eq:shell-sandwich} with the same profiles at budgets $\ell_nB_n$ and $h_nB_n$ with $\ell_n,h_n\to1$. Their logarithmic rates are unchanged, so the two bounds have the same limit.

\paragraph{A weak block occupying half the dimensions retains a positive shell gap.}
At even width take half the normalized singular values equal to
$2-\delta_n$ and half equal to $\delta_n$, with $0<\delta_n<1$ and
$\delta_n\to0$. Each gradient has full rank. The limiting nuclear shares
are $(1,0)$. In the sphere dual the squared gain is therefore the minimum
of $1/u_1$, attained at
$u_1=(1+\sqrt{1-e^{-4b}})/2<1$ for every finite $b$.
The limiting gap is $1/\sqrt{u_1}-1>0$, also on the matched shell.
The single small direction in Section~\ref{conf:outlier-proof} occupies
a vanishing dimension fraction and has a vanishing gain gap.

\section{Polar Steps for Rank-Deficient Gradients}\label{app:rank}
The criterion in Section~\ref{sec:criterion} uses a full-column-rank gradient, for which \(P\) preserves every input length. For the compact polar factor \(P_r\) of a nonzero rank-deficient gradient, the sphere hinge has a nonstationary point at \(cP_r\). We also determine when a different positive length can give a global minimum.
Let \(G=U_r\Sigma_rV_r^\top\) have rank \(0<r<n\), where \(U_r,V_r\) have \(r\) orthonormal columns and \(\tau=\operatorname{tr}\Sigma_r\). Its compact polar factor is \(P_r=U_rV_r^\top\). For a uniform unit input \(\zeta\), the squared displacement under \(P_r\) is
\[
 U=\|P_r\zeta\|_2^2=\|V_r^\top\zeta\|_2^2,
 \qquad P_r^\top P_r=V_rV_r^\top .
\]
In orthogonal coordinates beginning with the columns of \(V_r\), represent \(\zeta=g/\|g\|_2\) with \(g\) standard Gaussian. The sums of squares in the first \(r\) and remaining \(n-r\) coordinates are independent \(X\sim\chi_r^2\) and \(Y\sim\chi_{n-r}^2\). Their ratio \(U=X/(X+Y)\) has the \({\rm Beta}(r/2,(n-r)/2)\) law, with positive density on \((0,1)\).

Consider the hinge \(H(s)=B(s-c)_+\), \(c>0,\ B>0\). At \(cP_r\) the displacement \(c\sqrt U\) is strictly below the kink almost surely. For every fixed matrix direction \(D\), the penalty has zero pointwise derivative there, and \(B\|D\|_{\rm op}\) bounds its absolute difference quotient. Dominated convergence gives zero for the derivative of the expected penalty. Thus \(\nabla J_{\rm sph}(cP_r)=-G\ne0\).

Can another length \(t>0\) yield a global optimum on this ray? For a given input, the penalty gradient is \(B(Q\zeta)\zeta^\top/\|Q\zeta\|_2\) when \(\|Q\zeta\|_2>c\) and zero below \(c\). At \(Q=tP_r\) the norm is \(t\sqrt U\), so the factor \(t\) cancels on the active event. The equality \(t\sqrt U=c\) has probability zero. The same difference-quotient bound therefore gives the gradient of the expected penalty as
\[
 B\,\mathbb E\!\left[\mathbf1_{t\sqrt U>c}
      \frac{P_r\zeta\zeta^\top}{\sqrt U}\right]
 =\frac Br\,\mathbb E[\sqrt U\,\mathbf1_{t\sqrt U>c}]P_r.
\]
In the right singular basis, coordinate sign changes cancel all mixed terms. Permutations of the first \(r\) coordinates make the entries in the active diagonal block equal. Their sum before multiplication by \(B\) is \(\mathbb E[\sqrt U\,\mathbf1_{t\sqrt U>c}]\). These facts determine the coefficient of \(P_r\).

Optimality on the ray requires the derivative in the direction \(P_r\) to vanish. Taking the Frobenius inner product of the loss gradient with \(P_r\), and using \(\|P_r\|_F^2=r\), gives the scalar equation
\[
 \tau=\langle G,P_r\rangle
 = B\,\mathbb E[\sqrt U\,\mathbf1_{t\sqrt U>c}].
\]
The right side is zero up to \(t=c\) and then increases continuously and strictly to \(B\mathbb E\sqrt U\). Thus a finite ray optimum exists and is unique exactly when \(B\mathbb E\sqrt U>\tau\). To see what happens at the boundary, write the gain relative to zero as
\[
 t\tau-B\mathbb E(t\sqrt U-c)_+
 =t(\tau-B\mathbb E\sqrt U)+B\mathbb E\min(t\sqrt U,c).
\]
If \(B\mathbb E\sqrt U=\tau\), the gain increases to \(Bc\) as \(t\to\infty\) because \(U>0\) almost surely. It stays below \(Bc\) at every finite \(t\), since \(\Pr(t\sqrt U<c)>0\). If \(B\mathbb E\sqrt U<\tau\), the gain diverges to infinity. When a ray optimum exists, global optimality over matrices requires the entire gradient to vanish, that is,
\[
 G=\frac Br\,\mathbb E[\sqrt U\,\mathbf1_{t\sqrt U>c}]P_r.
\]
At the optimal ray scale this is equivalent to all \(r\) positive singular values of \(G\) being equal to \(\tau/r\).

\section{Sample Size for Empirical Optimality on the Sphere}\label{sec:sample}
Section~\ref{sec:criterion} averages over every sphere direction. For a finite sample and \(A=0\), optimality at some finite right slope becomes a convex-hull membership problem. Corollary~\ref{cor:samples} uses the sphere estimates to bound the median number of independent uniform directions needed for membership.
Let \(n\ge2\). Condition on a full-column-rank \(G\), normalize \(\tau=1\), and set \(C=S/\tau\) and \(A=0\). For unit directions \(v_1,\ldots,v_N\), consider the objective with input law \(\nu_N=N^{-1}\sum_{i=1}^N\delta_{v_i}\) and put \(X_i=v_iv_i^\top\). At \(cP\), every sample has displacement \(c\). The directional derivative argument of Appendix~\ref{app:selector} therefore gives the exact condition for empirical optimality,
\[
 C=N^{-1}\sum_{i=1}^N h_iX_i,\qquad 0\le h_i\le B .
\]
Taking traces gives \(\sum_i h_i/N=1\). Thus some finite slope suffices exactly when \(C\in\operatorname{conv}\{X_1,\ldots,X_N\}\). For any realizing convex weights \(w_i\), taking \(h_i=Nw_i\) works with \(B=N\). A smaller cap specified in advance requires convex weights with \(w_i\le B/N\). For \(C=I_n/n\), this feasibility question is frame scalability \citep{frames}. For the probability bound below, draw the \(v_i\) independently and uniformly from the sphere, independently of \(G\).

\begin{corollary}[Median sample size]\label{cor:samples}
For a positive definite trace-one target \(C\), let
\[
 N_{1/2}=\min\{N:\Pr(C\in\operatorname{conv}\{X_1,\ldots,X_N\})\ge1/2\},
\]
where probability is over the independently sampled directions. Write \(T=nC,\ D=-\tfrac12\log\det T,\ d=n(n+1)/2-1\), and \(U_n=\min(500n,800\sqrt{n\|T\|_{\rm op}})\). Then
\[
 D-\log 2\le\log N_{1/2}\le D+\log\bigl((3d+1)U_n\bigr).
\]
\end{corollary}
\begin{proof}
Centering each projector at \(C\) gives an independent copy of \(X-C\), where \(X=vv^\top\) for a uniform \(v\). These centered matrices belong to the space of symmetric matrices satisfying one trace constraint, of dimension \(d\). For the origin in this space the halfspace depth is
\(\alpha=\inf_{H=H^\top}\Pr\{\operatorname{tr}(H(X-C))\ge0\}\).
The sample hull contains \(C\) precisely when the centered hull contains the origin. Thus \citet[Theorem~16]{hull} gives
\[
 \frac{1}{2\alpha}\le N_{1/2}\le
 \left\lceil\frac{3d}{\alpha}\right\rceil\le \frac{3d+1}{\alpha}.
\]
Here \(\alpha=d_{\rm H}(C)\) and \(D=I_W(T)\). Equation~\eqref{eq:depth-determinant} gives \(D\le-\log\alpha\le D+\log U_n\), hence \(e^{-D}/U_n\le\alpha\le e^{-D}\). Substitution gives \(e^D/2\le N_{1/2}\le (3d+1)U_ne^D\), as claimed.
\end{proof}
The determinant thus gives \(\log N_{1/2}=D+O(\log n)\). For a flat target \(C=I_n/n\), \(D=0\), yet every combination of \(N\) rank-one projectors has rank at most \(N\), so \(N_{1/2}\ge n\). Independence enters the probability bound. If all \(v_i\) coincide with one uniform draw, each marginal remains uniform and the sample hull consists of a single rank-one matrix. A gradient computed on the same data as the directions would require their joint law.

\section{Frozen-Checkpoint Protocol and Results}
\label{app:checkpoints}
Section~\ref{sec:checkpoint-tests} compares held-out losses after a single update and tests how loss changes under a rotation of the output directions. Here we specify how the gradients, step lengths and paired document intervals were obtained. We also report the sensitivity of both endpoints to evaluation text, matrix, amplitude and training stage.

\subsection{Training and held-out text}

The unit of evaluation was one document at one frozen weight matrix. The same update and its competitors were scored on the same text. Training seeds varied the initialized weights, and two shared document panels allowed us to vary the data used to fit and evaluate a step at a given set of weights.

We trained a 124M-parameter decoder with 12 Transformer blocks, width 768 and 12 attention heads on GPT-2-tokenized FineWeb \citep{fineweb}. Initialization seeds 20261121, 20261122 and 20261123 saw identical training tokens in identical order. All training and measurements ran sequentially on one A100 PCIe 40\,GB using PyTorch 2.5.0 and CUDA 11.8. Each update used 512 sequences of length 1,024 (microbatch size eight, 64 accumulation steps). Each seed saw 402,653,184 prediction targets by step 768. Training used BF16 autocast for the forward pass. For the hidden-layer weight matrices, Muon \citep{muon} used momentum 0.95, five Newton--Schulz iterations and a peak learning rate of 0.01. The remaining parameters, including the shared token embedding and output weight, used AdamW with peak learning rate 0.0036, \((\beta_1,\beta_2)=(0.9,0.95)\), \(\epsilon=10^{-8}\) and zero weight decay. After \(k\) completed updates the next matrix learning rate was \(0.01\min\{1,(768-k)/192\}\), giving \(0.01/192\) for the last update. We saved all three models at steps 192, 576 and 768. Before training we fixed step 768 as the primary endpoint, the other two stages as secondary, and the matrices, panels, step grid and comparisons specified below.

We fixed the two disjoint panels before training and reused them at every stage of every run. Within each panel, sets \(A,B,C\) contained 32 gradient, 64 calibration and 128 evaluation documents. We selected documents by a salted hash from the first 4,096 records of shard 014 of FineWeb \emph{sample-10BT}, then used a second hash to choose one contiguous window of 513 tokens per document. With the first token as context, each window supplies 512 prediction targets. The 448 document hashes were distinct and disjoint from the 590,291 hashes in the training and prior-study exclusion index. None of the 448 token windows matched a substring of the complete training stream. The audit covered exact duplicates only. Every loss below is mean next-token cross entropy with the natural logarithm, in nats per predicted token.

At each of the nine checkpoints, we perturbed the \(768\times768\) key and value projections in blocks 0, 5 and 11 (indexed from zero). We recomputed the gradient from the frozen weights for each panel. An intervention changed only the selected matrix, and the code restored its original weights before the next intervention. We used FP32 for gradients and forward passes, disabled TF32, computed the SVD in FP64 and accumulated token losses into document averages in FP64. All remeasured baseline document losses agreed with their original values. Each candidate used exactly the same evaluation texts and frozen values of the other parameters. There were 18 combinations of seed, stage and panel, each with six matrix measurements.

\subsection{Constructing the polar, two-block and raw-gradient updates}

Take the current gradient \(G_A=U\operatorname{diag}(\sigma_i)V^\top\) of the mean loss over \(A\), with \(\sigma_1\ge\cdots\ge\sigma_{768}\). The polar candidate uses the full SVD and sets each singular value to one. The raw-gradient candidate keeps every \(\sigma_i\). The two-block candidate keeps the singular vectors but replaces the singular values within each half of the ordered spectrum by their mean. Previous checkpoint work studied allocations along momentum singular directions \citep{wu_spectral}. Here the current gradient determines two relative lengths which we apply jointly to one matrix.

\paragraph{Why the two spectral means determine the relative block lengths.}
Let \(I_j\) denote half \(j\), \(k_j=|I_j|\), \(j(i)\) the half containing index \(i\), and \(\bar\sigma_j=k_j^{-1}\sum_{i\in I_j}\sigma_i\). For a fixed aligned update \(Q=U\operatorname{diag}(x_{j(i)})V^\top\) with one length \(x_j\ge0\) per half, \(U^\top U=I\) gives \(G_A^\top Q=V\operatorname{diag}(\sigma_i x_{j(i)})V^\top\). Taking the trace and using \(V^\top V=I\) yields
\[
 \langle G_A,Q\rangle
 = \sum_{j=1}^2\sum_{i\in I_j}\sigma_i x_j
 = \sum_{j=1}^2 k_j\bar\sigma_j x_j .
\]
Thus the linear gain for every such \(Q\) is unchanged if we substitute
\(\widetilde G=U\operatorname{diag}(\bar\sigma_{j(i)})V^\top\) for \(G_A\). Holding \(Q\) fixed also preserves its displacement \(-Qz\) on every input \(z\) and hence its radial penalty. This gives exact equality of surrogate gains on the restricted class (Appendix~\ref{sec:continuum}).

To read the limiting allocation from Theorem~\ref{thm:hinge-phase}, write \(n\) for width and \(T_n=\sum_{i=1}^n\sigma_i\) for the nuclear norm. In the normalized spectrum \(\widetilde G/T_n\), the fraction of singular-value mass in block \(j\) is \(s_{j,n}=k_j\bar\sigma_j/T_n\), and its dimension fraction is \(\theta_{j,n}=k_j/n\). Therefore
\[
   \frac{s_{j,n}}{\theta_{j,n}}=\frac{n}{T_n}\bar\sigma_j .
\]
For uniform inputs on the unit sphere, take \(n\to\infty\) with positive limiting fractions \(s_j,\theta_j\), and use the displacement penalty \(H_n(r)=B_n(r-1)_+\) with \(B_n\to\infty\) and \((\log B_n)/n\to0\). The theorem gives limiting optimal block lengths proportional to \(s_j/\theta_j\). Since \(n/T_n\) is common to both blocks, we use the two measured means as relative weights at width 768. This last step proposes a direction to test on the network. We evaluated it on actual network losses without imposing a sphere or hinge assumption.

\paragraph{Normalizing the directions and fitting their step lengths.}
We normalize each candidate \(D\) to \(\|D\|_F=\sqrt{768}\). The two-block direction becomes \(D_{\rm two}=\alpha\widetilde G\) with \(\alpha=\sqrt{768}/\|\widetilde G\|_F\). For each direction separately we select the smallest minimizer \(t_D\) of the mean calibration loss \(L_B(W-tD)\) over the common 16-point grid
\[
\begin{aligned}
 \mathcal T={}&\{0\}\cup
 \{0.01/k:k\in\{384,192,96,48,24,12\}\}\\
 &{}\cup\{0.00125,0.0025,0.005,0.01,0.02,0.04,0.08,0.16,0.32\}.
\end{aligned}
\]
The two-block step lengths are \(t_{D_{\rm two}}\alpha\bar\sigma_1\) and \(t_{D_{\rm two}}\alpha\bar\sigma_2\). Both vanish when \(t_{D_{\rm two}}=0\). For example, at the step-768 value projection of block 0, seed 20261121 and panel 1 yielded means 0.00488106 and 0.000512674, a ratio of 9.52. Calibration selected \(t_{D_{\rm two}}=0.01\), so the two actual lengths were 0.01406 and 0.001477. The common grid fits one scalar for each candidate, including the raw gradient. The evaluation documents are reserved for scoring the selected steps.

\subsection{Held-out gain difference and output-rotation contrast}
\label{app:checkpoint-estimands}

For evaluation document \(i\), let \(\ell_i(W)\) be its loss at the frozen matrix and let
\(g_i(D)=\ell_i(W)-\ell_i(W-t_DD)\) be its absolute gain. For the first endpoint, take the paired difference
\[
 g_i(D_{\rm two})-g_i(D_{\rm pol})
 =\ell_i(W-t_{D_{\rm pol}}D_{\rm pol})-\ell_i(W-t_{D_{\rm two}}D_{\rm two}) .
\]
We call the weighted mean of these differences \(G_{\rm test}\). Positive values favor two blocks. The two fitted candidates are evaluated on the same documents. By comparison, \(\Delta_n\) in \eqref{conf:gap} uses the exact unrestricted and ray optima of the surrogate.

For the second endpoint, start with the polar direction \(D\) and use two fixed signed row permutations \(O\) (seeds 104729 and 130363). Define
\[
 E_{i,t}(D)=\tfrac12\{\ell_i(W+tD)+\ell_i(W-tD)\}-\ell_i(W),
 \qquad R_{i,t}(O)=E_{i,t}(OD)-E_{i,t}(D).
\]
Since \(O^\top O=I\), each layer input \(x\) has the same displacement length \(t\|Dx\|\) under both orientations and both signs. The two signs cancel the linear term exactly in \eqref{eq:su}, whatever the document gradient. For a radial penalty the remaining even cost in either orientation is \(\mathbb E[H(t\|Dx\|)]-H(0)\), so its predicted difference is zero for every finite \(t\) and any input law. The measured even loss uses actual model evaluations at finite \(t\) and can include even terms above second order. The second endpoint \(R_t\) averages \(R_{i,t}(O)\) over both permutations at prespecified \(t=0.01\). We also measured \(t=0.0025\) and \(0.04\). These perturbation amplitudes scale the weight change. The \(B_n\) of the surrogate specifies the slope of a penalty.

\paragraph{Paired document intervals conditional on the three checkpoints at each stage.}
Each endpoint gives equal weight to documents within a panel, the six matrices, the two panels and the three specified seeds. For each of 10,000 bootstrap replicates (seed 11781), we sampled 128 evaluation document indices with replacement independently in each panel. Within a panel the same draw was used for all seeds, stages, matrices, methods, permutations and signs. For each endpoint \(j\), first average its contrasts over the fixed seeds, matrices and, when applicable, permutations within each document. If \(v_{pj}\) is the sample variance of these averages across documents in panel \(p\), the standard error is \(\mathrm{SE}_j=\sqrt{(v_{1j}+v_{2j})/(4\cdot128)}\). Let \(\hat\mu_j\) be its observed mean and \(\hat\mu_j^*\) its bootstrap mean. For the two primary intervals we took the 95th percentile \(q\) of \(\max_j\left(|\hat\mu_j^*-\hat\mu_j|/\mathrm{SE}_j\right)\) and reported \(\hat\mu_j\pm q\,\mathrm{SE}_j\). Both standard errors were positive and \(q=2.2234\). Other document intervals are ordinary 2.5th and 97.5th bootstrap percentiles, except for the separate rotation families specified below. These intervals describe document sampling conditional on the observed checkpoints, chosen matrices, permutations and fixed \(A,B\) samples. The shared documents support paired stage and seed comparisons. The intervals do not quantify variation over a population of training seeds.

\subsection{Final-step results and sensitivity checks}

Table~\ref{tab:checkpoint-audit} separates relative loss differences from absolute improvements over zero. All effects in this subsection are in \(10^{-6}\) nats per token. The polar direction had lower mean held-out cross entropy than the two-block direction at step 768. Against the zero step, both point estimates of improvement were negative and their document intervals included zero. The raw-gradient comparison was also unresolved. The seed means of \(G_{\rm test}\) ranged from \(-16.83\) to \(-2.85\) with sample standard deviation 7.30. For \(R_{0.01}\) the range was \([-28.51,-25.31]\) and the sample standard deviation was 1.62.

\begin{table}[htbp]
\centering
\caption{Step-768 effects averaged over the three fixed seeds, two panels and six matrices, in \(10^{-6}\) nats per token. Gain over zero is the unperturbed loss minus the post-step loss. The two primary intervals are simultaneous across these two endpoints only.}
\label{tab:checkpoint-audit}
\begin{tabular}{@{}lrrl@{}}
\toprule
Endpoint & Estimate & 95\% interval & Interval type\\
\midrule
Polar loss minus two-block loss, \(G_{\rm test}\) & \(-8.63\) & \([-13.91,-3.35]\) & Joint\\
Rotated minus original even cost, \(R_{0.01}\) & \(-26.78\) & \([-27.49,-26.08]\) & Joint\\
Polar gain over zero & \(-15.59\) & \([-38.90,7.17]\) & Ordinary\\
Two-block gain over zero & \(-24.22\) & \([-49.77,0.63]\) & Ordinary\\
Raw-gradient gain over zero & \(-14.70\) & \([-35.61,6.24]\) & Ordinary\\
Raw-gradient loss minus two-block loss & \(-9.52\) & \([-19.84,0.78]\) & Ordinary\\
\bottomrule
\end{tabular}
\end{table}

\paragraph{Linear gain and finite-step remainder at the selected amplitudes.}
Table~\ref{tab:frozen-gain-components} in Section~\ref{sec:checkpoint-tests} defines the linear gain \(a_D\) and finite-step remainder \(u_D\). For each of the 36 final-step fits, we evaluated both terms on the same held-out documents at the amplitude previously selected on \(B\). Computing the paired two-block minus polar contrast before rounding gives \(2.378\) in linear gain and \(11.004\) in additional remainder, for a measured difference of \(-8.626\), all in \(10^{-6}\) nats per token.

To check how well local curvature accounts for this remainder, use the symmetric finite difference from the calibration documents at \(h=0.0025\),
\[
 \widehat k_B(D)=
 \frac{L_B(W+hD)+L_B(W-hD)-2L_B(W)}{h^2}.
\]
Replacing \(u_D\) with \(t_D^2\widehat k_B(D)/2\) yields a mean two-block minus polar gain of \(-7.59\). Across the 36 fits, the mean absolute error in the paired gain contrast falls from 29.04 using the held-out linear term alone to 1.85 with this quadratic correction. Both calculations use the evaluation gradient and the same previously selected amplitudes. This diagnostic locates the discrepancy in the cost of the actual steps. It cannot apportion that cost among output orientation, input directions and scalar step selection. The finite difference measures curvature along a weight-space direction. Inferring the hinge slope would additionally require a choice of displacement units and kink and a common length-dependent penalty consistent with these orientations.

\paragraph{Layer-input energy in the upper gradient spectral half.}
Let \(V_1\) contain the right singular vectors in the top spectral half of \(G_A\). For each fit we measured \(p_A=\sum_x\|V_1^\top x\|^2/\sum_x\|x\|^2\) over the layer inputs on \(A\). A uniform spherical population assigns half its energy to any fixed half-dimensional subspace. Across the 36 fits, \(p_A\) ranged from 0.823 to 0.881 even though the within-fit coefficient of variation of input radii was only 0.004 to 0.026. The empirical root-mean-square displacement of the two-block direction divided by that of polar, at the same weight amplitude, is
\[
 \left[\frac{2\{p_A\bar\sigma_1^2+(1-p_A)\bar\sigma_2^2\}}
 {\bar\sigma_1^2+\bar\sigma_2^2}\right]^{1/2}.
\]
It ranged from 1.280 to 1.325. Hence Frobenius normalization, which equalizes the population mean squared displacement under spherical inputs, leaves the two-block displacement appreciably larger on these inputs. The gradient and activation diagnostics share the fitting sample, so the range describes their observed alignment.

\paragraph{Swapping evaluation panels with the fitted update fixed.}
Table~\ref{tab:checkpoint-panels} shows all four combinations of fit and evaluation panel. Rows correspond to initializations. Each diagonal fit uses its own \(A,B,C\) and contributes to the primary endpoint. For the off-diagonal comparisons, we evaluated the already chosen direction and step from panel 1 on \(C_2\), or the direction and step from panel 2 on \(C_1\). This adds 36 single-matrix evaluations and keeps the weights, directions and selected steps fixed when changing the evaluation text. Two of the six seed-by-panel diagonal means were slightly positive. All four means across the fixed seeds favored polar. The primary orientation means for panels 1 and 2 were \(-26.16\) and \(-27.41\), respectively.

\begin{table}[htbp]
\centering
\caption{Polar minus two-block cross entropy at step 768 by fit and evaluation panel, averaged over six matrices, in \(10^{-6}\) nats per token. A positive number favors two blocks.}
\label{tab:checkpoint-panels}
\begin{tabular}{@{}lrrrr@{}}
\toprule
 & \multicolumn{2}{c}{Fit on \(A_1,B_1\)} & \multicolumn{2}{c}{Fit on \(A_2,B_2\)}\\
\cmidrule(lr){2-3}\cmidrule(lr){4-5}
Seed & Evaluate on \(C_1\) & Evaluate on \(C_2\) & Evaluate on \(C_1\) & Evaluate on \(C_2\)\\
\midrule
20261121 & \(-6.15\) & \(-4.44\) & \(-4.28\) & \(0.45\)\\
20261122 & \(0.91\) & \(-6.07\) & \(-2.91\) & \(-13.29\)\\
20261123 & \(-18.95\) & \(-22.42\) & \(-11.22\) & \(-14.71\)\\
Mean     & \(-8.07\) & \(-10.98\) & \(-6.13\) & \(-9.19\)\\
\bottomrule
\end{tabular}
\end{table}

\paragraph{Rotation contrasts by matrix and amplitude.}
Table~\ref{tab:checkpoint-rotations} reports the rotation contrasts by matrix, amplitude and permutation. For each of the six seed--panel pairs at step 768, the maximum standardized bootstrap deviation gave a simultaneous 95\% band over all 36 contrasts (six matrices, two permutations, three amplitudes). Every interval lay below zero. At \(t=0.01\), any shared radial prediction for the original and rotated even costs has maximum absolute error at least \(|R_t|/2\) on their two means, by the triangle inequality. The estimated bound is 13.39, with a lower confidence bound of 13.04 from the primary joint interval. Across the three increasing amplitudes, \(2R_t/t^2\) was \(-0.537,-0.536,-0.536\) nats per token per squared weight amplitude, consistent with near-quadratic scaling over this range.

\begin{table}[htbp]
\centering
\caption{Rotated minus original even cost, averaged over three seeds and two panels at step 768, in \(10^{-6}\) nats per token. \(O_1,O_2\) denote the two prespecified signed row permutations. K and V denote the key and value projections.}
\label{tab:checkpoint-rotations}
\begin{tabular}{@{}lrrrrrr@{}}
\toprule
&\multicolumn{2}{c}{\(t=0.0025\)}&\multicolumn{2}{c}{\(t=0.01\)}&\multicolumn{2}{c}{\(t=0.04\)}\\
\cmidrule(lr){2-3}\cmidrule(lr){4-5}\cmidrule(lr){6-7}
Matrix & \(O_1\) & \(O_2\) & \(O_1\) & \(O_2\) & \(O_1\) & \(O_2\)\\
\midrule
Block 0, K & \(-2.66\)&\(-2.68\)&\(-42.46\)&\(-42.73\)&\(-680.30\)&\(-684.47\)\\
Block 0, V & \(-5.82\)&\(-5.76\)&\(-92.89\)&\(-92.14\)&\(-1487.96\)&\(-1476.19\)\\
Block 5, K & \(-0.47\)&\(-0.46\)&\(-7.48\)&\(-7.41\)&\(-119.61\)&\(-118.51\)\\
Block 5, V & \(-0.88\)&\(-0.88\)&\(-14.12\)&\(-13.95\)&\(-225.99\)&\(-223.35\)\\
Block 11, K & \(-0.11\)&\(-0.12\)&\(-1.79\)&\(-1.78\)&\(-28.78\)&\(-28.66\)\\
Block 11, V & \(-0.14\)&\(-0.14\)&\(-2.35\)&\(-2.31\)&\(-37.63\)&\(-37.04\)\\
\bottomrule
\end{tabular}
\end{table}

\paragraph{Selected amplitudes and grid endpoints.}
At step 768, none of 108 candidate fits reached the upper grid endpoint (Table~\ref{tab:checkpoint-steps}). Polar minus two-block loss averaged \(-36.20\) in the 15 fits selecting the same positive step, \(+13.68\) in the 17 selecting different steps, and zero in the four selecting zero for both. Calibration determined these groups, so the comparison is descriptive. At steps 192 and 576, respectively, 14 and five of 108 fits reached the grid maximum.

\begin{table}[htbp]
\centering
\caption{Final-step choices for all six matrices. A triplet gives the grid indices for polar, two-block and raw-gradient steps, respectively. In ascending order, indices 0 through 6 denote zero followed by the six \(0.01/k\) values in \(\mathcal T\). Indices 7 through 15 denote the nine values from 0.00125 to 0.32. Column labels give block index and projection.}
\label{tab:checkpoint-steps}
\begin{tabular}{@{}llcccccc@{}}
\toprule
Seed & Panel & 0 K & 0 V & 5 K & 5 V & 11 K & 11 V\\
\midrule
20261121 & 1 & 9/9/6 & 10/10/6 & 11/10/8 & 10/9/6 & 11/10/7 & 12/11/8\\
20261121 & 2 & 0/0/0 & 10/9/5 & 11/11/8 & 11/11/7 & 4/8/0 & 11/10/7\\
20261122 & 1 & 0/0/0 & 10/9/6 & 11/10/3 & 11/11/7 & 11/11/7 & 11/11/7\\
20261122 & 2 & 10/9/5 & 10/10/6 & 9/8/5 & 11/10/6 & 10/8/9 & 12/11/8\\
20261123 & 1 & 8/8/4 & 9/8/4 & 0/0/0 & 10/10/6 & 12/12/9 & 13/13/9\\
20261123 & 2 & 0/0/0 & 10/10/5 & 11/10/7 & 8/8/5 & 11/11/8 & 12/11/7\\
\bottomrule
\end{tabular}
\end{table}

\paragraph{Earlier stages and short-schedule pilot.}
At steps 192 and 576, respectively, \(G_{\rm test}\) was \(-320.81,-89.88\) and \(R_{0.01}\) was \(-110.48,-33.93\). From step 576 to 768 the paired-document change in \(G_{\rm test}\) was \(81.26\) (ordinary 95\% interval \([40.28,121.37]\)). All nine checkpoints came from Muon trajectories.

In a separate pilot with learning-rate decay ending at step 192, polar-minus-two-block losses were \(-8.29\) and \(-0.40\) on an RTX 4090 and \(+13.32\) on an A100, in seed order. Each seed used different documents. The 768-step study held hardware, data and calibration budgets fixed across seeds.

\end{document}